%% file: main.tex
\pdfoutput=1
\documentclass{article}

\usepackage{microtype}
\usepackage{graphicx}
\usepackage{subfigure}
\usepackage{booktabs} 
\usepackage{amsmath}
\usepackage{amsthm}
\usepackage{thmtools}
\usepackage{algorithm}
\usepackage[noend]{algcompatible}
\usepackage{xspace}
\usepackage{enumitem}
\usepackage{color}

\usepackage{hyperref}

\PassOptionsToPackage{numbers, compress}{natbib}
\usepackage[preprint]{nips_2018}

\usepackage[utf8]{inputenc} 
\usepackage[T1]{fontenc}    
\usepackage{hyperref}       
\usepackage{url}            
\usepackage{booktabs}       
\usepackage{amsfonts}       
\usepackage{nicefrac}       
\usepackage{microtype}      

\input{macro}

\title{Interpreting Neural Network Judgments via \\ Minimal, Stable, and Symbolic Corrections}

\author{Xin Zhang \\ CSAIL, MIT \\ \And Armando Solar-Lezama \\ CSAIL, MIT \\ \And Rishabh Singh \\ Google Brain}

\begin{document}

\maketitle

\begin{abstract}
We present a new algorithm to generate minimal, stable, and symbolic corrections to an input that will cause a neural network with ReLU activations to change its output. We argue that such a correction is a useful way to provide feedback to a user when the network's output is different from a desired output. 
Our algorithm generates such a correction by solving a series of linear constraint satisfaction
problems.
The technique is evaluated on three neural network models:
one predicting whether an applicant will pay a mortgage,
one predicting whether a first-order theorem can be proved efficiently
by a solver using certain heuristics,
and the final one judging whether a drawing is an accurate rendition of a canonical drawing of a cat.

\end{abstract}

\input{intro}
\input{problem}
\input{algo}

\input{experiment}
\input{related}
\input{conclusion}

\bibliographystyle{abbrv}
\bibliography{main}

\input{appendix}

\end{document}

%% file: macro.tex
\def\av#1{\boldsymbol{#1}}
\def\funname#1{{\footnotesize\textsf{#1}}}
\def\dis{\funname{dis}}
\def\t#1{\text{#1}}
\def\tool{{\sc{Polaris}}\xspace}

\def\true{{\footnotesize\textsf{true}}}
\def\false{{\footnotesize\textsf{false}}}

\declaretheoremstyle[notefont=\bfseries]{bfnamethmstyle}
\declaretheoremstyle{exstyle}

\declaretheorem{theorem}
\declaretheorem[name=Definition,sibling=theorem,style=definition]{definition}

\def\PARAM#1{\STATE \textbf{\footnotesize PARAM} #1}

%% file: intro.tex
\section{Introduction}
\label{sec:intro}

When machine learning is used to make decisions about people in the real world, it is extremely important to be able to explain the rationale behind those decisions. Unfortunately, for systems based on deep learning, it is often not even clear what an explanation means; showing someone the sequence of operations that computed a decision provides little actionable insight. There have been some recent advances towards making deep neural networks more interpretable (e.g. \cite{NN:Understanding}) using two main approaches: i) generating input prototypes that are representative of abstract concepts corresponding to different classes~\cite{nguyen2016synthesizing} and ii) explaining network decisions by computing relevance scores to different input features~\cite{bach2015pixel}. However, these explanations 
do not provide direct actionable insights regarding how to cause the prediction to move from an undesirable class to a desirable class.

In this paper, we argue that for the specific class of \emph{judgment problems}, minimal, stable, and symbolic corrections are an ideal way of explaining a neural network decision. We use the term judgment to refer to a particular kind of binary decision problem where a user presents some information to an algorithm that is supposed to pass judgment on its input.  The distinguishing feature of judgments relative to other kinds of decision problems is that they are asymmetric; if I apply for a loan and I get the loan, I am satisfied, and do not particularly care for an explanation; even the bank may not care as long as on aggregate the algorithm makes the bank money. On the other hand, I very much care if the algorithm denies my mortgage application. The same is true for a variety of problems, from college admissions, to parole, to hiring decisions. In each of these cases, the user expects a positive judgment, and would like an actionable explanation to accompany a negative judgment. 

We argue that a \emph{correction} is a useful form of feedback; what could I have done differently to elicit a positive judgment? For example, if I applied for a mortgage, knowing that I would have gotten a positive judgment if my debt to income ratio (DTI) was 10\% lower is extremely useful; it is actionable information that I can use to adjust my finances. We argue, however, that the most useful corrections are those that are minimal, stable and symbolic. 

First, in order for a correction to be actionable, the corrected input should be as similar as possible from the original offending input. For example, knowing that a lower DTI would have given me the loan is useful, but knowing that a 65 year old billionaire from Nebraska would have gotten the loan is not useful. Minimality must be defined in terms of an error model which specifies which inputs are subject to change and how. For a bank loan, for example, debt, income and loan amount are subject to change within certain bounds, but I will not move to another state just to satisfy the bank. 

Second, the suggested correction should be stable, meaning that there should be a neighborhood of points surrounding the suggested correction for which the outcome is also positive. For example, if the algorithm tells me that a 10\% lower DTI would have gotten me the mortgage, and then six months later I come back with a DTI that is 11\% lower, I expect to get the mortgage, and will be extremely disappointed if the bank says, ``oh, sorry, we said 10\% lower, not 11\% lower''. So even though for the neural network it may be perfectly reasonable to give positive judgments to isolated points surrounded by points that get negative judgments, corrections that lead to such isolated points will not be useful. 

Finally, even if the correction is minimal and robust, it is even better if rather than a single point, the algorithm can produce a \emph{symbolic} correction that provides some insight about the relationship between different variables. For example, knowing that for someone like me the bank expects a DTI of between 20\% and 30\% is more useful than just knowing a single value. And knowing something about how that range would change as a function of my credit score would be even more useful still. 

In this paper, we present the first algorithm capable of 
computing minimal stable symbolic corrections. 
Given a neural network with ReLU activation,
 our algorithm produces a symbolic description of a space of corrections such that any correction in that space will change the judgment. In the limit, the algorithm will find the closest region with a volume above a given threshold. 
Internally, our algorithm reduces the problem into a series of linear constraint
satisfaction problems, which are solved using the Gurobi linear programming (LP) solver~\cite{gurobi}.
 We show that in practice, the algorithm is able to find good symbolic corrections in 12 minutes on average for small but realistic networks. 
 While the running time is dominated by solver invocations, only under 2\% of it is spent is spent on actual solving and
 the majority of the time is spent on creating these LP instances.
 We evaluate our approach on three neural networks: one predicting whether an applicant will pay a mortgage,
 one predicting whether a first-order theorem can be proved efficiently
 by a solver using certain heuristics,
 and the other judging whether a drawing is an accurate rendition of a canonical drawing of a cat.

\begin{figure}
	\begin{center}
	\begin{minipage}{0.3\textwidth}
		\begin{center}
			\includegraphics[width=0.99\linewidth]{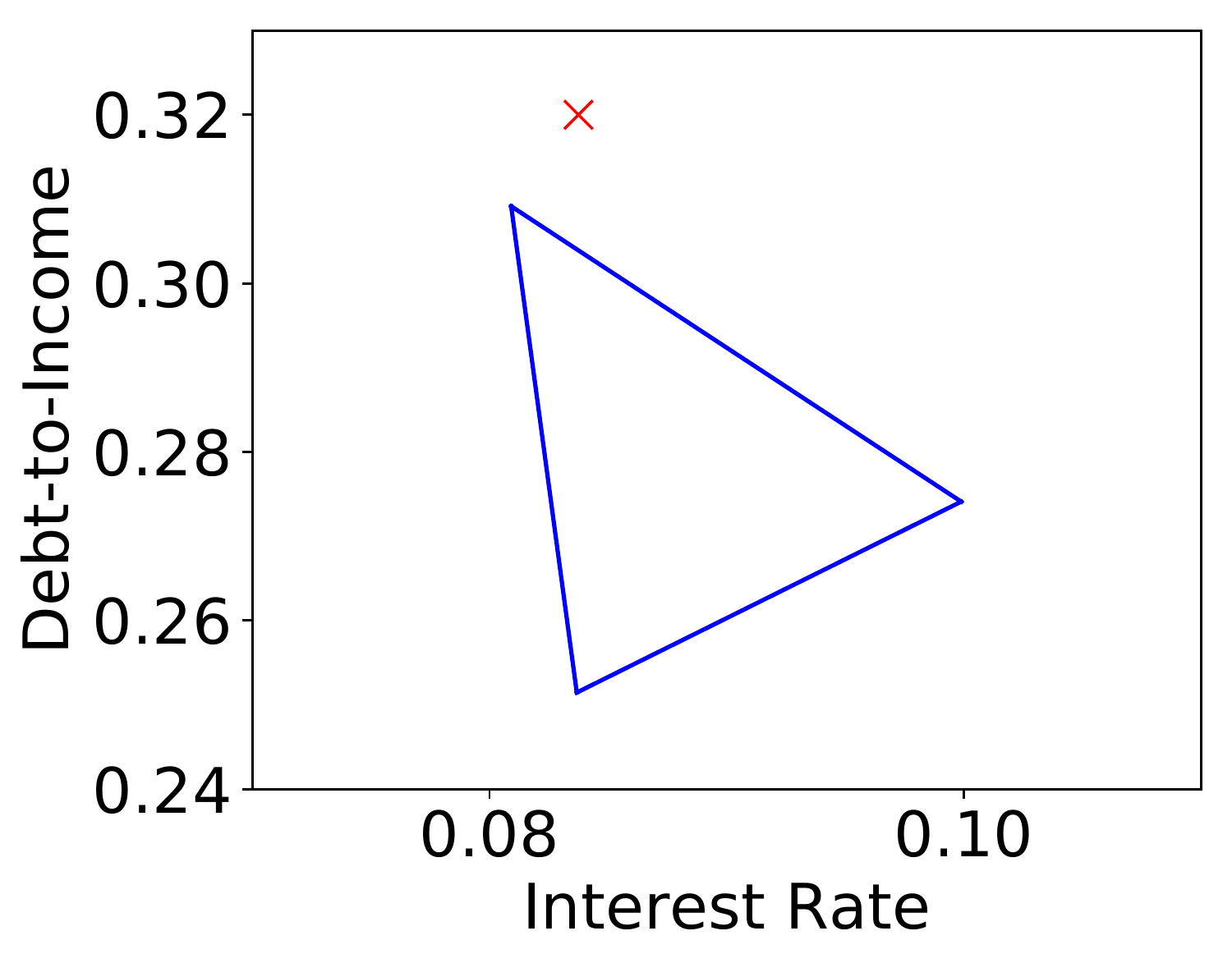}\\
			\vspace{-0.08in}
			{\small (a) Mortgage Underwriting}
		\end{center}
	\end{minipage}
	\begin{minipage}{0.3\textwidth}
		\begin{center}
		\includegraphics[width=0.9\linewidth]{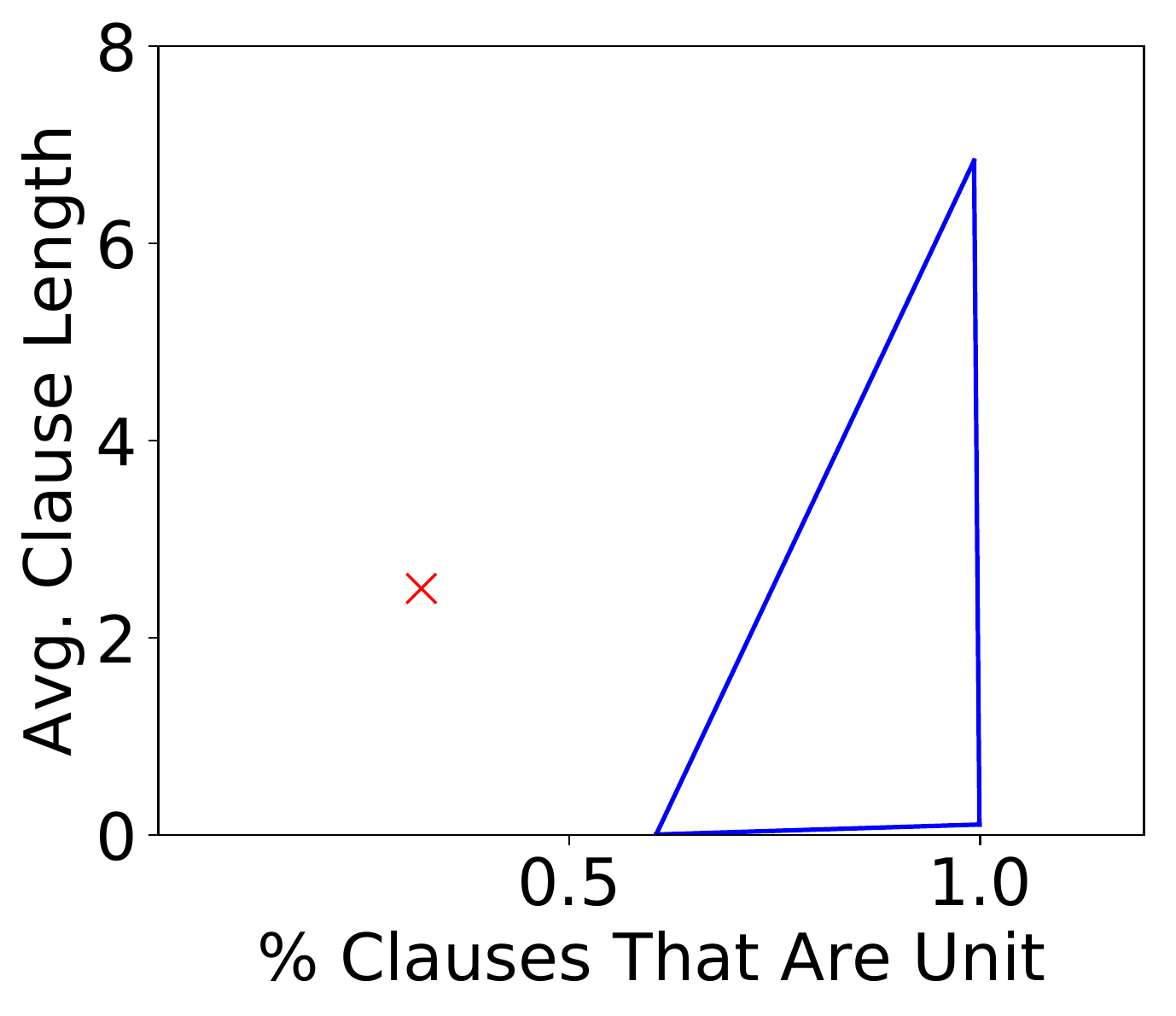} \\
					\vspace{-0.08in}
	{\small	(b) Theorem Proving}
		\end{center}
	\end{minipage}
	\begin{minipage}{0.3\textwidth}
		\begin{center}
		\includegraphics[width=\linewidth]{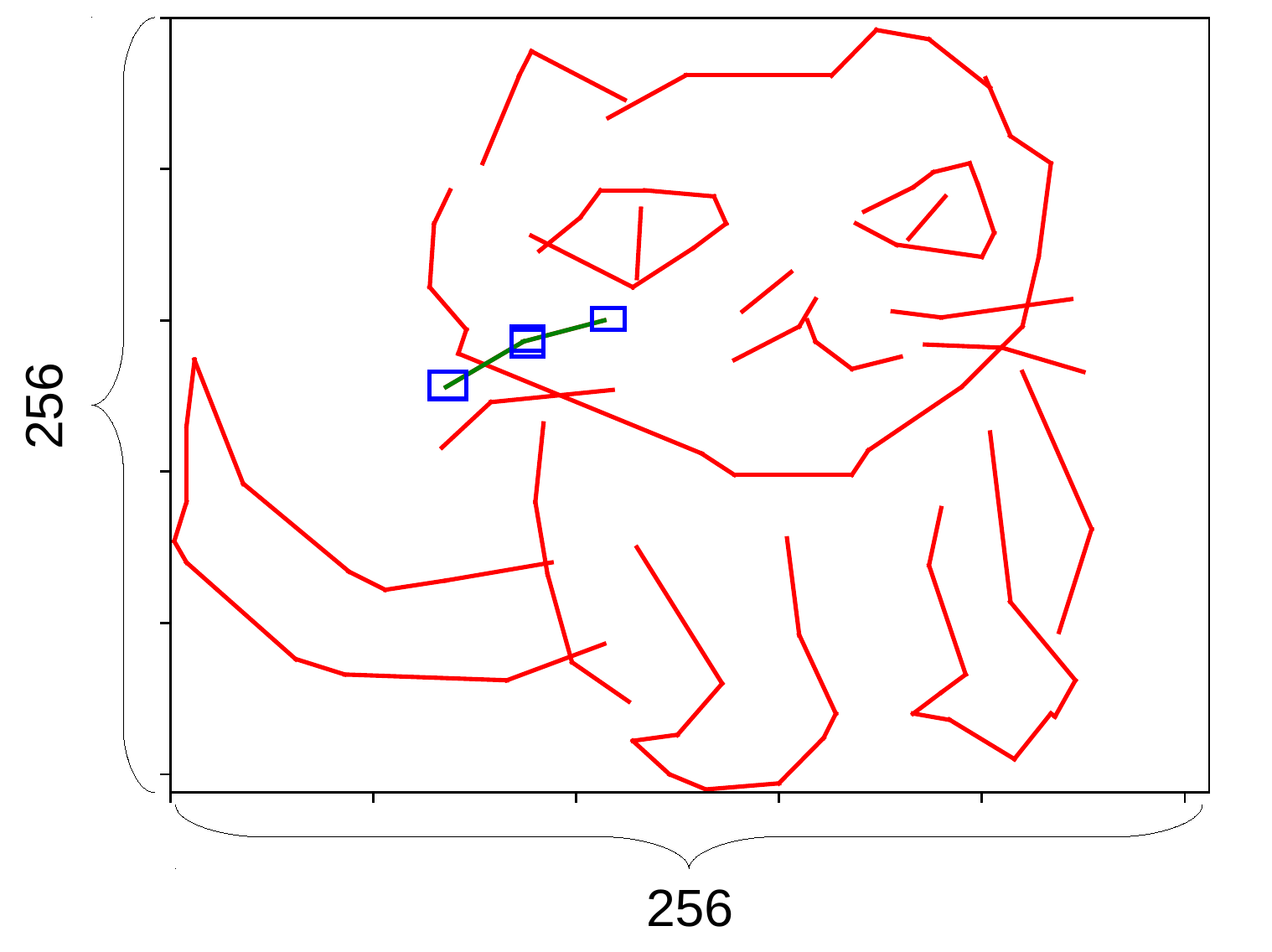} \\
					\vspace{-0.05in}
		{\small(c) Drawing Recognition}
		\end{center}
	\end{minipage}
	\end{center}
	\vspace{-0.16in}
	\caption{Symbolic explanations generated by our approach for neural networks in different domains.}
	\label{fig:showcase}
	\vspace{-0.3in}
\end{figure}
{\bf Explanation showcases.} Figure~\ref{fig:showcase} shows example explanations generated by our approach on the aforementioned networks.
Figure~\ref{fig:showcase}(a) suggests a mortgage applicant to change DTI and interest rate in order to get their application accepted.
While the red cross represents the original application, the blue triangle represents the symbolic correction (i.e. the region of points that all lead to a positive outcome).
Since the user may only be able to change DTI and interest rates often vary between applications, it is essential to provide a symbolic correction
rather than a concrete correction to make the feedback actionable.
Figure~\ref{fig:showcase}(b) suggests a user to reformulate a first-order theorem when the network predicts it as challenging to solve.
Intuitively, either reducing the problem size (by decreasing average clauses lengths) or providing a partial solution (by adding unit clauses) would reduce the problem complexity.
Finally, Figure~\ref{fig:showcase}(c) shows how to add lines to a drawing so that it gets recognized by the network as a canonical cat drawing.
The red lines represent the original input, while the blue boxes represent the symbolic correction and the green lines represent one concrete correction in it.
Briefly, any concrete correction whose vertices fall into the blue boxes would make the drawing pass the network's judgment.
Comparing to the previous two corrections which only involves 2 features, this correction involves 8 features (coordinates of each vertex).
This highlights our approach's ability to generate high-dimension complex corrections.

%% file: problem.tex
\section{Background and Problem Definition}
\label{sec:problem}

%
%
We first introduce some notations we will use in explaining our approach.
Suppose $F$ is a neural network with ReLU activation. 
In the model we consider, the input to $F$ is a (column) vector $\av{v}_0$ \footnote{Unless specified, all vectors in the paper are by columns.} of size $s_0$. The network computes the output of each layer as 
\[
\av{v}_{i+1} = f_i(\av{v}_i ) = \funname{ReLU}(\av{W}_i \av{v}_i + \av{b}_i)
\]
Where $\av{W}_i$ is an $s_{i+1} \times s_i$ matrix, $\av{b}_i$ is a vector of size $s_{i+1}$, and $\funname{ReLU}$ applies the rectifier function elementwise to the output of the linear operations.

We focus on classification problems, where the classification of input $\av{v}$ is obtained by
\[
	l_F(\av{v}) \in \funname{argmax}_i F(\av{v})[i].
\]
We are specifically focused on binary classification problems (that is, $l_F(\av{v}) \in \{0,1\}$).
The {\it judgment problem} is a special binary classification problem where one label is preferable than
the other.
We assume 1 is preferable throughout the paper.

The {\it judgment interpretation problem} concerns providing feedback in the form of corrections when $l_F(\av{v}) = 0$.
A correction $\av{\delta}$ is a real vector of input vector length such that 
$l_F(\av{v} + \av{\delta}) = 1$.
As mentioned previously, a desirable feedback should be a minimal, stable, and symbolic correction.
We first introduce what it means for a concrete correction $\av{\delta}$ to be 
minimal and stable. Minimality is defined in terms of a norm $\|\av{\delta}\|$
on $\av{\delta}$ that measures the
distance between the corrected input and the original input.
For simplicity, we use $L_1$ norm to measure the sizes of all vectors throughout Section~\ref{sec:problem} and Section~\ref{sec:algo}.
We say $\av{\delta}$ is $e$-stable if for any $\av{\delta}'$ such that if $\|\av{\delta} - \av{\delta}'\| \leq e$, we have
$l_F(\av{\delta}') = 1$.

A symbolic correction $\av{\Delta}$ is a connected set of concrete corrections.
More concretely, we will use a set of linear constraints to represent a symbolic correction.
We say a symbolic correction is $e$-stable if there exists a correction $\av{\delta} \in \av{\Delta}$ such that for
any $\av{\delta}'$ where $\|\av{\delta}' - \av{\delta} \| \leq e$, we have $\av{\delta}' \in \av{\Delta} $.
We call such a correction a stable region center inside $\av{\Delta}$.
To define minimality, we define the distance of $\av{\Delta}$ from the original input using the distance of a stable region center that has the smallest distance among all stable region centers.
More formally:
\[
\dis_e(\av{\Delta}) := \funname{min}_{\av{\delta} \in S} \| \av{\delta}\|,
\]
where $S := \{ \av{\delta} \in \av{\Delta} \mid \forall \av{\delta}'. \| \av{\delta}' - \av{\delta} \| \leq e  \implies \av{\delta}' \in \av{\Delta} \}$.
When $\av{\Delta}$ is not $e$-stable, $S$ will be empty, so we define $\dis_e(\av{\Delta}) := \infty$.

We can now define the judgment interpretation problem.

\begin{definition}{({\it Judgment Interpretation})}
Given a neural network $F$, an input vector $\av{v}$ such that $l_F(\av{v}) = 0$, and a real value $e$, a 
judgment interpretation is an $e$-stable symbolic correction $\av{\Delta}$ with the minimum distance
among all $e$-stable symbolic corrections.
	
\end{definition}

%% file: algo.tex
\section{Our Approach}
\label{sec:algo}

\begin{figure}
\def\funnames#1{{\fontsize{7pt}{9pt}\selectfont \textsf{#1}}}
\begin{minipage}[t]{0.485\textwidth}
	\ \\
	\ \\
	\ \\
	\ \\
	\ \\
	\ \\
\begin{algorithm}[H]\footnotesize
	\begin{algorithmic}[1]
		\REQUIRE{A neural network $F$ and an input vector $\av{v}$ such that $l_F(\av{v}) = 0$.}
		\ENSURE{A judgment interpretation $\av{\Delta}$.}
		\PARAM{A real value $e$ and an integer number $n$.}
		\STATE{$\av{S_n} := \{ \av{s} \mid \av{s} \text{ is a subarray of }[1,...,|\av{v}|]\linebreak\text{ with length }n\}$}
		\STATE{$\av{\Delta} := None, d := +\infty$}
		\FOR{ $\av{s} \in \av{S_n}$ }
		\STATE{ $\av{\Delta}_s := \funnames{findProjectedInterpretation}(F,\av{v},\av{s}, e)$}
		\IF{$\funnames{dis}_e(\av{\Delta}_s) < d$}
		\STATE{$\av{\Delta} := \av{\Delta}_s$, $d := \funnames{dis}_e(\av{\Delta}_s)$}
		\ENDIF
		\ENDFOR
		\STATE{ \bf{return} $\av{\Delta}$}
	\end{algorithmic}
	\caption{Finding a judgment interpretation.}	
	\label{alg:main}
\end{algorithm}
\end{minipage}
\ \ 
\begin{minipage}[t]{0.51\textwidth}
\begin{algorithm}[H]\footnotesize
	\begin{algorithmic}[1]
		\REQUIRE{A neural network $F$, an input vector $\av{v}$, an integer vector $\av{s}$, and a real number $e$.}
		\ENSURE{A symbolic correction $\av{\Delta}_s$ that only changes features indexed by $\av{s}$.}
		\PARAM{An integer $m$, the maximum number of verified linear regions to consider.}
		\STATE{$\t{regions} := \emptyset, \t{workList} := []$}
		\STATE{$\av{\delta}_0 := \funnames{findMinimumConcreteCorrection}(F, \av{v}, \av{s}$)}
		\STATE{$\av{a}_0 := \funnames{getActivations}(F,\av{\delta}_0 +\av{v})$}
		\STATE{$L_0 := \funnames{getRegionFromActivations}(F, \av{a}_0, \av{v}, \av{s})$}
		\STATE{$\t{regions} := \t{regions}\cup \{L_0 \} $}
		\STATE{$\t{workList} := \funnames{append}(\t{workList}, \av{a}_0 )$}
		\WHILE{$\funnames{len}(\t{workList})!= 0$ }
		\STATE{$\av{a} := \funnames{popHead}(\t{workList})$}
		\FOR{$p \in [1, \funnames{len}(\av{a})]$}	
		\IF{$\funnames{checkRegionBoundary}(F, \av{a}, p, \av{v}, \av{s})$}
		\STATE{$\av{a'}:= \funnames{copy}(\av{a})$}
		\STATE{$\av{a'}[p] := \neg\av{a'}[p]$}
		\STATE{$L' := \funnames{getRegionFromActivations}(F, \av{a'}, \av{v}, \av{s})$}
		\IF{$L' \notin \t{regions}$}
		\STATE{$\t{regions} := \t{regions}\cup \{ L' \}$}
		\IF{$\funnames{len}(\t{regions}) = m$}
		\STATE{$\t{workList} := []$}
		\STATE{\bf{ break}}
		\ENDIF
		\STATE{$\t{workList} := \funnames{append}(\t{workList}, \av{a}' )$}
		\ENDIF
		\ENDIF
		\ENDFOR
		\ENDWHILE
		
		\STATE{{\bf return} $\funnames{inferConvexCorrection}(\t{regions})$}
	\end{algorithmic}
	\caption{\funname{findProjectedInterpretation}}	
	\label{alg:search}
\end{algorithm}
\end{minipage}
\end{figure}
%

Algorithm~\ref{alg:main} outlines our approach to find a judgment interpretation for a given neural network $F$ and an input vector $\av{v}$.
Besides these two inputs, it is parameterized by a real $e$ and an integer $n$.
The former specifies the radius parameter in our stability definition, while the latter
specifies how many features are allowed to vary to produce the judgment interpretation.
We parameterize the number of features to change as high-dimension interpretations can be 
hard for end users to understand.
For instance, it is very easy for a user to understand if the explanation says their mortgage would be approved as long as they change the DTI and the credit score while keeping the other features as they were.
On the other hand, it is much harder to understand an
an interpretation that involves all features (in our experiment, there are 21 features for the mortgage underwriting domain).
The output is a judgment interpretation that is expressed in a system of linear constraints, which are in the form of 
\[
\av{A} \av{x}  +\av{b} \geq 0,
\]
where $\av{x}$ is a vector of variables, $\av{A}$ is a matrix, and $\av{b}$ is a vector.
%
%

Algorithm~\ref{alg:main} finds such an interpretation by iteratively invoking $\funname{findProjectedInterpretation}$ (Algorithm~\ref{alg:search}) to find an interpretation that varies a list of $n$ features $\av{s}$.
It returns the one with the least distance.
Recall that the distance is defined as 
$
\dis_e(\av{\Delta}) = \funname{min}_{\av{\delta} \in S} \| \av{\delta}\|,
$
which can be evaluated by solving a sequence of linear programming problems when $L_1$ norm is used.

We next discuss $\funname{findProjectedInterpretation}$ which is the heart of our approach.

\subsection{Finding a Judgment Interpretation along given features}

In order to find a judgment interpretation, we need to find a set of linear constraints that are {\bf minimal}, {\bf 
stable}, and {\bf verified} (that is, all corrections satisfying it will make the input classified as 1).
None of these properties are trivial to satisfy given the complexity of any real-world neural network.

We first discuss how we address these challenges at a high level, then dive into the details of the algorithm.
To address minimality, we first find a single concrete correction that is minimum by leveraging an existing adversarial
example generation technique~\cite{adv-gradient} and then generate a symbolic correction by expanding upon it.
To generate a stable and verified correction, 
we exploit the fact that ReLU-based neural networks are piece-wise linear functions.
Briefly, all the inputs that activate the same set of neurons can be characterized by a set of linear constraints.
We can further characterize the subset of inputs that are classified as 1 by adding an additional linear constraint.
Therefore, we can use a set of linear constraints to represent a set of verified concrete corrections under certain activations.
We call this set of corrections a {\emph{verified linear region}} (or {\emph{region} for short).
We first identify the region that the initial concrete correction belongs to, then grow the set of regions by identifying
regions that are connected to existing regions.
Finally, we infer a set of linear constraints whose concrete corrections are a subset of ones enclosed by the set of discovered regions.
Algorithm~\ref{alg:search} details our approach, which we describe below.

{\bf Generating the initial region.}
We first find a minimum concrete correction $\av{\delta}_0$ by leveraging a modified version of
the fast signed gradient method~\cite{adv-gradient} that minimizes the $L_1$ distance (on line 3).
More concretely, starting with a vector of $0$s, we calculate $\av{\delta}_0$ by iteratively adding a modified gradient that takes the sign of the most significant 
dimension among the selected features until $l_F(\av{v} + \av{\delta}_0) = 1$.
For example, if the original gradient is $[0.5, 1.0, 6.0, -6.0]$, the modified gradient would be
$[0,0,1.0,0]$ or $[0,0,0,-1.0]$.
Then we obtain the ReLU activations $\av{a_0}$ for $\av{v}+\av{\delta}_0$ (by invoking $\funname{getActivations}$ on line 4), which is a Boolean vector where each Boolean value represents whether a given neuron is activated.
Finally, we obtain the initial region that $\av{\delta}_0$ falls into by invoking $\funname{getRegionFromActivations}$ (on line 5), which is defined below:

\begin{small}
$
\begin{array}{@{}r@{\,}c@{\,}l}
\funname{getRegionFromActivations}(F, \av{a}, \av{v}, \av{s}) &:=&  \funname{activationConstraints}(F, \av{a}, \av{v})\ \wedge \ 
	\funname{classConstraints}(F, \av{a}, \av{v})	 \\
&&
\wedge \ 
	\funname{featureConstraints}(\av{s})
	, 
\end{array}
$
\end{small}

\begin{small}
$
\begin{array}{@{}r@{\,}c@{\,}l}
	\text{\normalsize{where }} \funname{activationConstraints}(F, \av{a}, \av{v}) &:=& \quad\ \, \bigwedge_{j \in [1, k]}
	\bigwedge_{m \in [1, |f_j|]}
	\{ G_r^{\av{a}}(\av{x}+\av{v}) \geq 0  \text{ if } \av{a}[r] = \true \} 
	 \\
	&& \wedge \ \  
	 \bigwedge_{j \in [1, k]}
	\bigwedge_{m \in [1, |f_j|]}
	\{ G_r^{\av{a}}(\av{x}+\av{v}) < 0  \text{ if }  \av{a}[r] = \false\} , \\
	&& \text{where } G_r^{\av{a}}(\av{x}+\av{v}):=\av{w}_r \cdot f_0^{\av{a}}(f_1^{\av{a}}(...f_{m-1}^{\av{a}}(\av{x}+\av{v}))) + b_r, \\
	&& \qquad\qquad\quad\quad \quad r := \sum_{i \in [1, j-1]} |f_i| + m
	\\[5pt]
	 \funname{classConstraints}(F, \av{a}, \av{v}) &:=& F^{\av{a}}(\av{x}+\av{v})[1] > F^{\av{a}}(\av{x}+\av{v})[0], \\[5pt]
	 \funname{featureConstraints}(\av{s}) &:=& \bigwedge_{j \notin \av{s}} \av{x}[j] = 0.
	 
\end{array}
$
\end{small}

In the definition above, we use the notation $f_i^{\av{a}}$ to refer to layer $i$ with its activations ``fixed'' to $\av{a}$. More formally, 
$
f_i^{\av{a}}(\av{v}_i) = \av{W}^{\av{a}}_i \av{v}_i + \av{b}_i^{\av{a}}
$
where $\av{W}^{\av{a}}_i $ and $\av{b}_i^{\av{a}}$ have zeros in all the rows where the activation indicated that rectifier in the original layer had produced a zero.
We use $k$ to represent the number of ReLU layers and $|f_j|$ to represent the number of neurons in the $j$th layer.
Integer $r$ indexes the $m$th neuron in $j$th layer.
Vector $\av{w}_r$ and real number $b_r$ are the weights and the bias of neuron $r$ respectively.
Intuitively, $\funname{activationConstraints}$ uses a set of linear constraints to encode
the activation of each neuron.

{\bf Expanding to connecting regions.} After generating the initial region, Algorithm 1 tries to grow the set of
concrete corrections by identifying regions that are connected to existing regions (line 6-20).
How do we know whether a region is connected to another efficiently?
There are $2^n$ regions for a network with $n$ neurons and checking whether two
sets of linear constraints intersect can be expensive on high dimensions.
Intuitively, two regions are likely connected if their activations only differ by one ReLU.
However, this is not entirely correct given a region is not only constrained by
the activations by also the desired classification.

Our key insight is that, {\emph{since a ReLU-based neural network is a continuous function,
two regions are connected if their activations differ by one neuron, and there are
concrete corrections on the face of one of the corresponding convex hulls, and this face corresponds to the differing neuron.
}}
Intuitively, on the piece-wise function represented by a neural network,
the sets of concrete corrections in two adjacent linear pieces are connected 
if there are concrete corrections on the boundary between them.
Following the intuition, we define \funname{checkRegionBoundary}:

\begin{small}
$
\begin{array}{@{}r@{\,}c@{\,}l}
	\funname{checkRegionBoundary}(F, \av{a}, p, \av{v}, \av{s}) &: = &
\funname{isFeasible}(
\funname{boundaryConstraints}(F, \av{a}, \av{v} , p)  \\ 
&& \qquad\quad\ \ \ \wedge\ 
	\funname{classConstraints}(F, \av{a}, \av{v})
\wedge \ 
	\funname{featureConstraints}(\av{s}))
\end{array}
$	
\end{small}
where

\begin{small}
$
\begin{array}{@{}r@{\,}c@{\,}l}
	\funname{boundaryConstraints}(F, \av{a}, p, \av{v}) &:=& \quad 
		\bigwedge_{j \in [1, k]}
	\bigwedge_{m \in [1, |f_j|]}
	\{ G_r^{\av{a}}(\av{x}+\av{v}) = 0  \text{ if } r = p  \} 
	 \\
		&& \wedge \

\bigwedge_{j \in [1, k]}
	\bigwedge_{m \in [1, |f_j|]}
	\{ G_r^{\av{a}}(\av{x}+\av{v}) \geq 0  \text{ if } \av{a}[r] = \true \text{ and }
r!= p \} 
	 \\
&& \wedge \ 
%
 \bigwedge_{j \in [1, k]}
	\bigwedge_{m \in [1, |f_j|]}
	\{ G_r^{\av{a}}(\av{x}+\av{v}) < 0  \text{ if }  \av{a}[r] = \false \text{ and }
 r!= p \} \\
		\end{array} 
$

$
\begin{array}{@{}l}
	\quad \text{\normalsize{where} } G_r^{\av{a}}(\av{x}+\av{v}):= \av{w}_r \cdot f_0^{\av{a}}(f_1^{\av{a}}(...f_{m-1}^{\av{a}}(\av{x}+\av{v}))) + \av{b}_r \text{ \normalsize{and} }
	r := \sum_{i \in [1, j-1]} |f_i| + m.
	\\[5pt]
\end{array}
$
\end{small}

By leveraging $\funname{checkRegionBoundary}$, Algorithm~\ref{alg:search}
uses a worklist algorithm to identify regions that are connected or 
transitively connected to the initial region until no more such regions
can be found  
or the number of discovered regions reaches a predefined upper bound $m$ (line 8-20).

{\bf Infer the final explanation.} Finally, Algorithm~\ref{alg:search} infers a set of linear constraints
whose corresponding concrete corrections are contained in the discovered regions.
Moreover, to satisfy the stability constraint, we want this set to be as large as possible.
Intuitively, we want to find a convex hull (represented by the returning
constraints) that is contained in a polytope (represented by the regions),
such that the volume of the convex hull is maximized.
Further, we infer constraints that represent relatively simple shapes, such as
simplexes or boxes, for two reasons.
First, explanations in simpler shapes are easier for the end user
to understand;
secondly, it is relatively efficient to calculate the volume of a simplex or a box.

The procedure $\funname{inferConvexCorrection}$ implements the above process using a greedy algorithm.
In the case of simplexes, we first randomly choose a discovered region and randomly sample a simplex inside it.
Then for each vertex, we move it by a very small distance in a random direction such that
(1) the simplex is still contained in the set of discovered regions,
and (2) the volume increases.
The process stops until the volume cannot be increased further.
For boxes, the procedure is similar except that we move the surfaces rather than the vertices.

Note that our approach is sound but not optimal or complete.
In other words, whenever Algorithm~\ref{alg:main} finds a symbolic correction,
the correction is verified and stable,
but it is not guaranteed to be minimal.
Also, when our approach fails to find a stable symbolic correction, it does not mean that such corrections do
not exist.
However, in practice, we find that our approach is able to find stable corrections for most of the time
and the distances of the discovered corrections are small enough
to be useful (as we shall see in Section~\ref{sec:exp.results}). 

\autoref{sec:ext} discusses several extensions to our approach including how to handle categorical features.



%% file: experiment.tex
\section{Empirical Evaluation}
\label{sec:experiment}

We evaluate our approach on three neural network models from different domains.

\subsection{Experiment Setup}
{\bf Implementation.} We implemented our approach in a tool called \tool,
which is written in three thousand lines of Python code.
To implement $\funname{findMinimumConcreteCorrection}$, we used a customized version
of the CleverHans library~\cite{cleverhans}.
To implement $\funname{isFeasible}$ which checks feasibility of generated linear constraints,
we applied the commercial linear programming solver Gurobi 7.5.2~\cite{gurobi}.

\begin{table}[t]
	\caption{Summary of the neural networks used in our evaluation.}
	\label{tab:networks}
	\vspace{-0.1in}
	\begin{small}
		\begin{tabular}{@{}l@{\ \ \ }l@{\ \ }c@{\ \ \ }l@{\ }c@{\ }c@{\ }c@{}}
			\hline
			Application & Network Structure & \# ReLUs & Dataset (train/val./test: 50/50/25) &\# features & F1  & Accuracy\\
			\hline
			\begin{tabular}{@{}l@{}}Mortgage\\ underwriting\end{tabular} & \begin{tabular}{@{}l@{}} 5 dense layers of\\ 200 ReLUs each\end{tabular} & 1,000  &  \begin{tabular}{@{}l@{}}Applications and performance of \\ 34 million Single-Family loans \cite{fanniemae} \end{tabular} & 21 & 0.118 & 0.8 \\
			\hline
			\begin{tabular}{@{}l@{}}Theorem \\ proving\end{tabular} &  \begin{tabular}{@{}l@{}}8 dense layers of \\100 ReLUs each \end{tabular} & 800 &
			\begin{tabular}{@{}l@{}}Statistics of 6,118 first-order\\theorems and their solving times \cite{theoproving}\end{tabular} & 51 & 0.74 & 0.792
			\\ 
			\hline
			\begin{tabular}{@{}l@{}}Drawing\\ recognition\end{tabular}& 
			\begin{tabular}{@{}l@{}}
				3 1-D conv. layers\\(filter shape: [5,4,8])\\ and 1 dense layer\\ of 1,024 ReLUs
			\end{tabular}
			& 4,096 & \begin{tabular}{@{}l@{}}
				0.12 million variants of a canonical\\ cat drawing and 0.12 million of cat\\ drawings from Google QuickDraw\cite{quickdraw}
	\end{tabular} & 512 & 0.995 & 0.995 \\
	\hline
\end{tabular}
\end{small}
\vspace{-0.2in}
\end{table}

{\bf Neural networks.} Table~\ref{tab:networks} summarizes the statistics of the neural networks.
The \emph{mortgage underwriting} network predicts whether an applicant would default on the loan.
Its architecture is akin to state-of-the-art neural networks for predicting mortgage risks~\cite{mortgagenet},
and has a recall of 90\% and a precision of 6\%.
It is trained to have a high recall to be conservative in accepting applications.
The \emph{theorem proving network} predicts whether a first-order theorem can be solved efficiently by a solver based
on static and dynamic characteristics of the instance.
We chose its architecture using a grid search.
The \emph{drawing recognition} network judges whether a drawing is an accurate rendition of a canonical drawing of a cat.
A drawing is represented by a set of line segments on a $256\times256$ canvas, each of which is represented by the coordinates
of its vertices.
A drawing comprises up to 128 lines, which leads to 512 features.

{\bf Evaluation inputs.} For the first two applications, we randomly chose 100 inputs in the test sets that were rejected by the networks.
For drawing recognition, we used 100 variants of the canonical drawing and randomly removed subsets of line segments so that they get rejected
by the network.  

{\bf Algorithm configurations.} Our approach is parameterized by
the number of features $n$ allowed to change simultaneously,
the maximum number of regions to consider $m$, the stability metric, the distance metric, and the shape of the generated symbolic correction.
We set $n=2$ for mortgage underwriting and theorem proving as corrections of higher dimensions on them
are hard for end users to understand.
Moreover, we limit the mutable features to 5 features each that are plausible for the end user to change.
Details of these features are described in Appendix \ref{sec:ext-setup}.
As for drawing recognition, we set $n\in [1,20]$, which allows us to add up to 5 line segments.
To reduce the computational cost, we use a generative network to recommend the features to change rather than
enumerating all combinations of features.
The network is a variational autoencoder that completes drawing sketches~\cite{sketchrnn}.
For the stability metric and the distance metric, we use a weighted $L_\infty$ norm and a weighted $L_1$ respectively
for both mortgage underwriting and theorem proving,
%
which are described in Appendix \ref{sec:ext-setup}.
For drawing recognition, we measure the distance of a correction by the number of features changed ($L_0$), which
reflects how many lines are added.
We say a correction is stable if it contains at least 3 pixels in each dimension.
Finally, we use triangles to represent the corrections for mortgage underwriting and theorem proving, while we use axis-aligned
boxes for drawing recognitions.
The blue rectangles in Figure~\ref{fig:showcase}(c) are projections of a box correction on coordinates of added line vertices.

{\bf Experiment environment.}
All the experiments were run on a Dell XPS 8900 Desktop with 16GB RAM and an Intel I7 4GHZ quad-core processor running Ubuntu 16.04.

\subsection{Experiment Results}
\label{sec:exp.results}
We first discuss how often \tool generates stable corrections and 
how far away these corrections are from the original input.
%
%
Then, we study the efficiency of \tool.
Finally, we discuss the effect of varying $m$, the maximum number of regions to consider.
%

\begin{figure}[t]
	\begin{center}
	\includegraphics[width=0.31\linewidth]{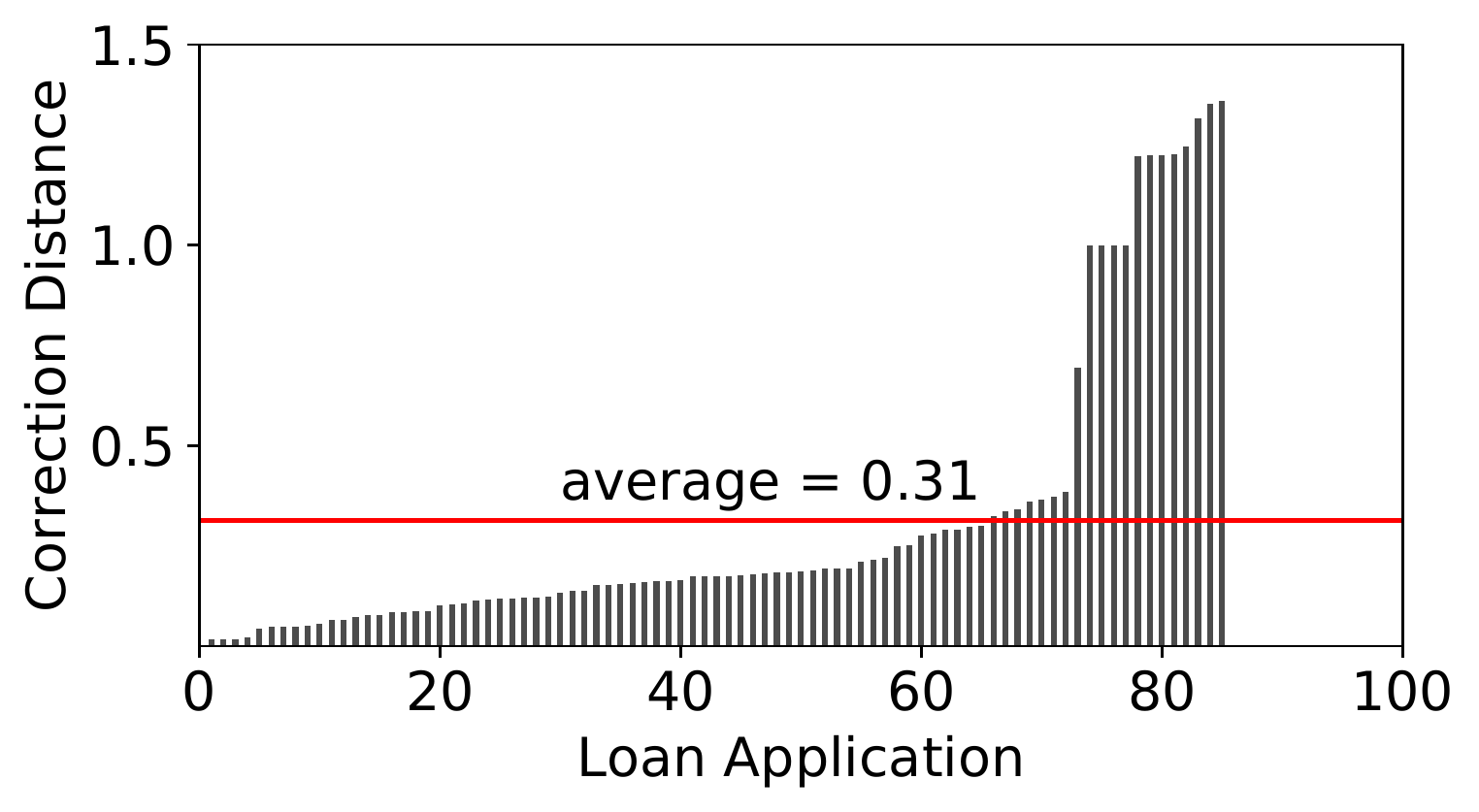}
	\includegraphics[width=0.31\linewidth]{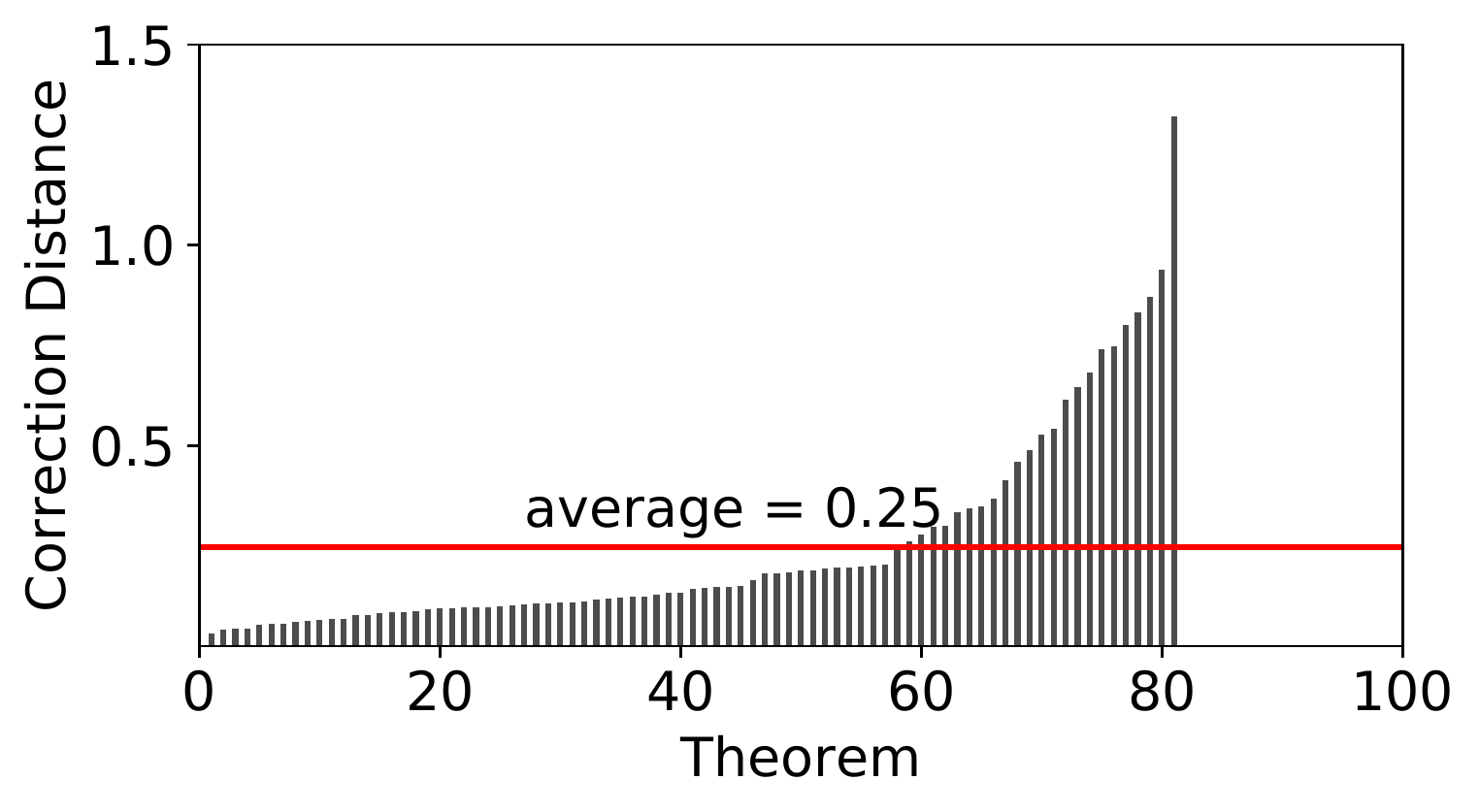}
	\includegraphics[width=0.31\linewidth]{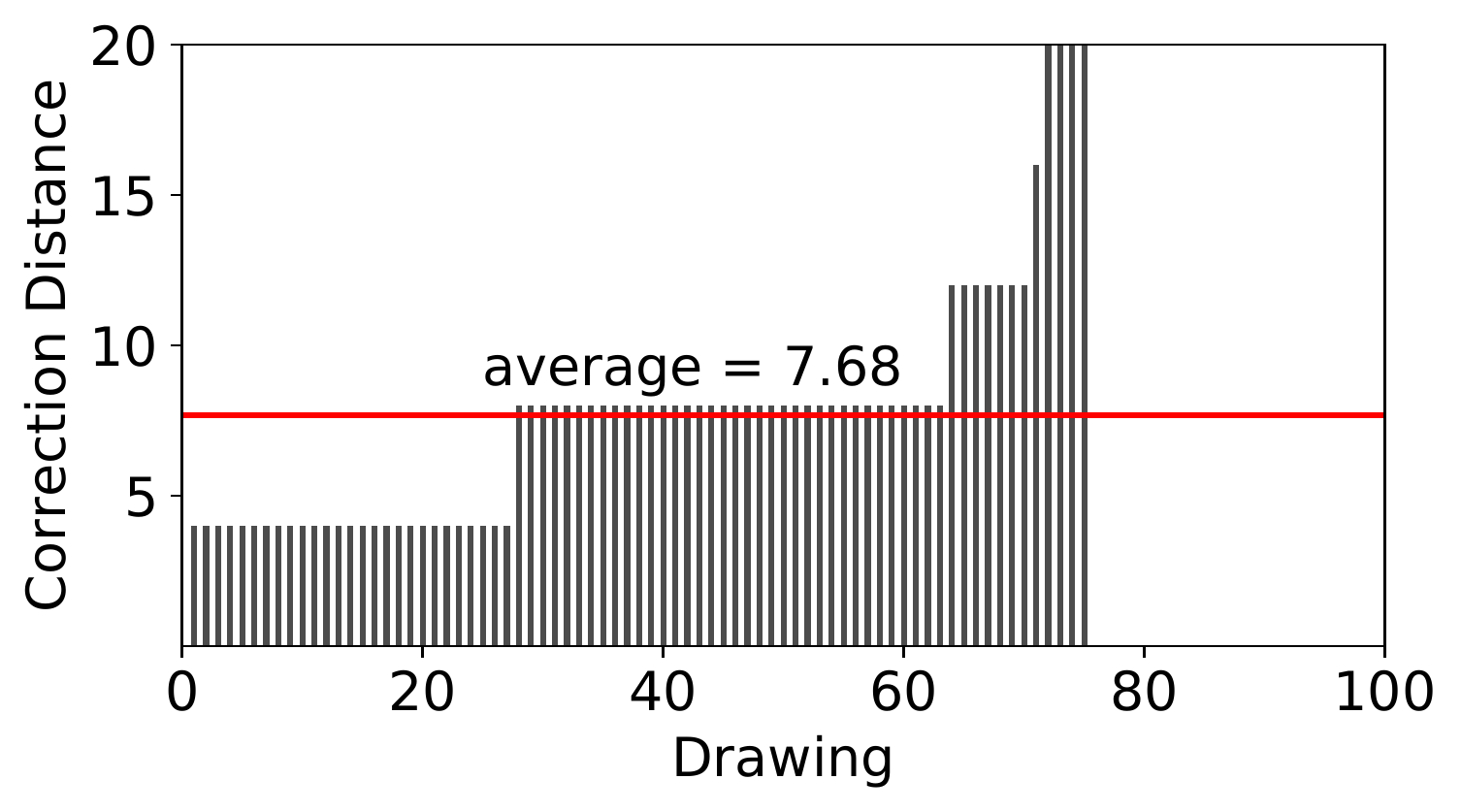} \\
					\vspace{-0.05in}
	{\small	(a) Mortgage Underwriting \qquad\qquad (b) Theorem Proving \qquad\qquad (c) Drawing Recognition}
	\end{center}
		\vspace{-0.15in}
	\caption{Distances of judgment interpretations generated by \tool.}
	\label{fig:distance}
\end{figure}

{\bf Stability and minimality.}
For the selected 100 inputs that are rejected by each network,
\tool successfully generated symbolic corrections for 85 inputs
of mortgage underwriting, 81 inputs of theorem proving, and 75 inputs of drawing recognition.
For the remaining inputs, it is either the case
that the corrections found by \tool were discarded for being unstable,
or the case that \tool failed to find an initial concrete correction
due to the incompleteness of the applied adversarial example generation algorithm. 
These results show that \tool is effective in finding symbolic corrections
that are stable and verified.

We next discuss how similar these corrections are to the original input.
Figure~\ref{fig:distance} lists the sorted distances of the aforementioned
85 symbolic corrections.
For mortgage application and theorem proving, the distance is defined using a weighted $L_1$ norm, where the weight for
each feature is 1/(max-min) (see Appendix \ref{sec:ext-setup}).
The average distances of corrections generated on these two applications are 0.31 and 0.25 respectively.
Briefly, the former would mean, for example, to decrease the DTI by 19.5\% or increase the interest rate by 3\%,
while the latter would mean, for example, to add 25\% more unit clauses or horn clauses.
Moreover, the smallest distances for these two applications are only 0.016 and 0.03.
As for drawing recognition, the distance is measured by the number of features to change (that is, number of added lines $\times$ 4).
As figure~\ref{fig:distance}(c) shows, the sizes of the corrections range from 1 line to 5 lines, with 2 lines being the average. 
In conclusion,
the corrections found by it are often small enough to be actionable
for end users.

To better understand these corrections qualitatively, we inspect several corrections more closely in Appendix~\ref{sec:case-study}.
We also include more example corrections in Appendix~\ref{sec:examples}.

\begin{figure}[t]
	\begin{center}
		\includegraphics[width=0.31\linewidth]{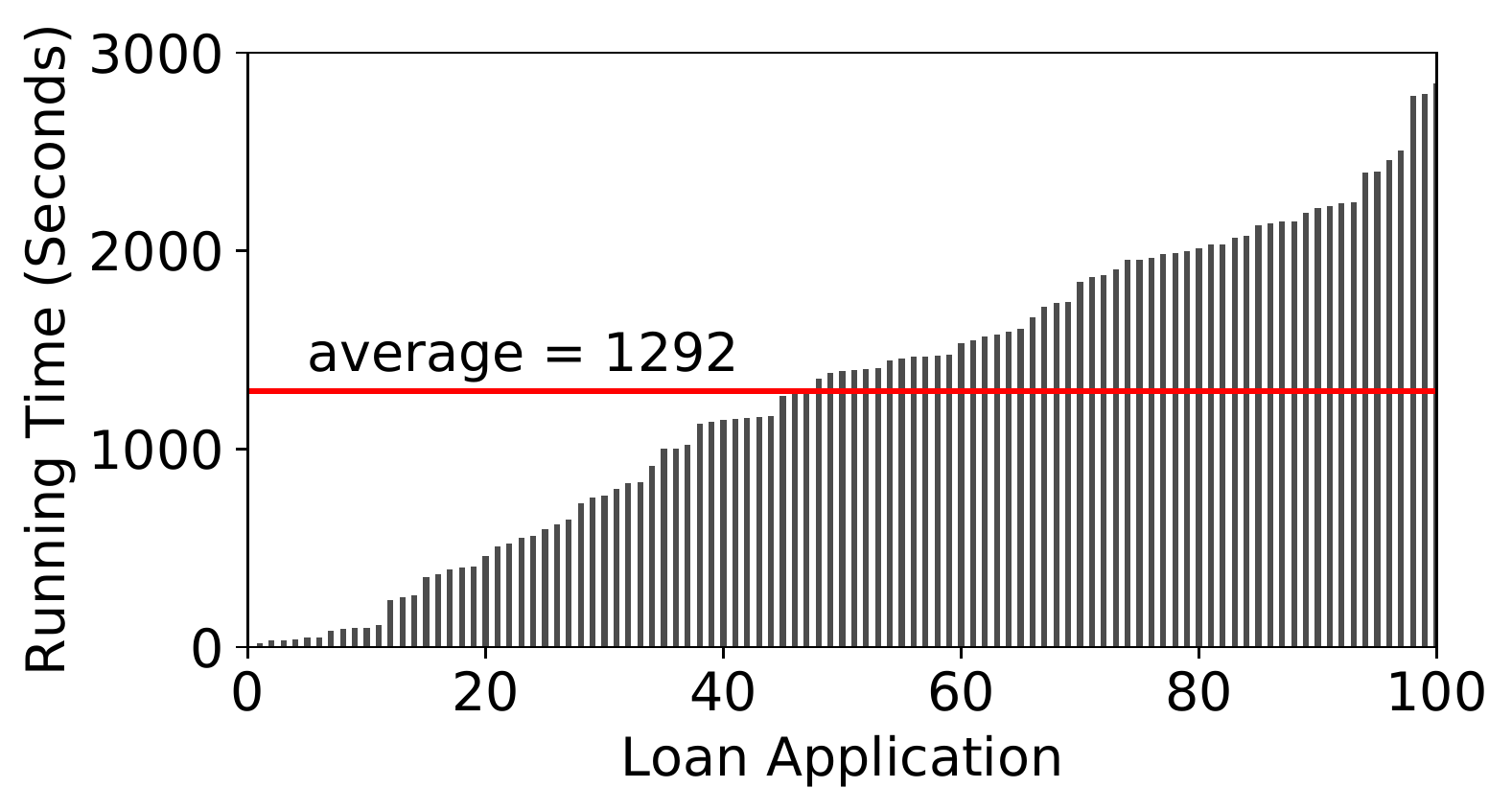}
		\includegraphics[width=0.31\linewidth]{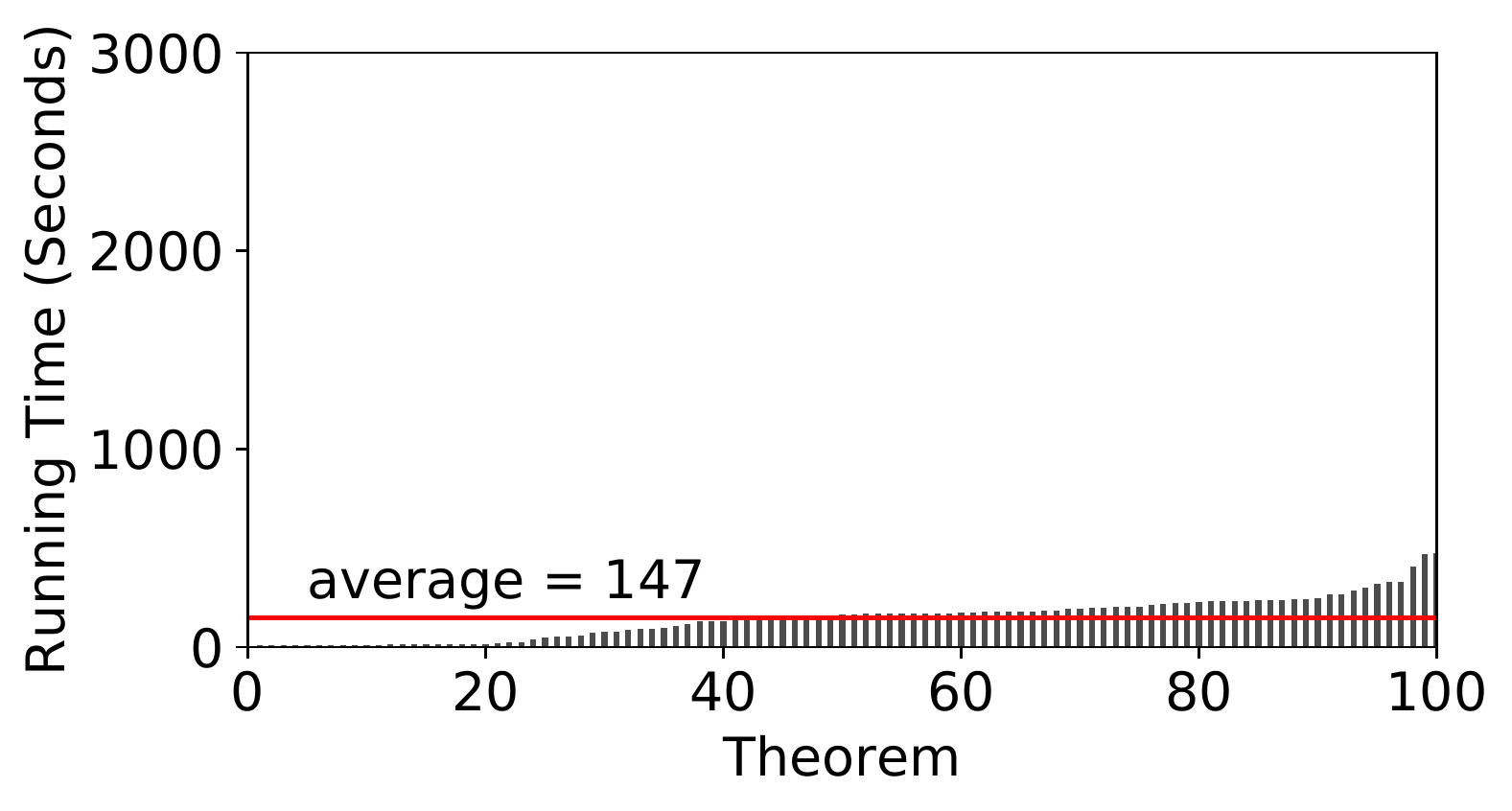}
		\includegraphics[width=0.31\linewidth]{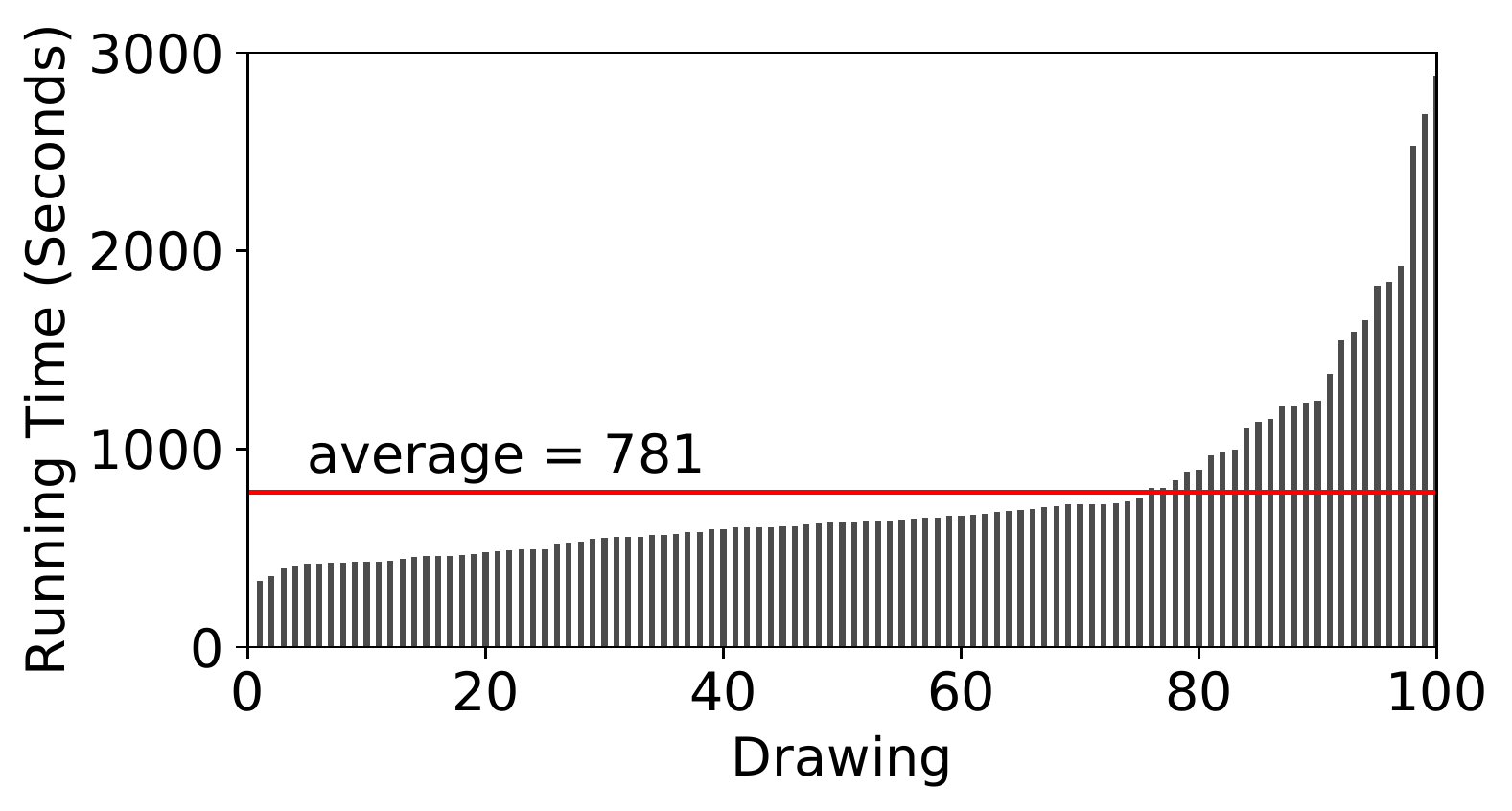} \\
				\vspace{-0.05in}
	{\small	(a) Mortgage Underwriting \qquad\qquad (b) Theorem Proving \qquad\qquad (c) Drawing Recognition}
	\end{center}
	\vspace{-0.15in}
	\caption{Running time of \tool on each input.}
	\label{fig:time}
	\vspace{-0.1in}
\end{figure}
{\bf Efficiency.}
Figure~\ref{fig:time} shows the sorted running time of \tool across all inputs for our three applications.
On average, \tool takes around 20 minutes, 2 minutes, and 13 minutes to generate corrections for each input on the three applications respectively.
We first observe \tool consumes least time on theorem proving.
It is not only because theorem proving has the smallest network but also because the search often terminates much earlier
before reaching the maximum number of regions to consider (m=100).
On the other hand, \tool often reaches this limit on the other two applications.
Although drawing recognition has a larger network than mortgage underwriting, \tool consumes less time on it.
This is because \tool uses a generative network to decide which features to change on drawing recognition, which leads to
one invocation to Algorithm~\ref{alg:search} per input.
On the other hand, \tool needs to invoke Algorithm~\ref{alg:search} for multiple times per input which searches under a combination of
different features.
However, a single invocation to Algorithm~\ref{alg:search} is still faster for mortgage underwriting.

After closer inspection, we find the running time is dominated by invocations to the LP solver. 
We have two observations about the invocation time.
First, most of the time is spent in instance creation rather actual solving due to the poor python
binding of Gurobi.
For instance, in mortgage underwriting, while each instance creation takes around 60ms, the actual
solving typically only takes around 1ms.
As a result, \tool can be made even more efficient if we re-implement it using C++ or if Gurobi improves
the python binding.
Second, the LP solver scales well as the size of the network and the number of dimensions grow.
For example, compared to the solving time (1ms) in mortgage underwriting, where the network comprises
1,000 neurons and the corrections are 2-dimension, the solving time only grows up to around 7ms in drawing recognition,
where the network comprises 4,096 neurons and the corrections are up to 20-dimension.
This indicates that \tool has the potential to scale to even larger networks with higher input dimensions.

\begin{table}
	\caption{Effect of varying the maximum number of regions to consider.}
	\label{tab:vary-m}
	\vspace{-0.15in}
	\begin{center}
		\begin{small}
	\begin{tabular}{cccc}
		\hline
		m & \# explored regions & volume & time (in seconds) \\
		\hline
		100 & 88, 100, 100, 100
		& 2.4, 10.3, 9.2, 1.29
		& 102, 191, 141, 118
		\\
		\hline
		500 & 88, 205, 214, 500
		 &2.4, 26.3, 21.9, 6.9
		 & 100, 374, 288, 517
		  \\
		\hline
			1000 & 88, 205, 214, 1000  &2.4, 26.3, 21.9, 10.2 & 100, 375, 290, 1115 \\
				\hline
					2000 & 88, 205, 214, 1325  &2.4, 26.3, 21.9, 11.2 & 101, 375, 291, 1655 \\
					\hline
	\end{tabular}
	\end{small}
	\end{center}
	\vspace{-0.2in}
\end{table}

{\bf Varying maximum number of regions.}
Table~\ref{tab:vary-m} shows the results of varying maximum number of regions to consider ($m$) for four randomly selected inputs of mortgage underwriting.
To simplify the discussion, we only study corrections generated under DTI and interest rate.
As the table shows, both the volume and running time increase roughly linearly as the number of explored regions grows.


{\bf Comparing to sampling by a grid.}
An alternative approach to generate judgment interpretations is to sample by a grid.
Since there may be unviable inputs between two adjacent viable inputs, a grid with fine granularity is needed to produce a symbolic correction with high confidence. 
However, this is not feasible if there are continuous features or the input dimension is high.
For instance, the corrections generated on drawing recognition may involve up to 20 features.
Even if we only sample 3 pixels along each feature, it would require over 3 billion samples.
Our approach on the other hand, verifies a larger number of concrete corrections at once by verifying a linear region.

%% file: related.tex
\section{Related Work}
\label{sec:related}


Much work on interpretability has gone into analyzing the results produced by a convolutional network that
does image classification.
The Activation Maximization approach and its follow-ups visualize
learnt high-level features by finding inputs that maximize
activations of given neurons~\cite{am, hinton12, lee:icml2009, oord:icml2016, nguyen2016synthesizing}.
Zeiler and Fergus \cite{deconv} uses deconvolution to visualize what a network has learnt.
Not just limited to image domains, more recent works try to build interpretability as part of the network itself~\cite{pinhiro:cvpr2015, lei:emnlp2016, tan:asru2015, wu:is2016, li:aaai18}.
There are also works that try to explain a neural network
by learning a more interpretable model~\cite{lime, hmm, modelExtraction}.
Anchors~\cite{anchors} identifies features that are sufficient to preserve current classification.
As far as we know, the problem definition of judgement interpretation is new, and none
of the existing approaches can directly solve it. Moreover, these approaches typically generate a single input prototype or relevant features, but do not result in corrections or a space of inputs that would lead the prediction to move from an undesirable class to a desirable class.


%% file: conclusion.tex
\section{Conclusion}
\label{sec:conclusion}

We proposed a new approach to interpret a neural network
by generating minimal, stable, and symbolic corrections
that would change its output.
Such an interpretation is a useful way to provide feedback
to a user when the neural network fails to produce a desirable output.
We designed and implemented the first algorithm for generating such corrections, 
and demonstrated its effectiveness on three neural network models from different real-world domains.

%% file: appendix.tex
\newpage
\appendix

\section{Extensions to Our Approach}
\label{sec:ext}

In this section, we discuss several extensions to our approach.
%

\paragraph{Handling categorical features.}
%
Categorical features are typically represented using one-hot encoding and directly applying Algorithm~\ref{alg:search}
on the embedding can result in a symbolic correction comprising invalid concrete corrections.
To address this issue, we enumerate the embeddings representing different values of categorical features and apply Algorithm~\ref{alg:search}
to search symbolic corrections under each of them.
%

\paragraph{Extending for multiple classes. } Our approach can be easily extended for multiple classes as long as there is only one desirable class.
Concretely, we need to: 1) guide the initial concrete correction generation (to the desirable class), which has been studied in the literature of adversarial example generation; 2) extend \funname{classConstraints} so that the desired class gets a higher weight than any other class. That said, the focus of our paper is judgment problems which do binary classifications.

\paragraph{Extending to other norms.}
If we use norms other than $L_1$ to measure the sizes of vectors,
our algorithm largely remains the same, except for 
$\dis_e$, which measures the stability and size of a inferred symbolic correction.
When $L_\infty$ is used, we can still evaluate $\dis_e$ using linear programming.
However, when other norms are applied, evaluating $\dis_e$ would require solving one or more non-linear optimization problems.

\paragraph{Reducing the number of invocations to Algorithm~\ref{alg:search}.} When the input dimension is high, Algorithm~\ref{alg:main} may lead to a large number of invocations to Algorithm~\ref{alg:search} due to the many combinations of different features.
One way to avoid this problem is to use another machine learning model to predict which features would yield a desirable correction,
as we saw in the cat drawing experiment (Section~\ref{sec:experiment}).


\paragraph{Extending to non-ReLU activations.}
Our approach applies without any change as long as the activation functions are continuous and can be approximated using piece-wise linear functions.
For networks whose activations are continuous but cannot be approximated using piece-wise linear functions, we can still apply our algorithm but need
constraints that are more expressive than linear constraints to represent verified regions.
When activations are not continuous, our approach no longer applies as our method of testing whether two regions are connected relies on them being continuous.

\paragraph{Avoiding adversarial corrections.}
Adversarial inputs are inputs generated from an existing input via small
perturbations such that they are indistinguishable to end users from the original input
but lead to different classifications.
Adversarial inputs are undesirable and often considered as ``bugs'' of a neural network.
For simplicity, we did not consider them in previous discussions.
To avoid corrections that would result in adversarial inputs,
we rely on the end user to define a threshold $\sigma$ such that any concrete correction
$\delta$ where $\| \delta \| > \sigma$ is considered not adversarial.
Then we add $\| \av{x} \| > \sigma$ as an additional constraint to each
region.

\section{Experiment Details}
\label{sec:ext-exp}

\begin{table}[t!]
	\caption{Mutable features in corrections to mortgage underwriting.}
	\label{tab:features_mortgage}
	\centering
	\begin{small}
		\begin{tabular}{|@{\,}l@{\,}|@{\,}l@{\,} |@{\,} l@{\,} |@{\,}l@{\,}|@{\,}l@{}|}
			\hline
			& {\bf{Name}} & {\bf{Type}} & {\bf{Range}} & {\bf{Radius}}
			\\
			\hline
			1 & Interest Rate & Real & [0, 0.1] & 0.005 \\
			\hline
			2 & Credit Score  & Integer & [300, 850] & 25 \\	
			\hline
			3 & Debt-to-Income & Real & [0.01, 0.64] & 0.025 \\
			\hline
			4 & Loan-to-Value & Real & [0, 2] & 0.025\\
			\hline
			5 & Property Type & Category & 
			\begin{footnotesize}
				\begin{tabular}{@{}l@{}}
					[Cooperative Share, \\
					\ Manufactured Home, \\
					\ Planned Urban \\
					\ Development, \\
					\ Single-Family Home,\\
					\ Condominium] 
				\end{tabular}
			\end{footnotesize}
			& N.A.
			\\
			\hline
		\end{tabular}
	\end{small}
\end{table}

\begin{table}[t!]
	\caption{Mutable features in corrections to theorem proving.}
	\label{tab:features_proof}
	\centering
	\begin{small}
		\begin{tabular}{|@{\,}l@{\,}|@{\,}l@{\,} |@{\,} l@{\,} |@{\,}l@{\,}|@{\,}l@{}|}
			\hline
			& {\bf{Name}} & {\bf{Type}} & {\bf{Range}} & {\bf{Radius}}
			\\
			\hline
			1 & \% Clauses That Are Unit & Real & [0, 1] & 0.025 \\
			\hline
			2 & \%  Clauses That Are Horn & Real & [0, 1] & 0.025 \\	
			\hline
			3 & \% Clauses That Are Ground & Real & [0, 1] & 0.025 \\
			\hline
			4 & Avg. Clause Length & Real & [0, 10] & 0.25\\
			\hline
			5 & Avg. Clause Depth & Real & [0, 10] & 0.25\\
			\hline
		\end{tabular}
	\end{small}
\end{table}

\subsection{Experiment Setup}
\label{sec:ext-setup}

We describe details about the experiment setup in this subsection.

\paragraph{Mutable features.} Table~\ref{tab:features_mortgage} and Table~\ref{tab:features_proof} describe the features that are allowed to change in order to
generate symbolic corrections for mortgage underwriting and theorem proving.

\paragraph{Stability and distance metrics.} We first describe the operator $\funname{dist}_e$ for the mortgage application, which measures both
stability and distance.
Briefly, we used a weighted $L_1$ norm to evaluate the distance
of the correction and a weighted $L_\infty$ norm to evaluate the stability.
For distance, we use 1 / (max - min) as the weight for each numeric feature.
As for the categorical feature ``property type'', we charge 1 on the distance if the minimum stable concrete correction
in the symbolic correction (the minimum stable region center) would change
it, or 0 otherwise.
This is a relatively large penalty as changing the property type requires the applicant to switch to a different property.
For stability, we define a stability radius array $\av{r}$ and use
$1 / \av{r}[i]$ as the weight for feature $i$.
If the category feature is involved, we require the symbolic corrections to at least contain
two categories of the feature.
Table~\ref{tab:features_mortgage} defines the range and radius of each feature.
%
We define  $\funname{dist}_e$ as follow:
\begin{small}
	\begin{align*}
	\dis_e(\av{\Delta}) := \underset{{\av{\delta} \in S}}{\funname{min}}( \frac{|\av{\delta}[1]|}{0.1-0}
	+ \frac{|\av{\delta}[2]|}{850-300} + \frac{|\av{\delta}[3]|}{0.64-0.01} \\
	+\frac{|\av{\delta}[4]|}{2-0} + (0 \text{ if } \av{\delta}[5] \t{ leads to no change} \text{ else } 1)),
	\end{align*}
\end{small}
where
\begin{small}
	\begin{align*}
	S := \{ \av{\delta} \in \av{\Delta} \mid &
	\exists 1 \leq i < j \leq 4.
	\forall \av{\delta}'. | \av{\delta}'[i] - \av{\delta}[i] | \leq e*r[i] \\
	& \qquad \qquad \qquad \quad \ \ \wedge | \av{\delta}'[j] - \av{\delta}[j] | \leq e*r[j] \\
	& \qquad \qquad \qquad  \quad \ \ \wedge |\av{\delta}'[k] = \av{\delta}[k]| \text{ for } k\notin \{i,j\}\\
	& \qquad \qquad \implies \av{\delta}' \in \av{\Delta} \} \\
	%
	\cup\ \{ \av{\delta} \in \av{\Delta} \mid &
	\exists i \in [1,4] \t{ and a category } c \t{ of Feature 5} \\ 
	& \t{that differs from the category } \av{\delta}[5] \t{ leads to so that} \\
	& \quad\quad \forall \av{\delta}'. | \av{\delta}'[i] - \av{\delta}[i] | \leq e*r[i] \\
	& \ \ \ \qquad\wedge \av{\delta}'[5] = \av{\delta}[5] \t{ or $\av{\delta}'[5]$ leads to $c$} \\
	& \ \ \ \qquad\wedge |\av{\delta}'[k] = \av{\delta}[k]| \text{ for } k\notin \{i,5\} \\
	& \qquad \qquad \implies \av{\delta}' \in \av{\Delta} \}.
	\end{align*}
\end{small}
%
%
%
Note when the categorical feature property type is involved, we evaluate $\dis_e(\av{\Delta})$ by solving a sequence of integer linear
programming problems, which is also implemented using Gurobi.

The definition of $\funname{dist}_e$ for theorem proving is similar except that all mutable features are real values:
\begin{small}
	\begin{align*}
	\dis_e(\av{\Delta}) := \underset{{\av{\delta} \in S}}{\funname{min}}( \frac{|\av{\delta}[1]|}{1-0}
	+ \frac{|\av{\delta}[2]|}{1-0} + \frac{|\av{\delta}[3]|}{1-0} 
	+\frac{|\av{\delta}[4]|}{10-0} + \frac{|\av{\delta}[5]|}{10-0},
	\end{align*}
\end{small}
where
\begin{small}
	\begin{align*}
	S := \{ \av{\delta} \in \av{\Delta} \mid &
	\exists 1 \leq i < j \leq 5.
	\forall \av{\delta}'. | \av{\delta}'[i] - \av{\delta}[i] | \leq e*r[i] \\
	& \qquad \qquad \qquad \quad \ \ \wedge | \av{\delta}'[j] - \av{\delta}[j] | \leq e*r[j] \\
	& \qquad \qquad \qquad  \quad \ \ \wedge |\av{\delta}'[k] = \av{\delta}[k]| \text{ for } k\notin \{i,j\}\\
	& \qquad \qquad \implies \av{\delta}' \in \av{\Delta} \}. 
	\end{align*}
\end{small}
We set $e = 1$ for both applications in all runs in the experiment.

\subsection{Case Study}
\label{sec:case-study}

\begin{figure}[t!]
	

\begin{tabular}{@{}c@{}c@{}}
		\includegraphics[width=0.5\linewidth]{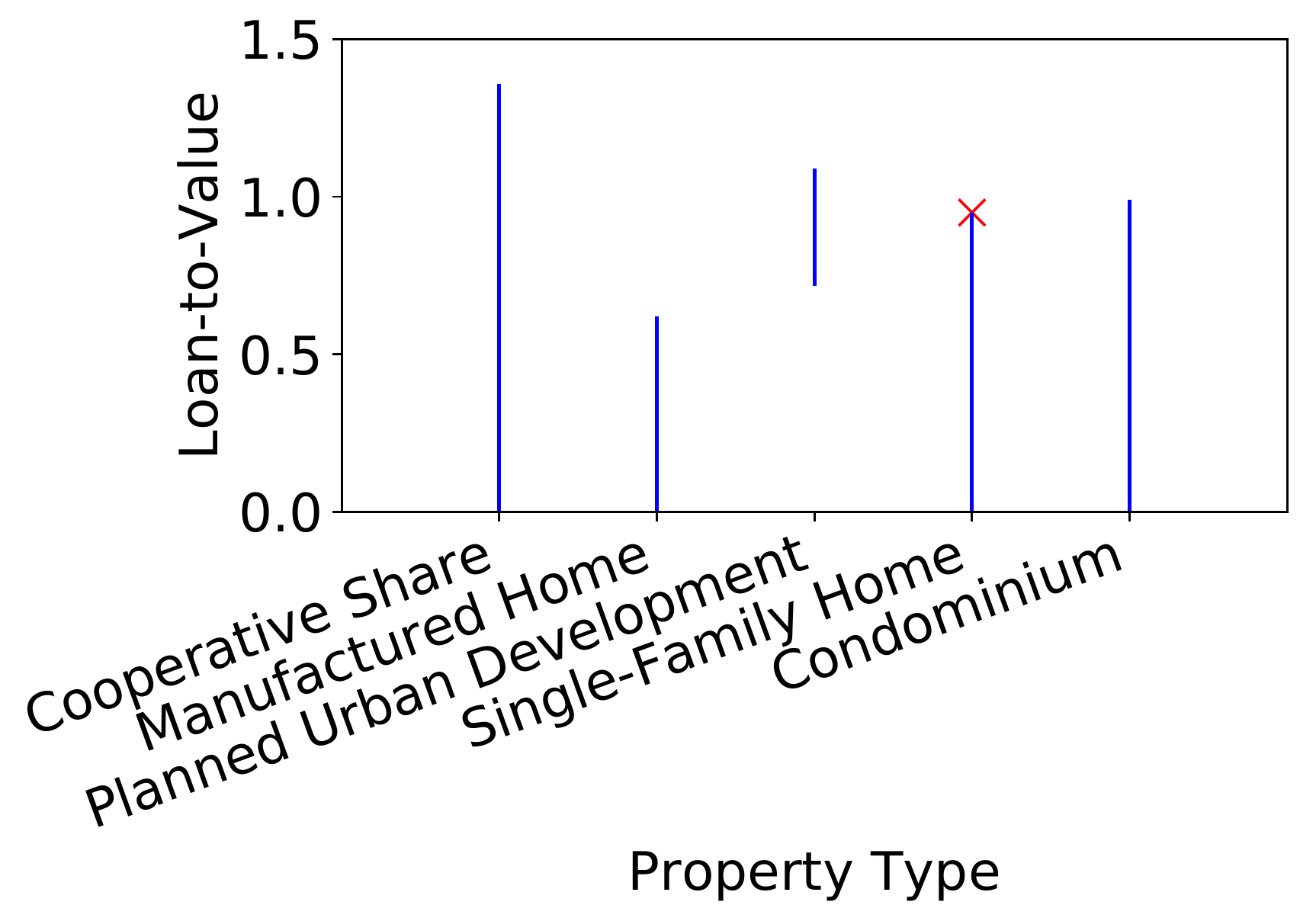} &
	\raisebox{.3\height}{ 
					\includegraphics[width=0.5\linewidth]{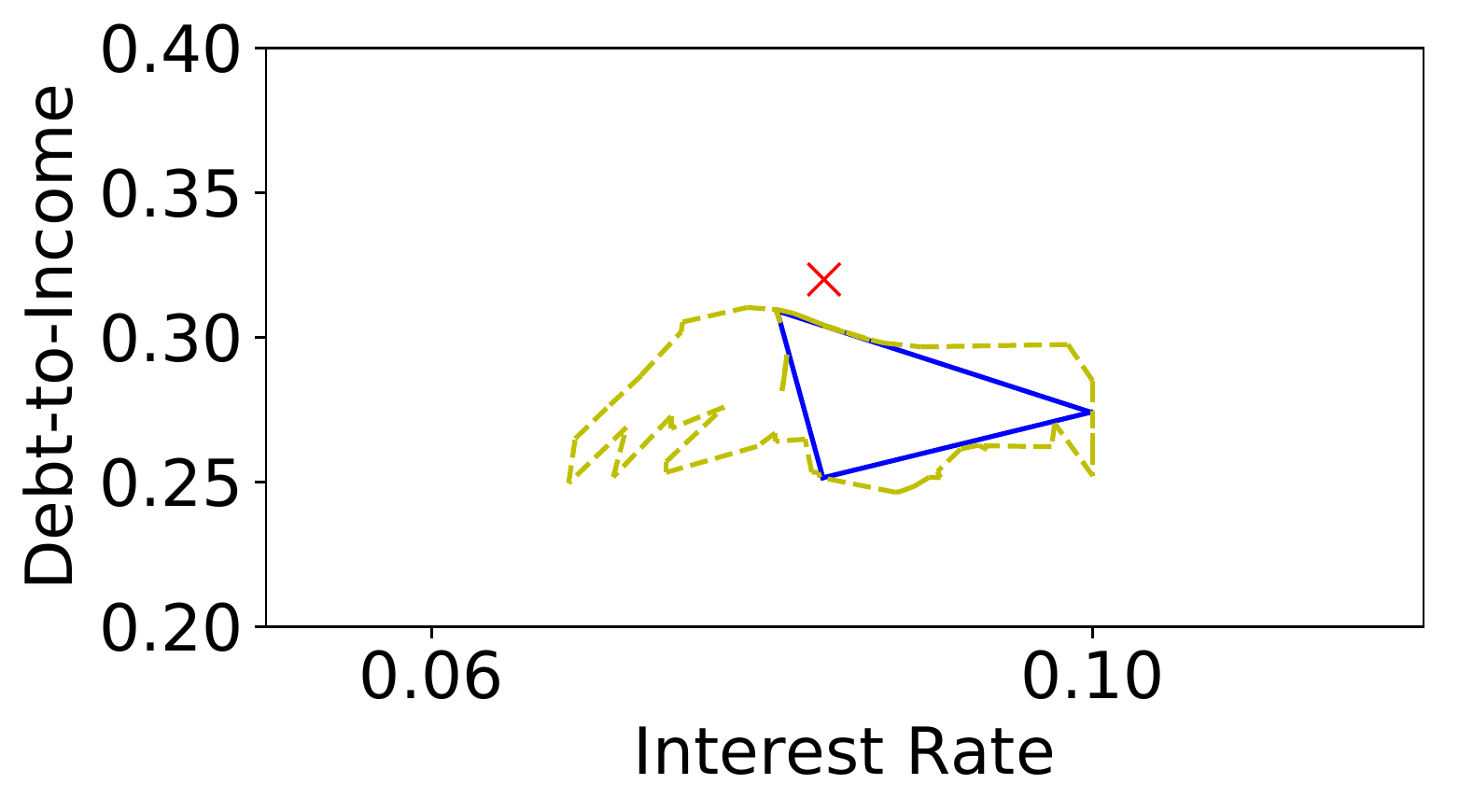}} \\
					(a) & (b) \\
						\raisebox{.3\height}{
					\includegraphics[width=0.5\linewidth]{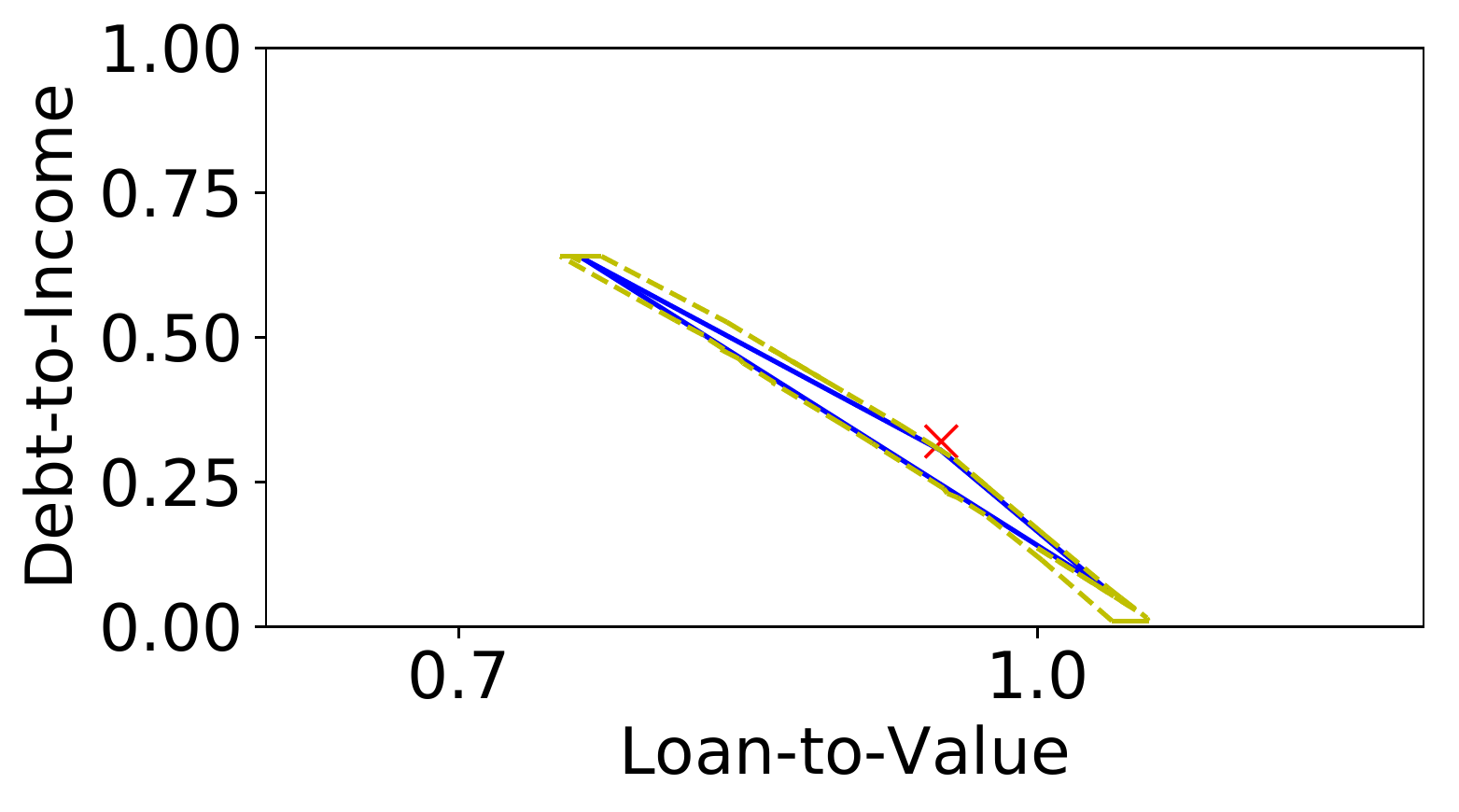}} 
					& 
					\includegraphics[width=0.5\linewidth]{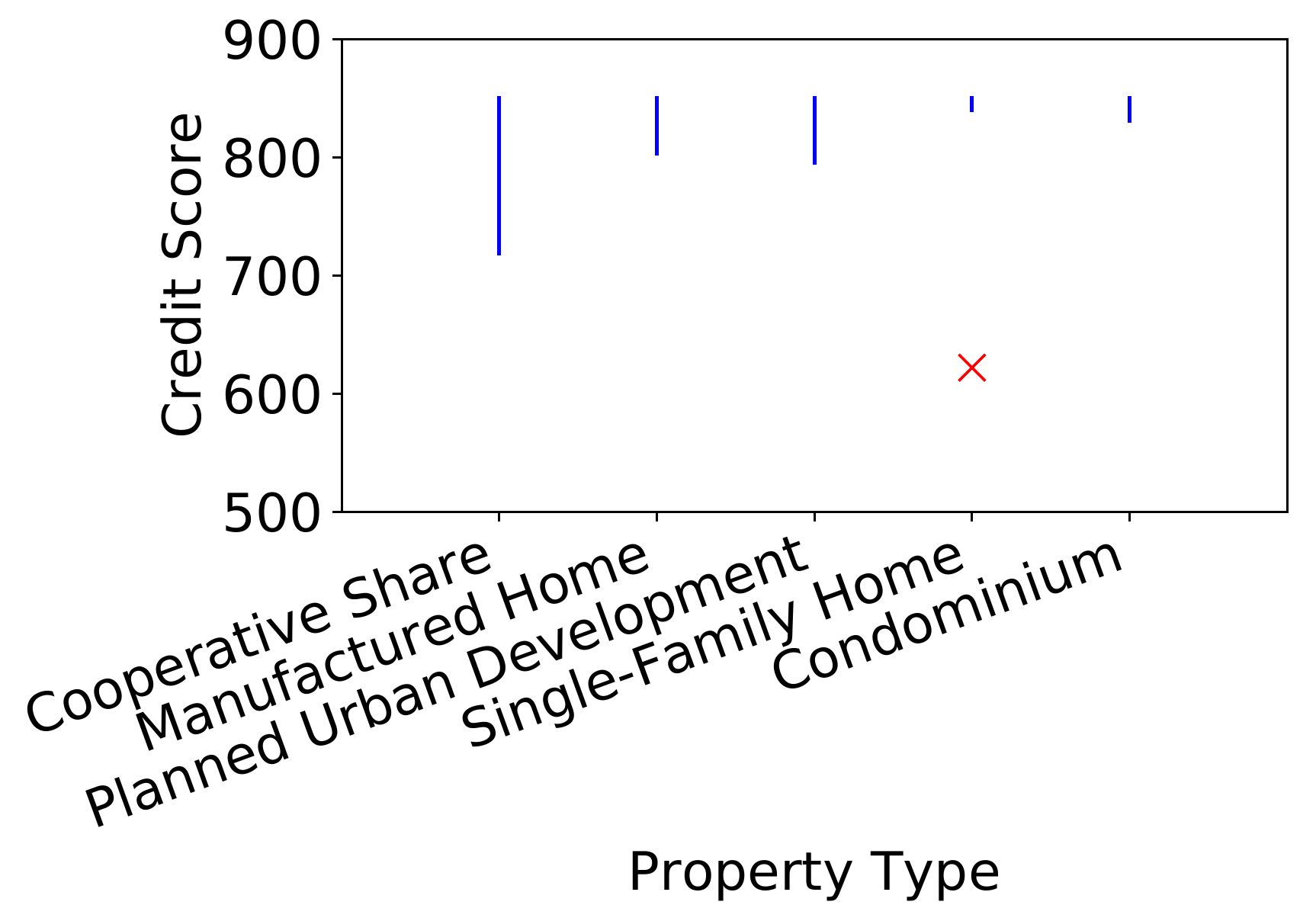} \\
					(c) & (d)
\end{tabular}
	
	\caption{Corrections for the mortgage application with the minimum judgment interpretation (a,b, and c) and a correction for the mortgage application with the maximum judgment interpretation (d).}
	\label{fig:min-point}
\end{figure}

While the discussion in Section~\ref{sec:exp.results} gives a high-level idea of the effectiveness of our approach,
we now look at individual generated symbolic corrections closely.
We are interested in answering two questions:
\begin{enumerate}[topsep=0pt,noitemsep]
	\item Are these corrections small and stable enough such that they are actionable to the applicant?
	\item Do they make sense?
\end{enumerate}

We study corrections generated for mortgage underwriting in detail to answer these two questions. 
More concretely, we inspect the symbolic corrections generated for the application with the minimum correction
among all applications and the ones generated for the application with maximum correction.
These two applications correspond to the rightmost and the leftmost bars on Figure~\ref{fig:distance}(a).

Figure~\ref{fig:min-point}(a)-(c) shows the symbolic corrections generated for the application with the minimum judgment
interpretation among all applications.
The application corresponds to the leftmost bar in Figure~\ref{fig:distance}(a).
Since \tool is configured to generate corrections involving two features out of five features,
there are ten possible corrections that vary different features.
For space reason, we study three of them.

Figure~\ref{fig:min-point}(a) shows the symbolic correction generated along loan-to-value ratio and
property type, which is the minimum correction for this application.
The red cross shows the projection of the original application on these two features,
while the blue lines represent the set of corrected applications that the symbolic correction
would lead to.
First, we observe that the correction is very small.
The applicant will get their loan approved if they reduce the loan-to-value ratio only by 
0.0076.
Such a correction is also stable.
If the applicant decides to stick to single-family home properties, they will get the loan approved as
long as the reduction on the loan-to-value ration is greater than 0.0076.
Moreover, they will get similar results if they switch to cooperative share properties or condominiums.
This correction also makes much sense, since reducing loan-to-value ratio often means to reduce the loan amount.
In practice, smaller loans are easier to approve.
Also, from the perspective of the training data, smaller loans are less likely to default.

Figure~\ref{fig:min-point}(b) shows the symbolic correction generated along debt-to-income ratio and interest rate,
which are two numeric features.
Similar to Figure~\ref{fig:min-point}(a), the red cross represents the projection of the original application,
while the blue triangle represents the symbolic correction.
In addition, we use a polytope enclosed in dotted yellow lines to represent the verified linear regions
collected by Algorithm~\ref{alg:search}.
We have two observations about the regions.
First, the polytope is highly irregular, which reflects the highly nonlinear nature of the neural network.
However, \tool is still able to generate symbolic corrections efficiently.
Secondly, the final correction inferred by our approach covers most area of the regions, which shows 
the effectiveness of our greedy algorithm applied in \funname{inferConvexCorrection}.
While this correction is also small and stable, its distance is larger than the previous correction
along loan-to-value ratio and property type.
Such a correction also makes sense from the training data perspective.
It is obvious that applicants with smaller debt-to-income ratios will less likely default.
As for interest rate, the correction leans towards increasing it.
It might be due to the fact that during subprime mortgage crisis (2007-2009), loans
were approved with irrationally low interest rate, many of which went into default later.

Figure~\ref{fig:min-point}(c) shows the correction generated along debt-to-income ratio and loan-to-value ratio.
Compared to the previous corrections, its distance is small but it is highly unstable (the triangle is very narrow).
In fact, it is discarded by \tool due to this. 

As a comparison to corrections generated on the previous application, Figure~\ref{fig:min-point}(d) shows
the final correction generated on the application that corresponds to the rightmost bar on Figure~\ref{fig:distance}(a).
In other words, its final correction has the largest distance among final corrections generated for all applications.
As the figure shows, such a large distance makes it hard for the applicant to adopt.
For most categories of property type, the applicant needs to raise their credit score by 100, and even to over 800 under some cases,
which is not very easy in practice.
As a result, \tool assigns a high distance for such a correction.

\subsection{More Example Corrections}
\label{sec:examples}

\subsubsection{Mortgage Underwriting}
The red cross shows the projection of the original application on these two features,
while the blue lines represent the set of corrected applications that the symbolic correction
would lead to.
In addition, we use a polytope enclosed in dotted yellow lines to represent the verified linear regions
collected by Algorithm~\ref{alg:search}.

\begin{figure}[h]
\includegraphics[width=0.5\textwidth]{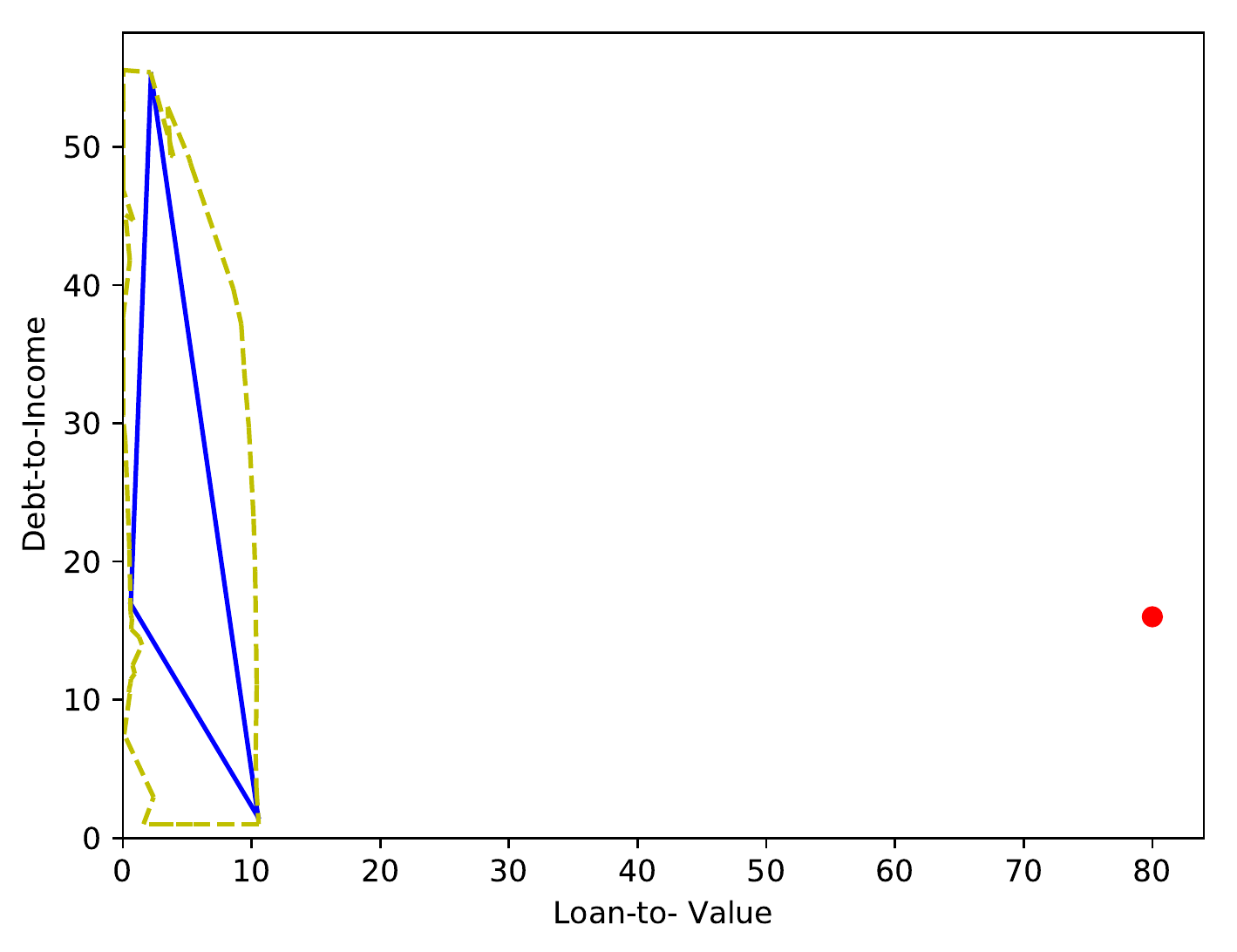}
\includegraphics[width=0.5\textwidth]{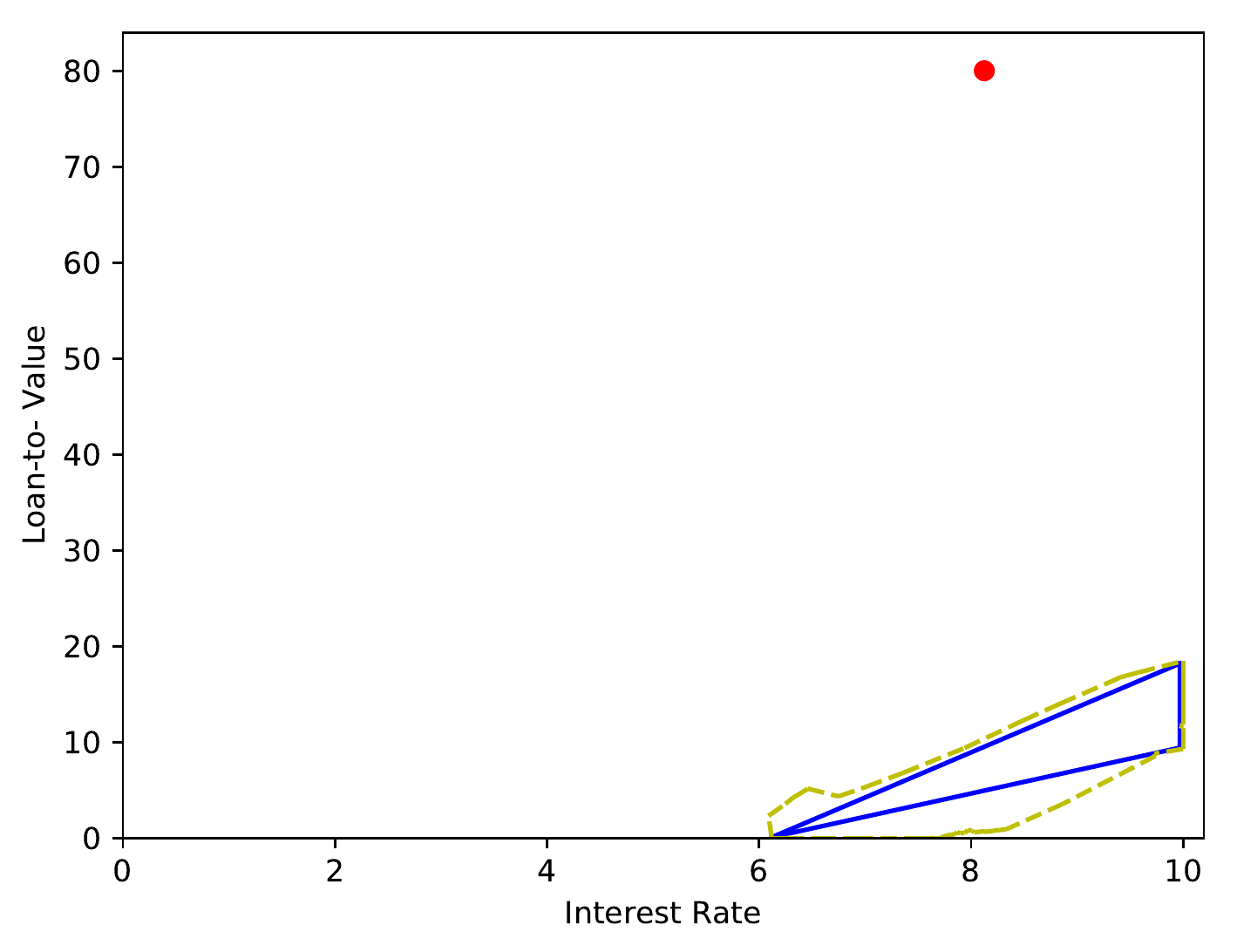}
\end{figure}

\begin{figure}[t]
\includegraphics[width=0.5\textwidth]{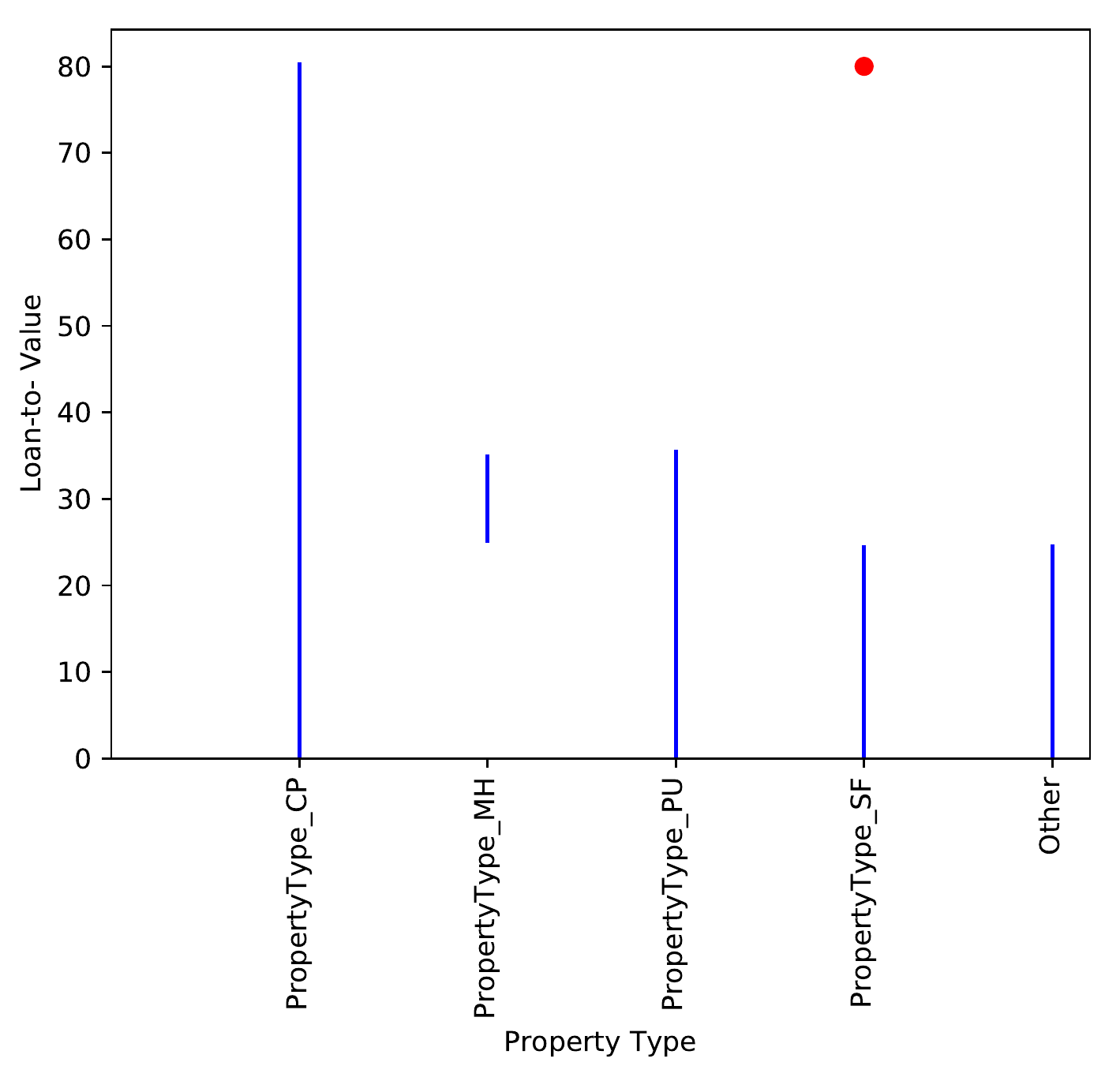}
\includegraphics[width=0.5\textwidth]{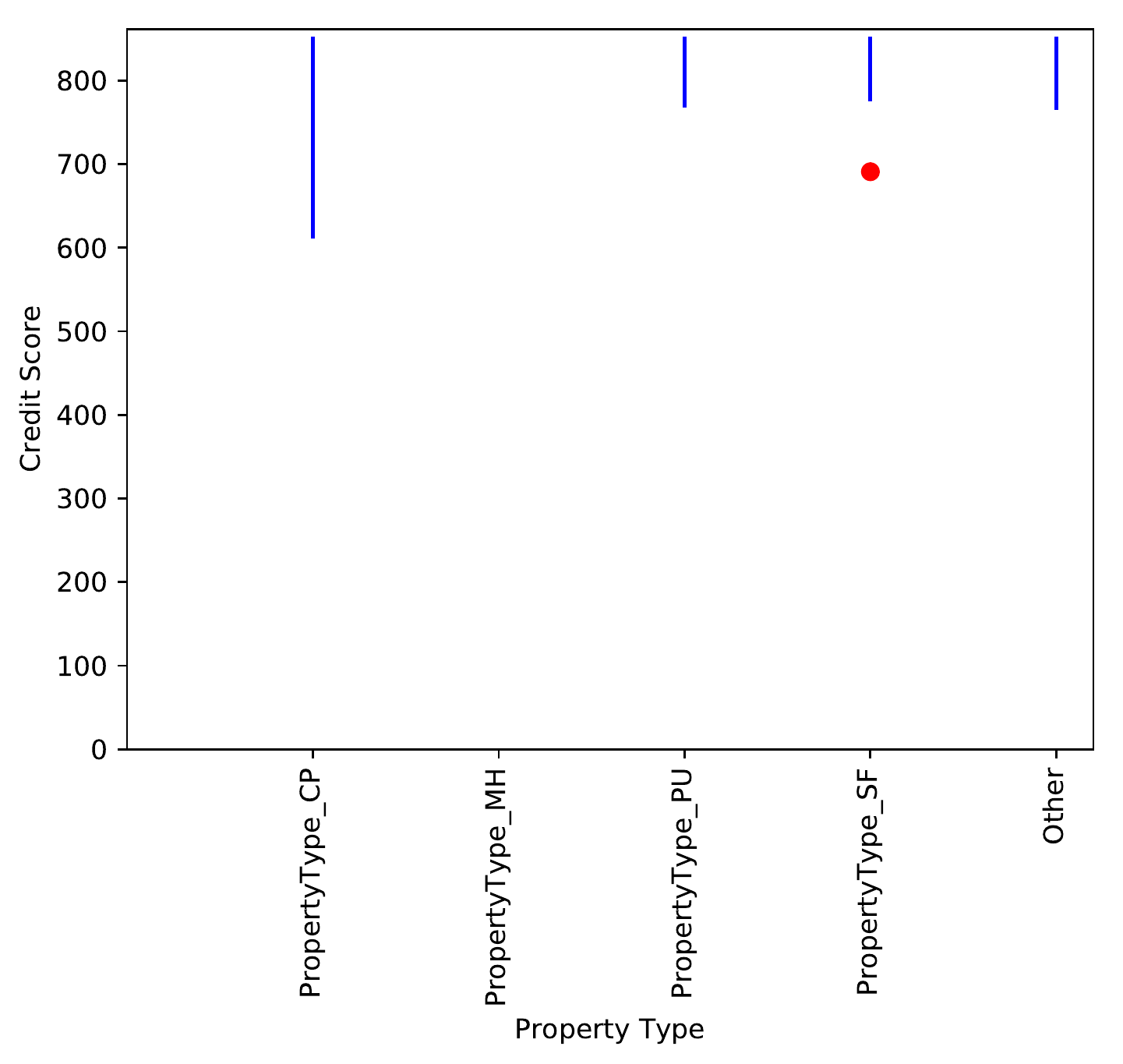} \\
\includegraphics[width=0.5\textwidth]{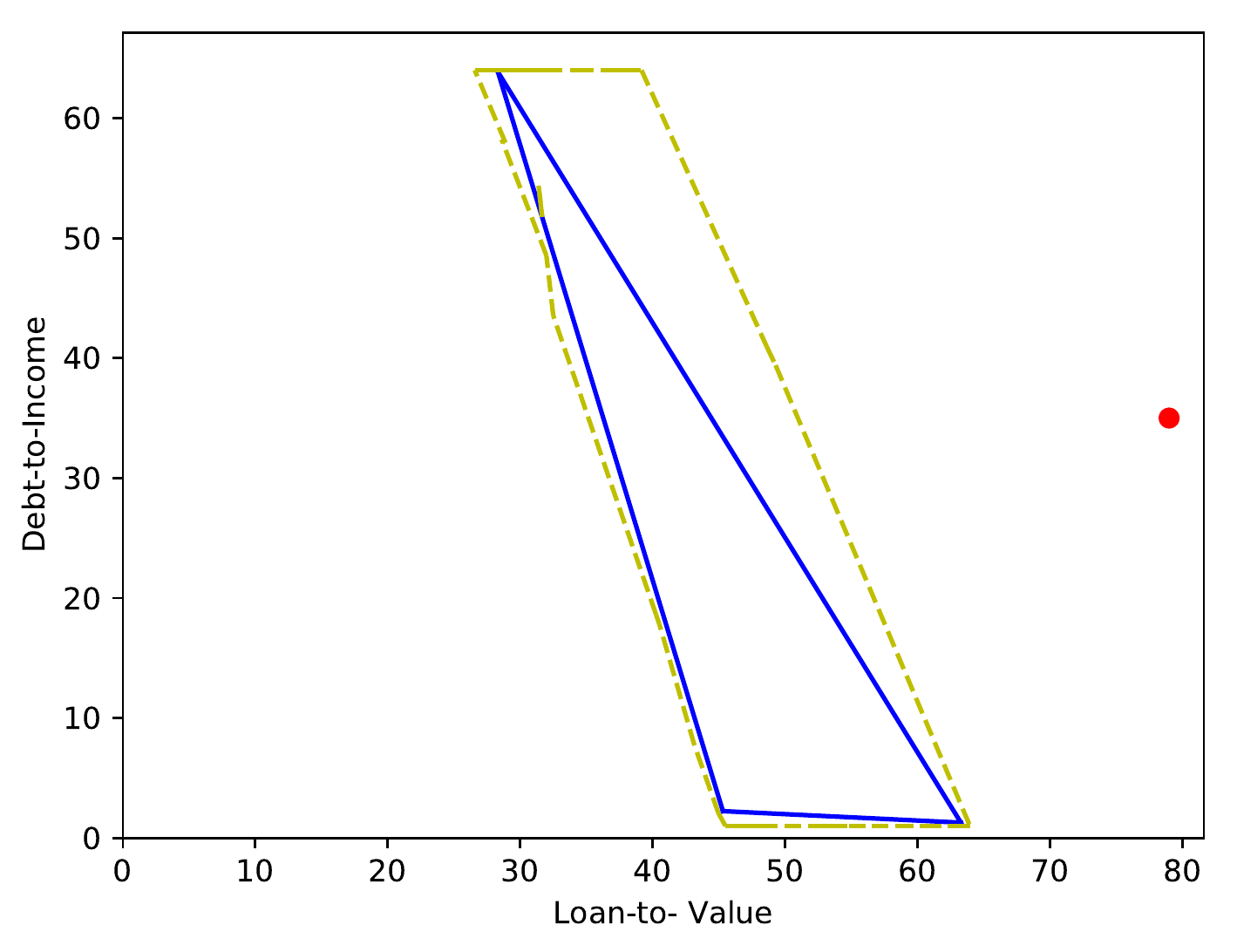}
\includegraphics[width=0.5\textwidth]{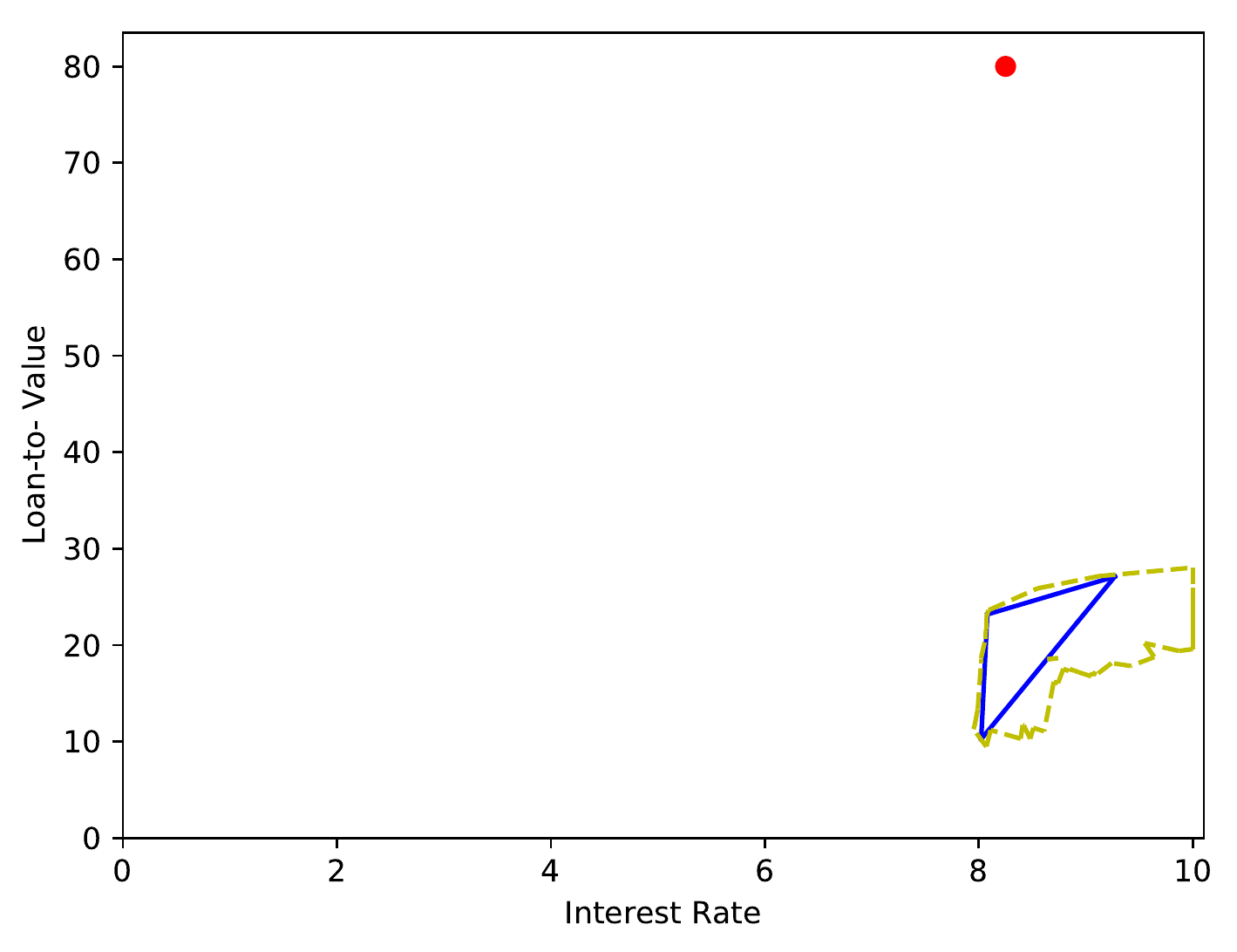} \\
\includegraphics[width=0.5\textwidth]{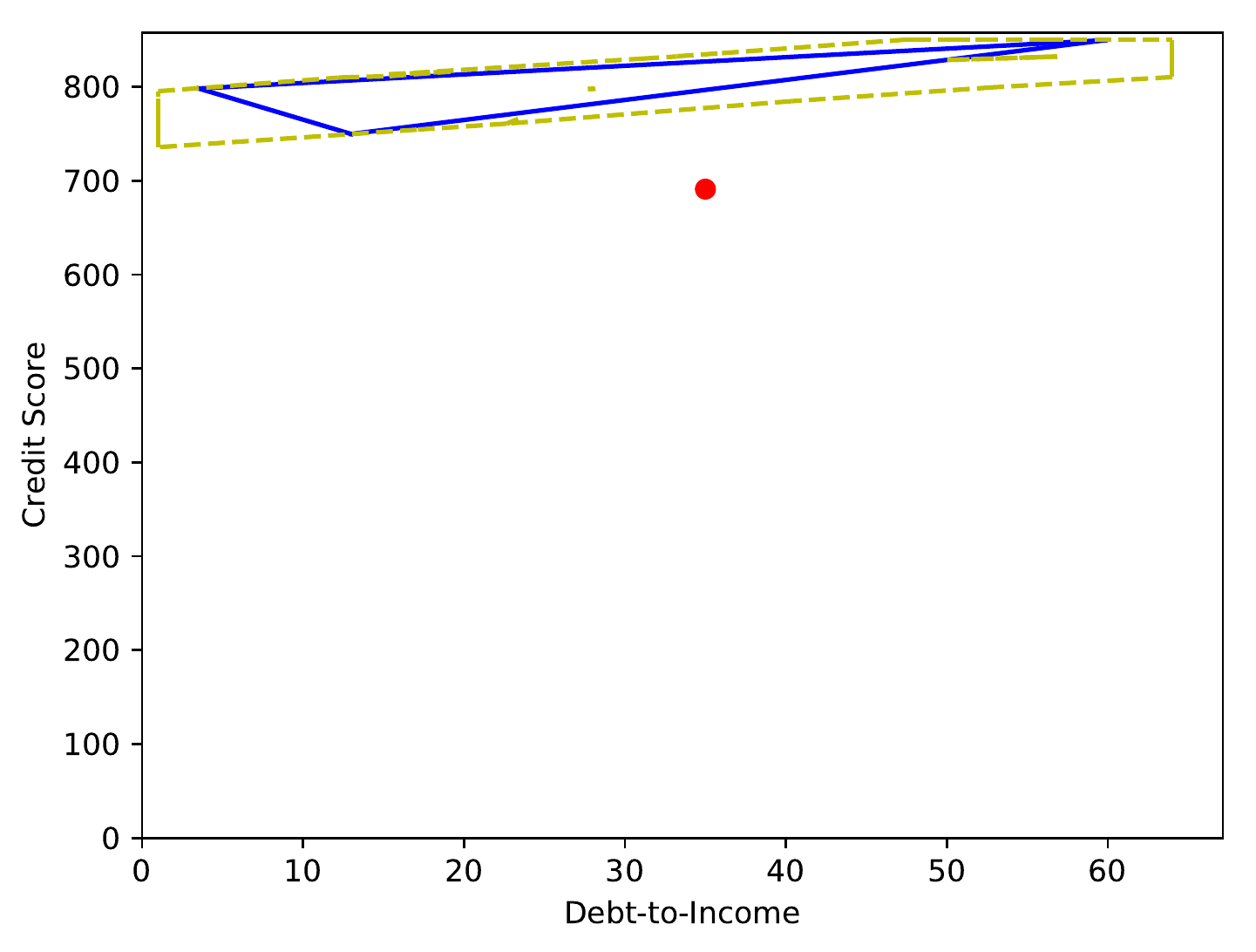}
\includegraphics[width=0.5\textwidth]{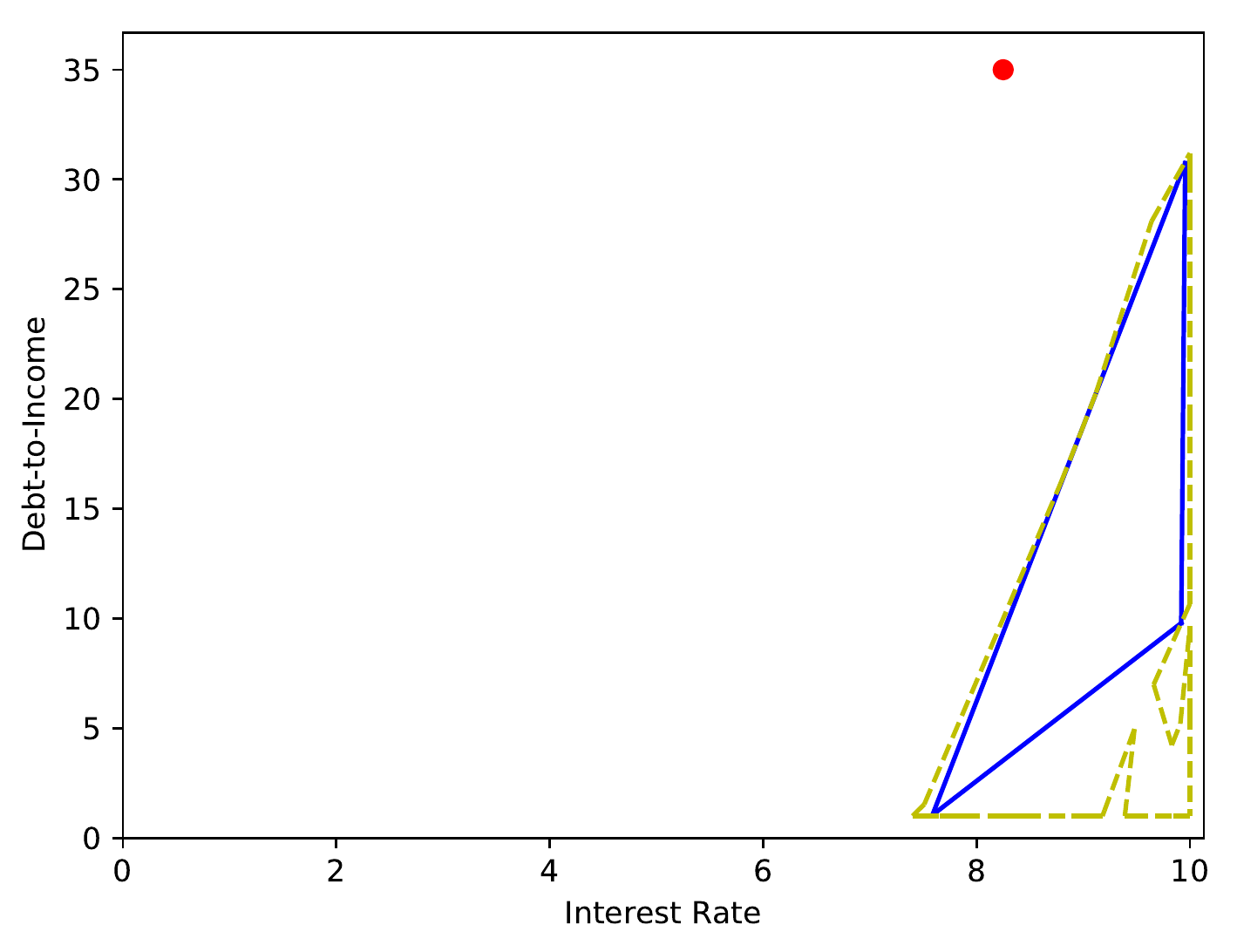} 
\end{figure}

\begin{figure}[t]
\includegraphics[width=0.5\textwidth]{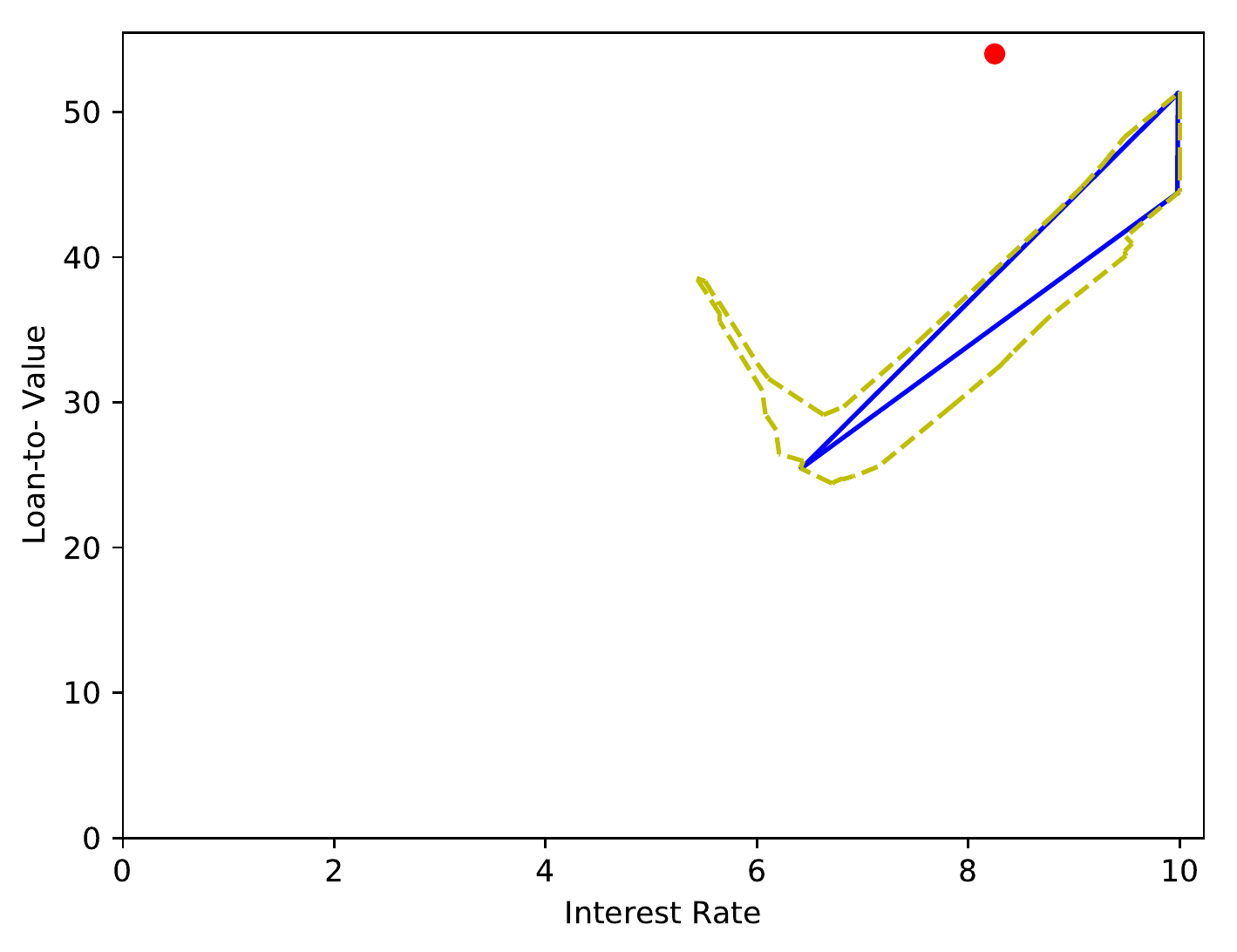}
\includegraphics[width=0.5\textwidth]{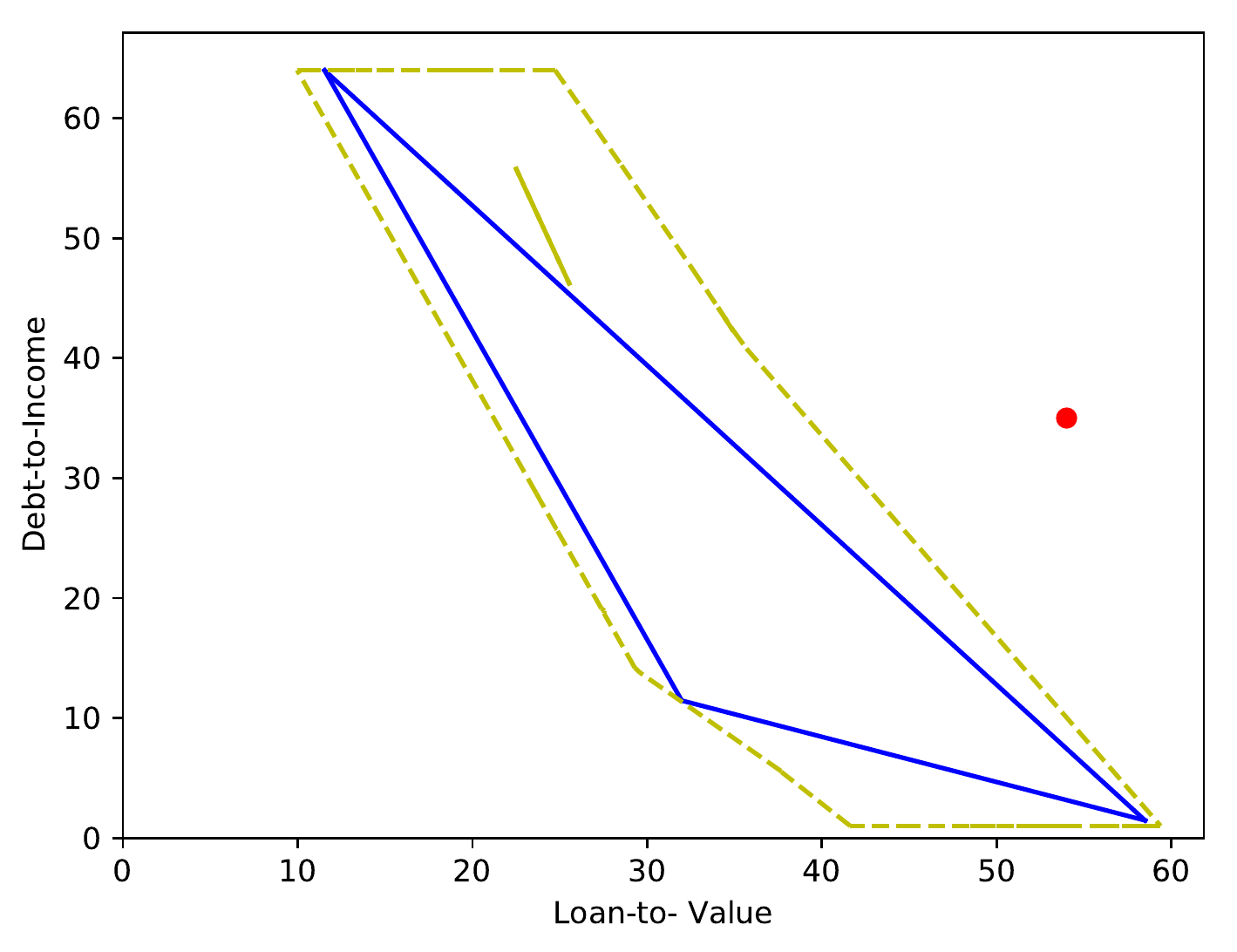} \\
\includegraphics[width=0.5\textwidth]{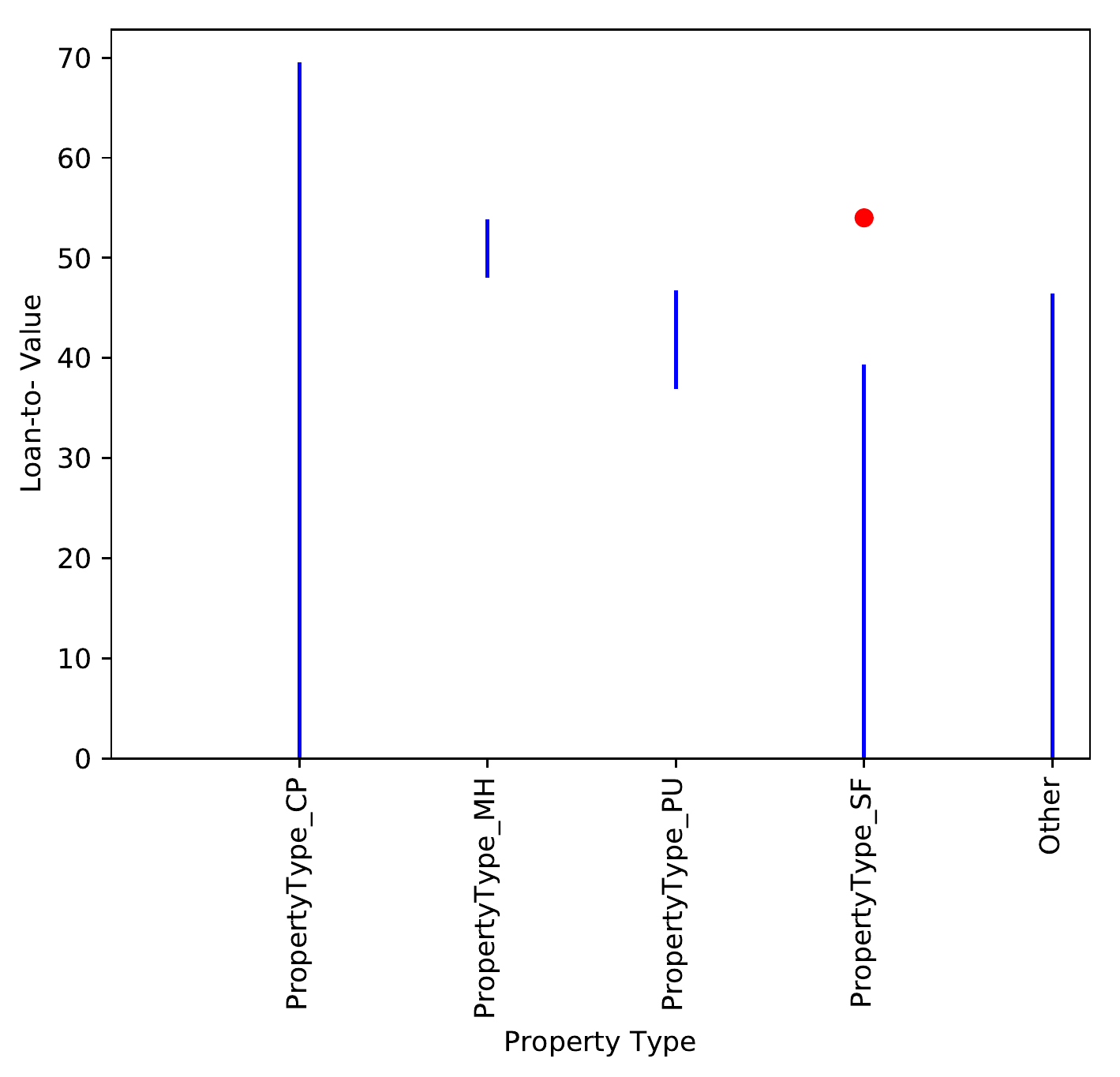}
\includegraphics[width=0.5\textwidth]{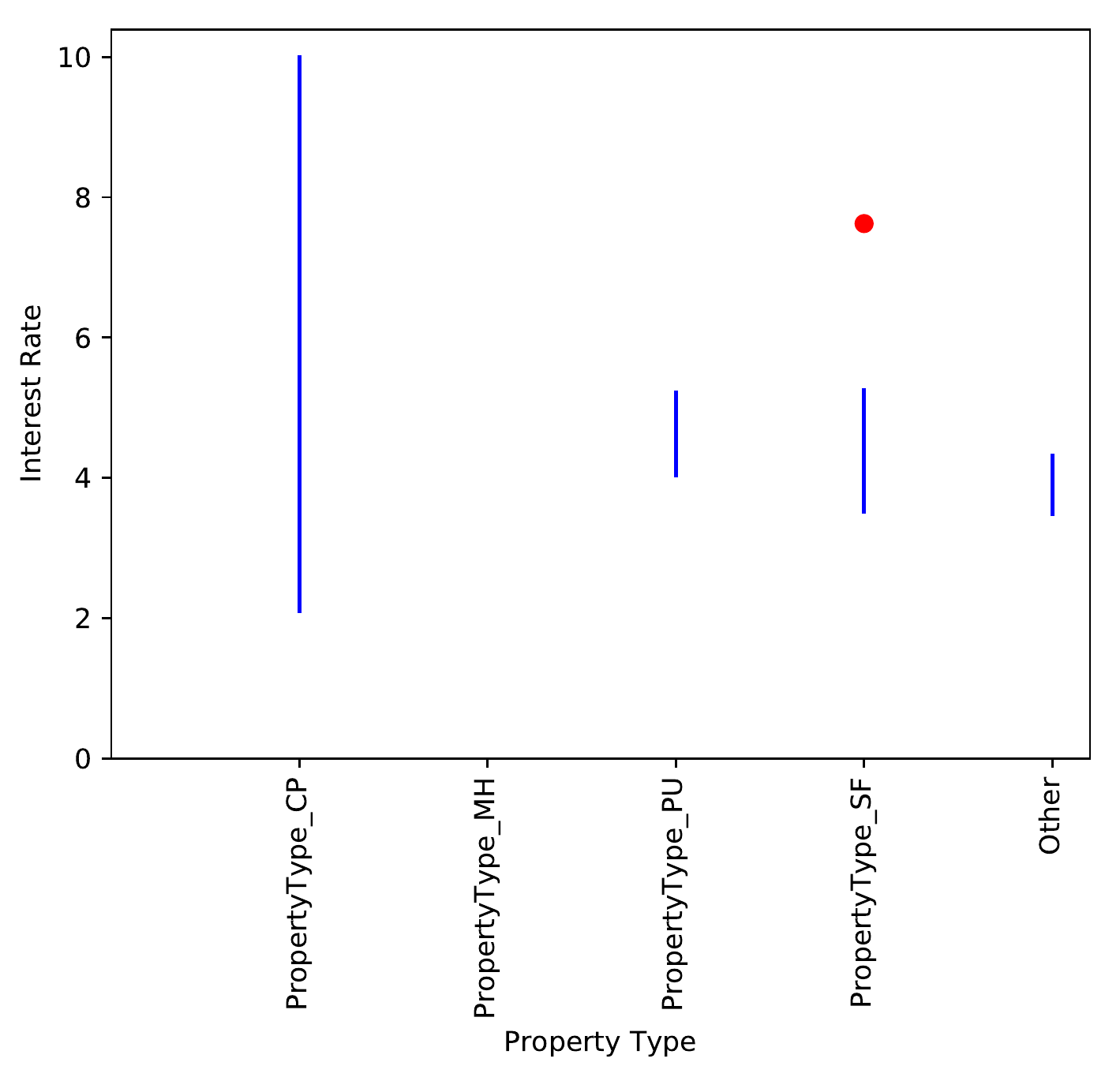} \\
\includegraphics[width=0.5\textwidth]{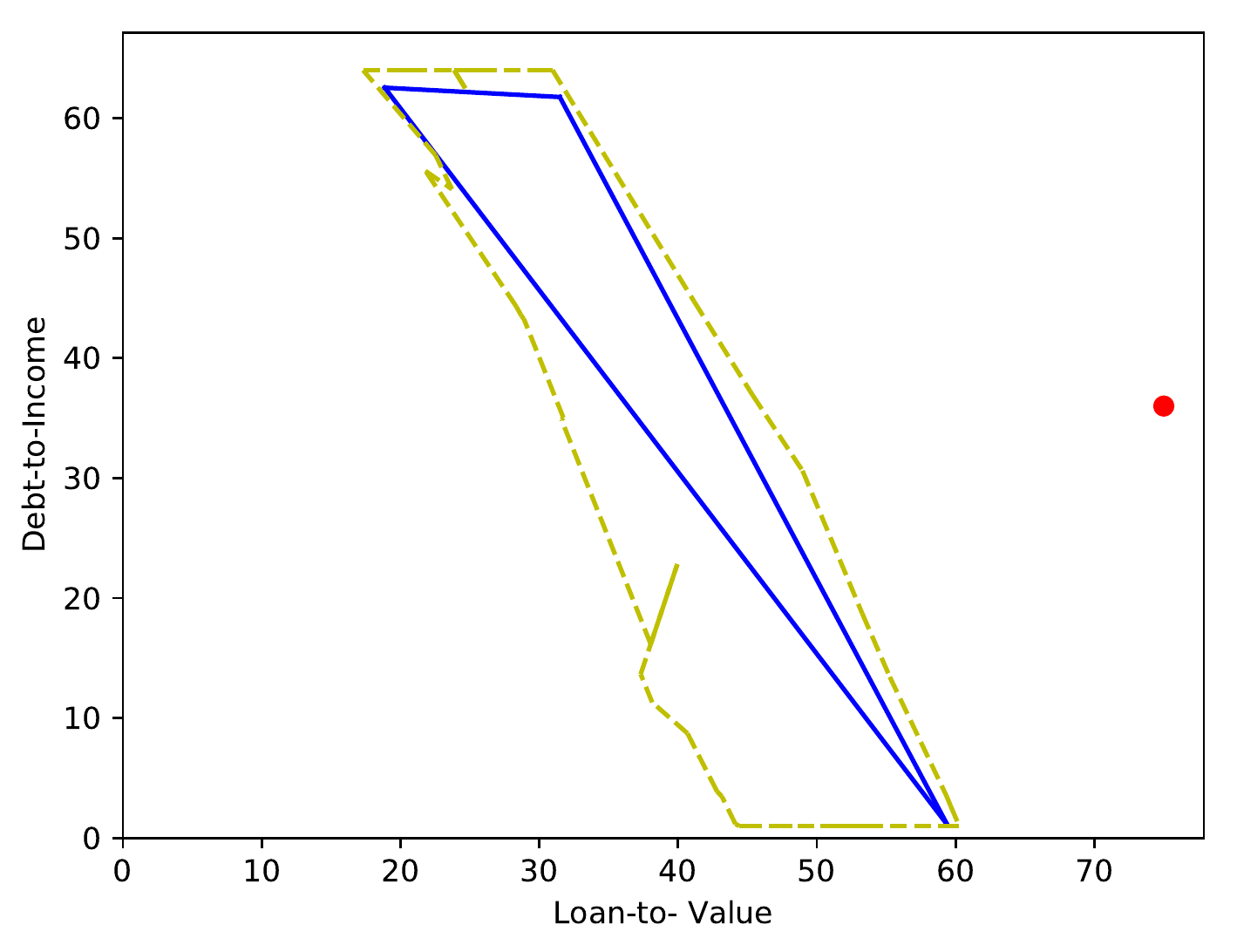}
\includegraphics[width=0.5\textwidth]{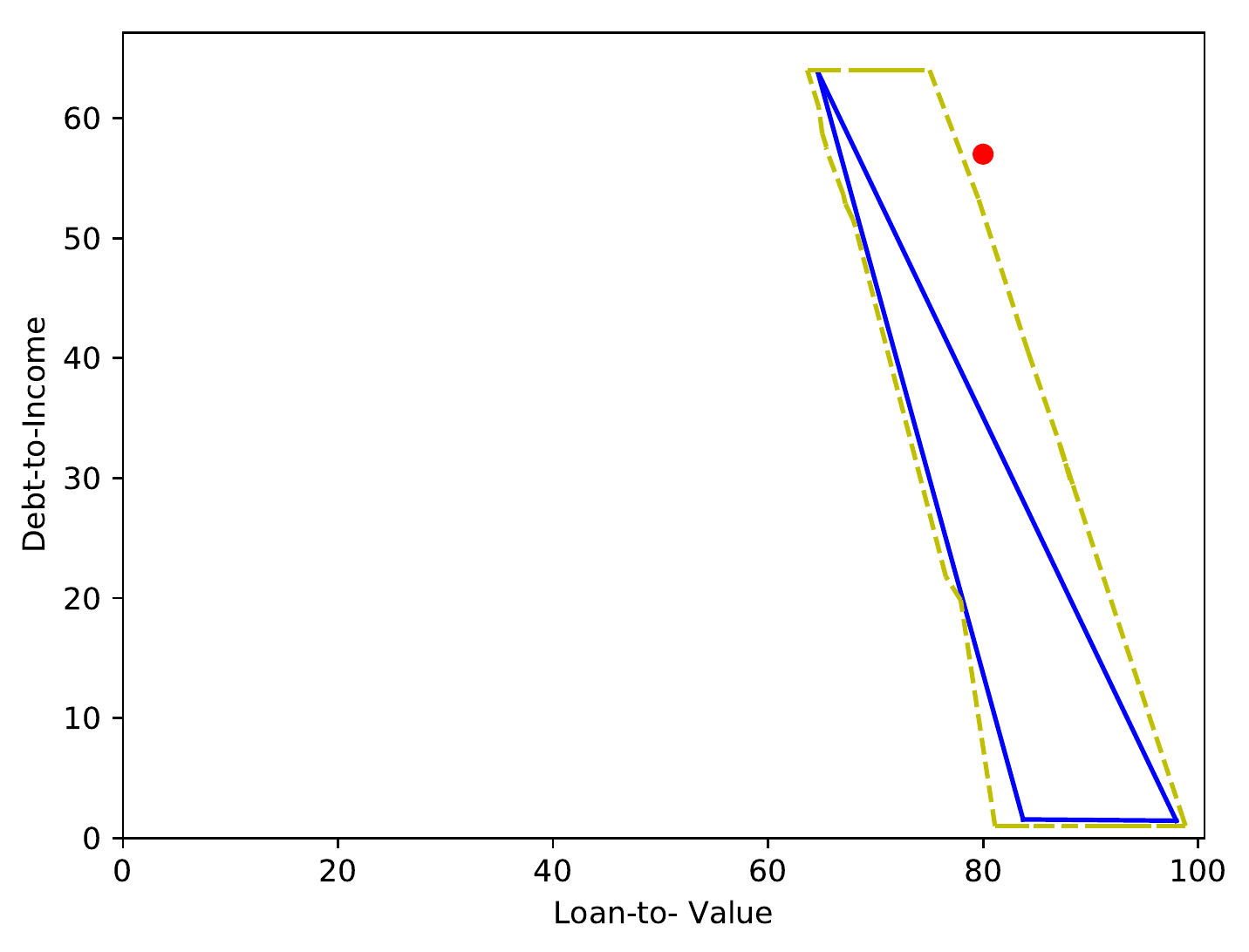} 
\end{figure}

\begin{figure}[t]
\includegraphics[width=0.5\textwidth]{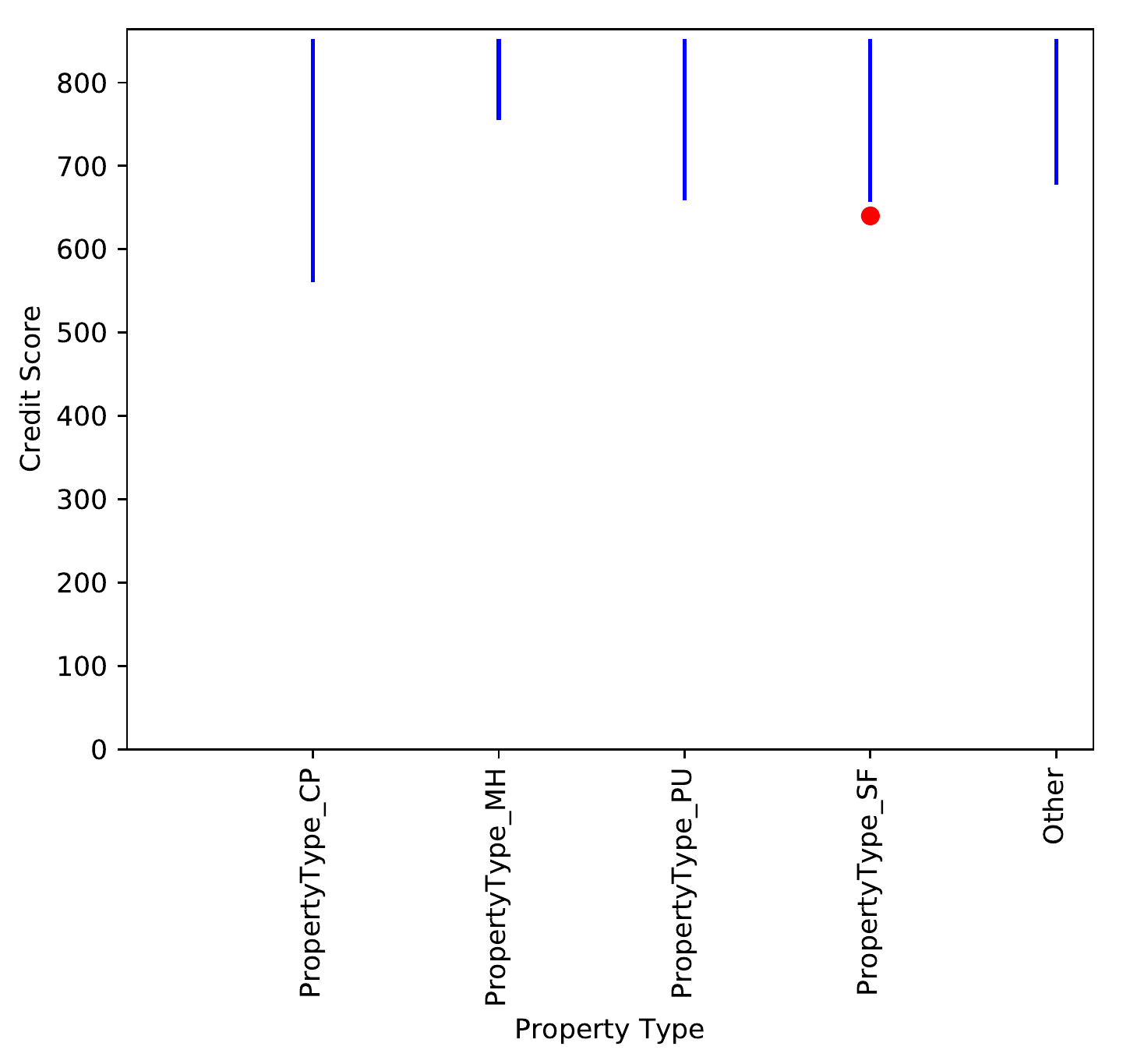}
\includegraphics[width=0.5\textwidth]{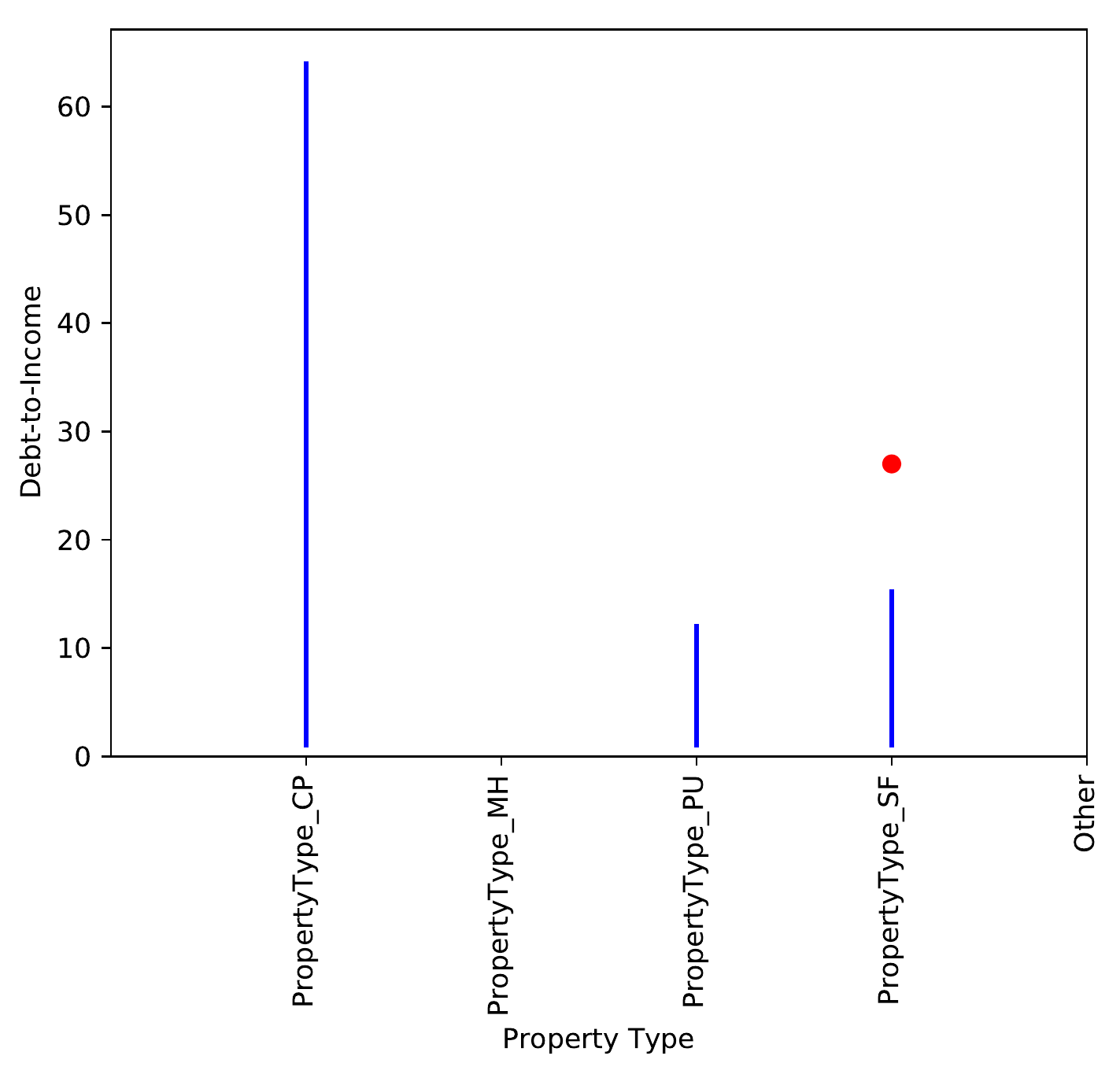} \\
\includegraphics[width=0.5\textwidth]{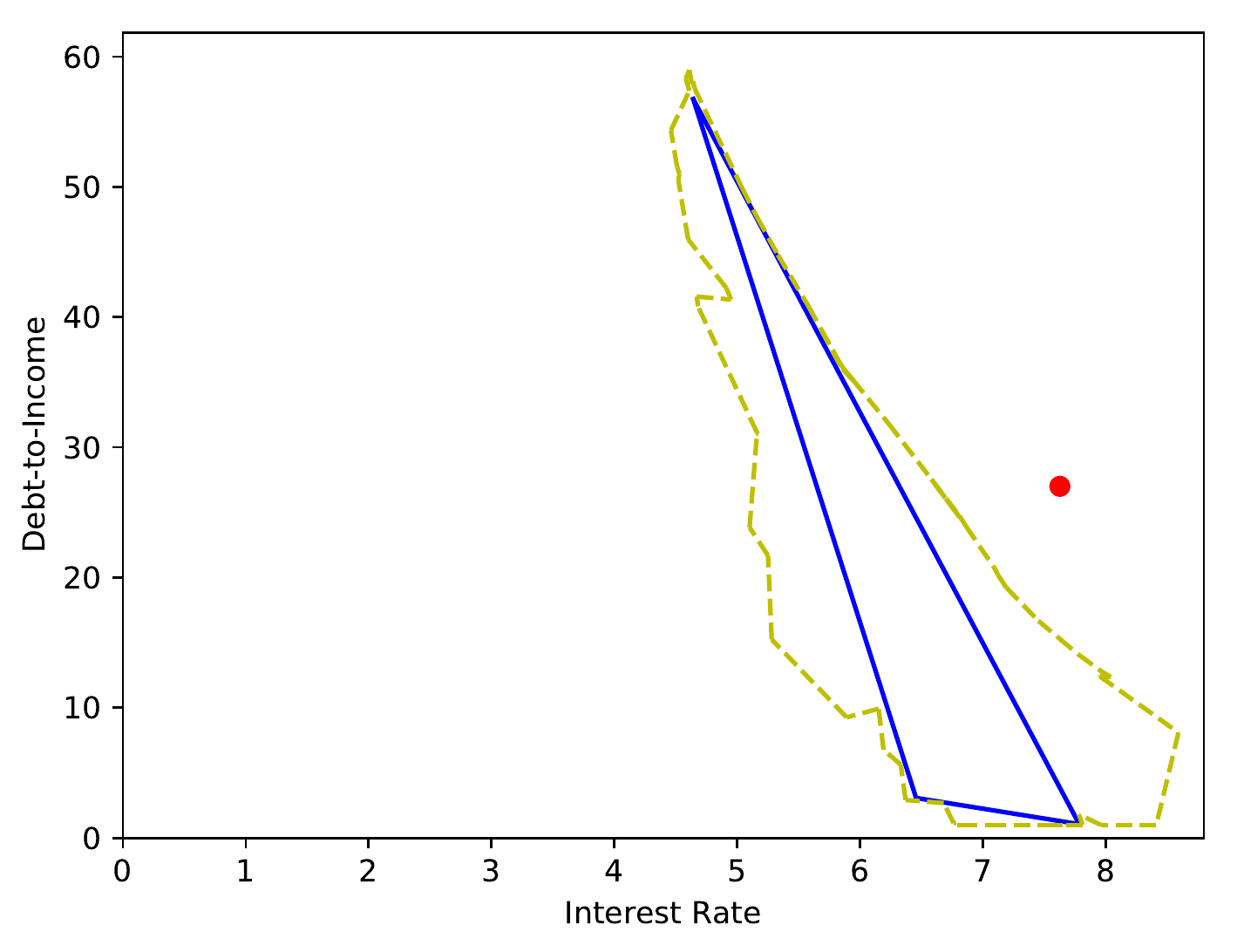}
\includegraphics[width=0.5\textwidth]{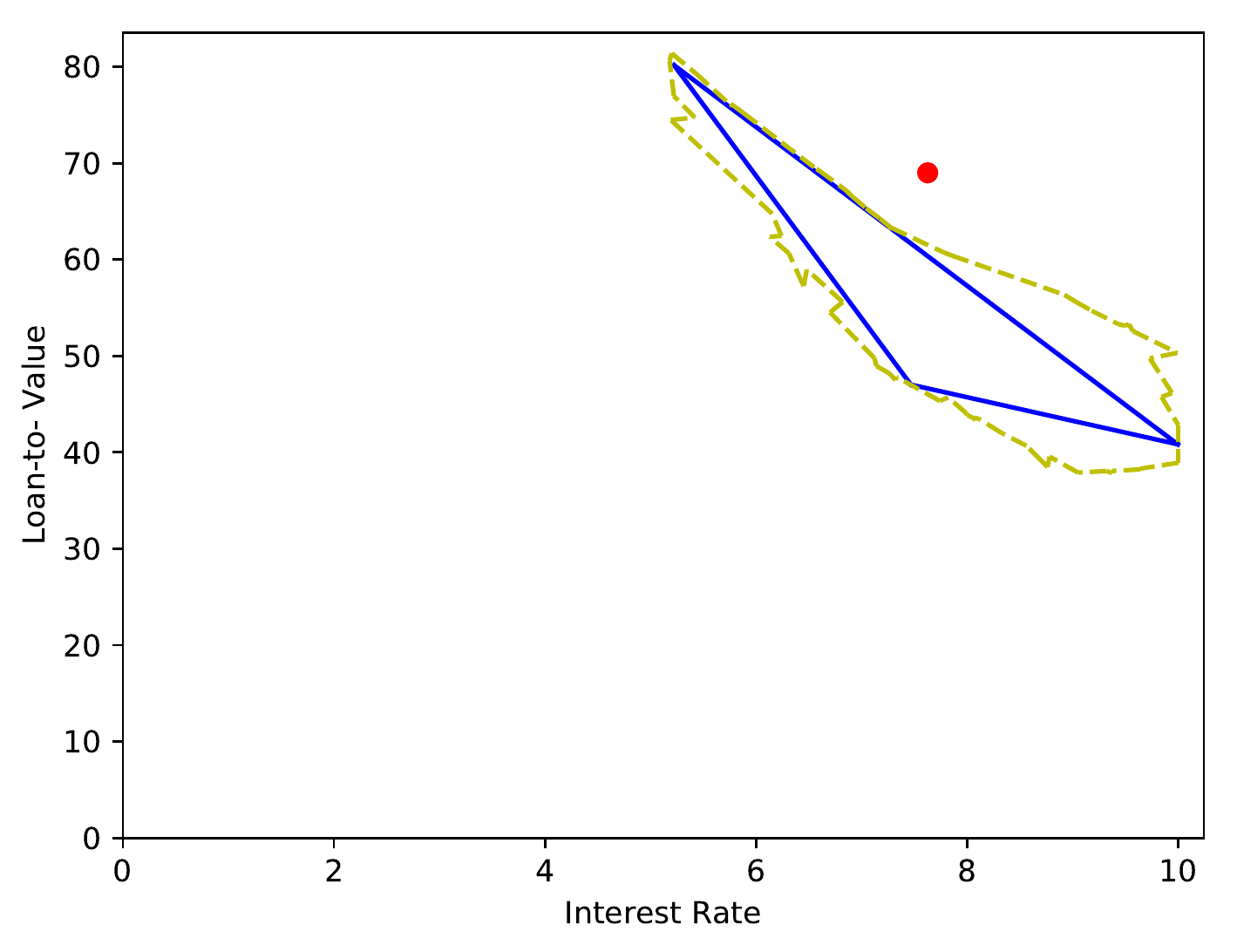} 
\end{figure}

\clearpage
\subsubsection{Theorem Proving}
The red cross shows the projection of the original application on these two features,
while the blue lines represent the set of corrected applications that the symbolic correction
would lead to.
In addition, we use a polytope enclosed in dotted yellow lines to represent the verified linear regions
collected by Algorithm~\ref{alg:search}.
\begin{figure}[h]
	\includegraphics[width=0.5\textwidth]{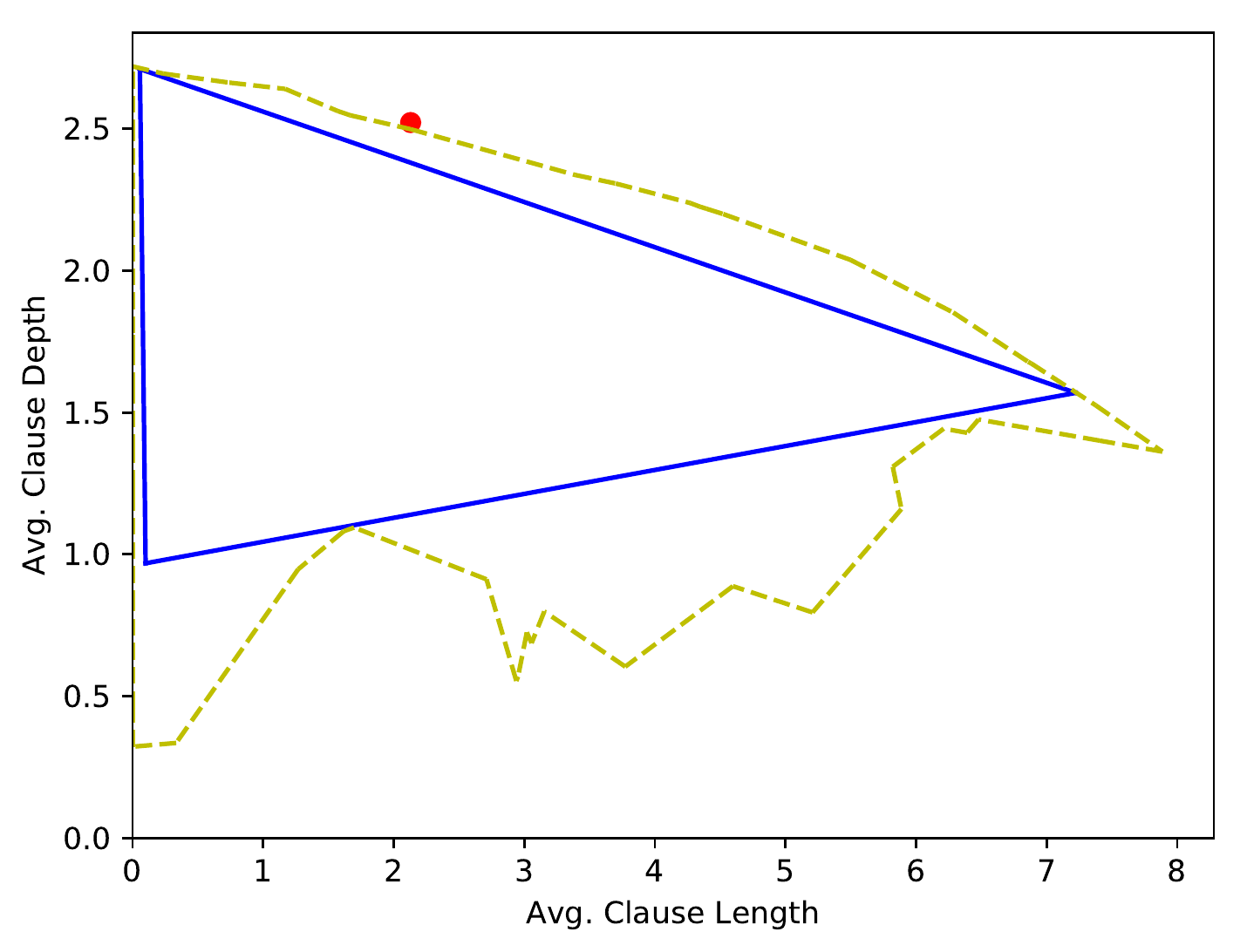}
	\includegraphics[width=0.5\textwidth]{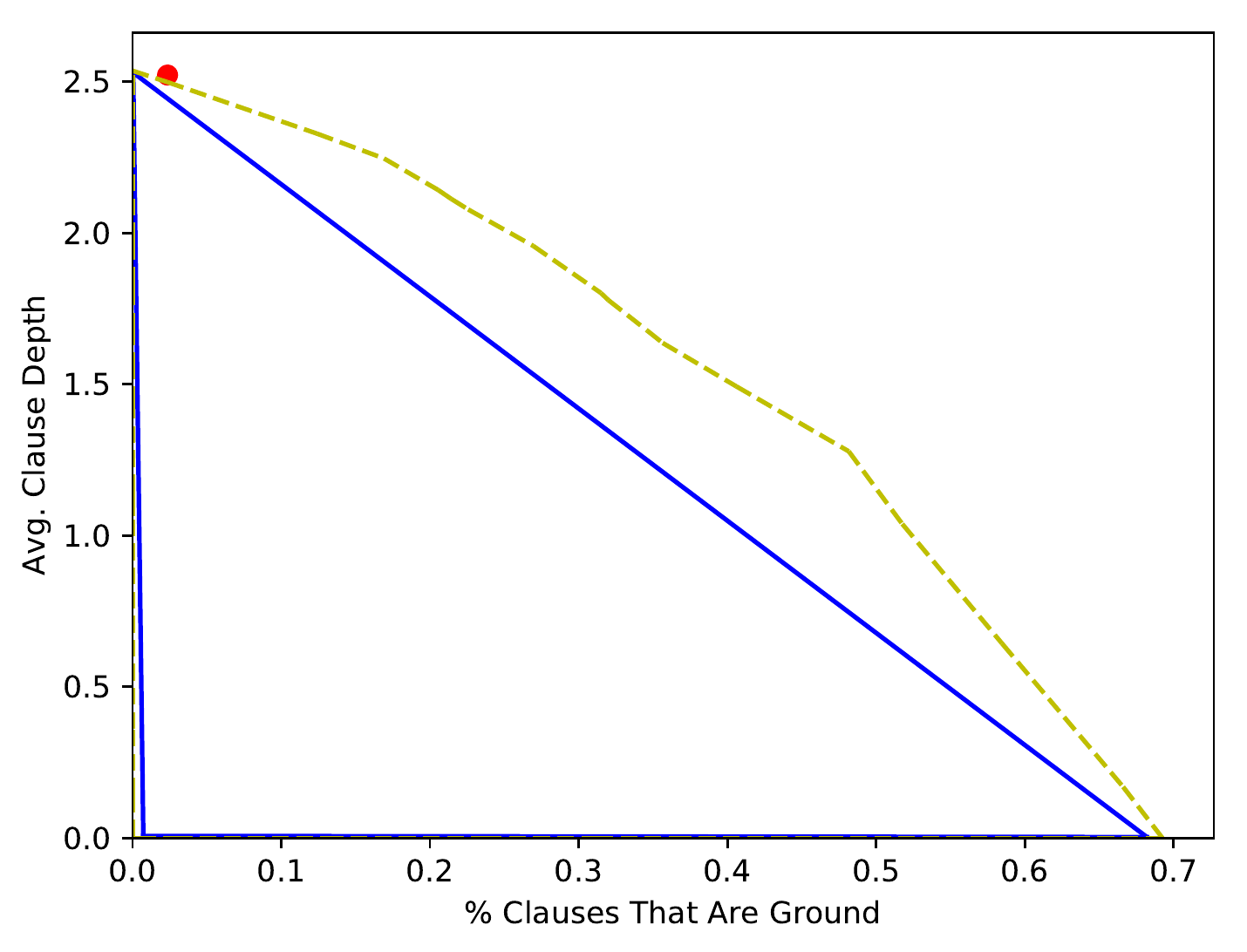} \\
	\includegraphics[width=0.5\textwidth]{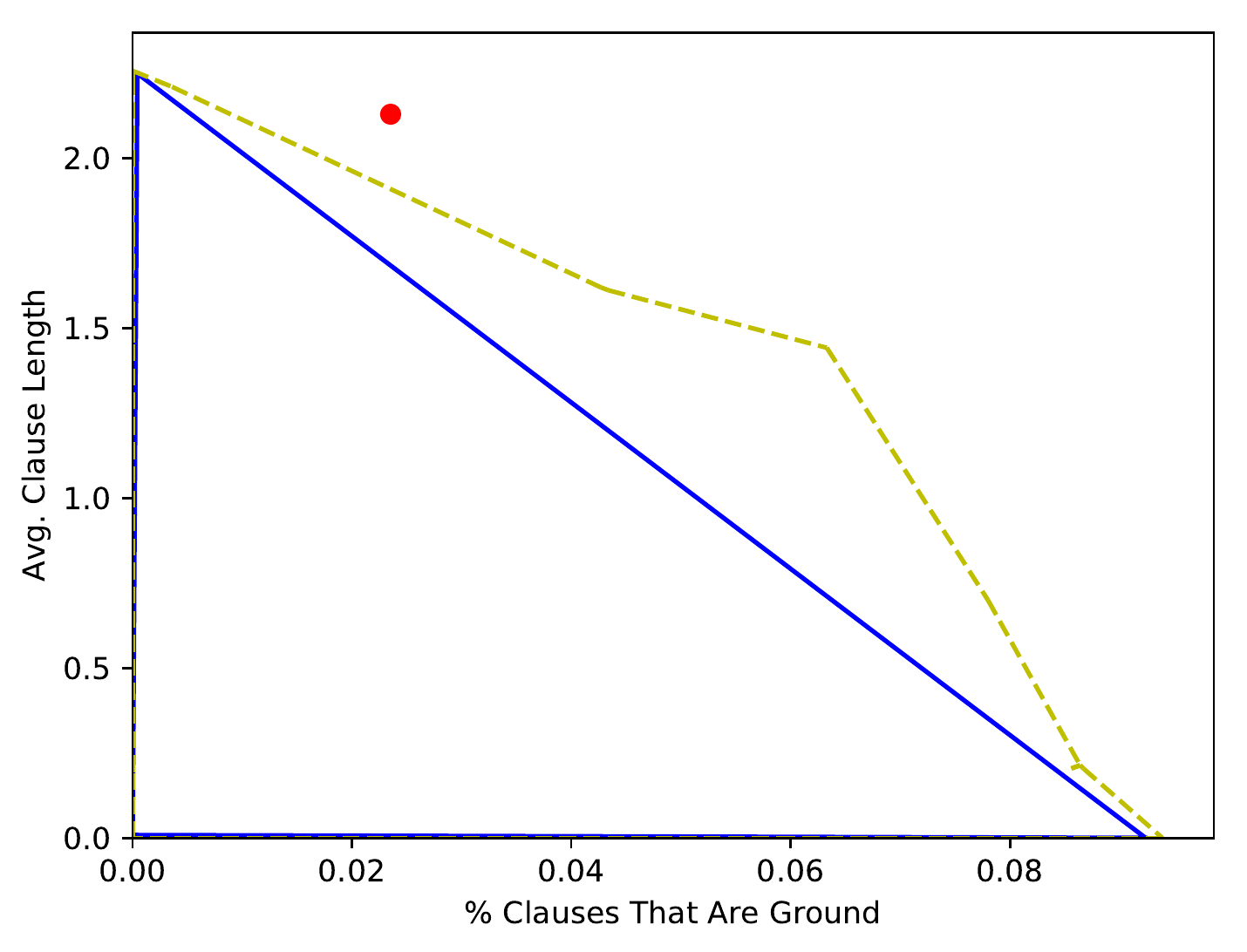}
	\includegraphics[width=0.5\textwidth]{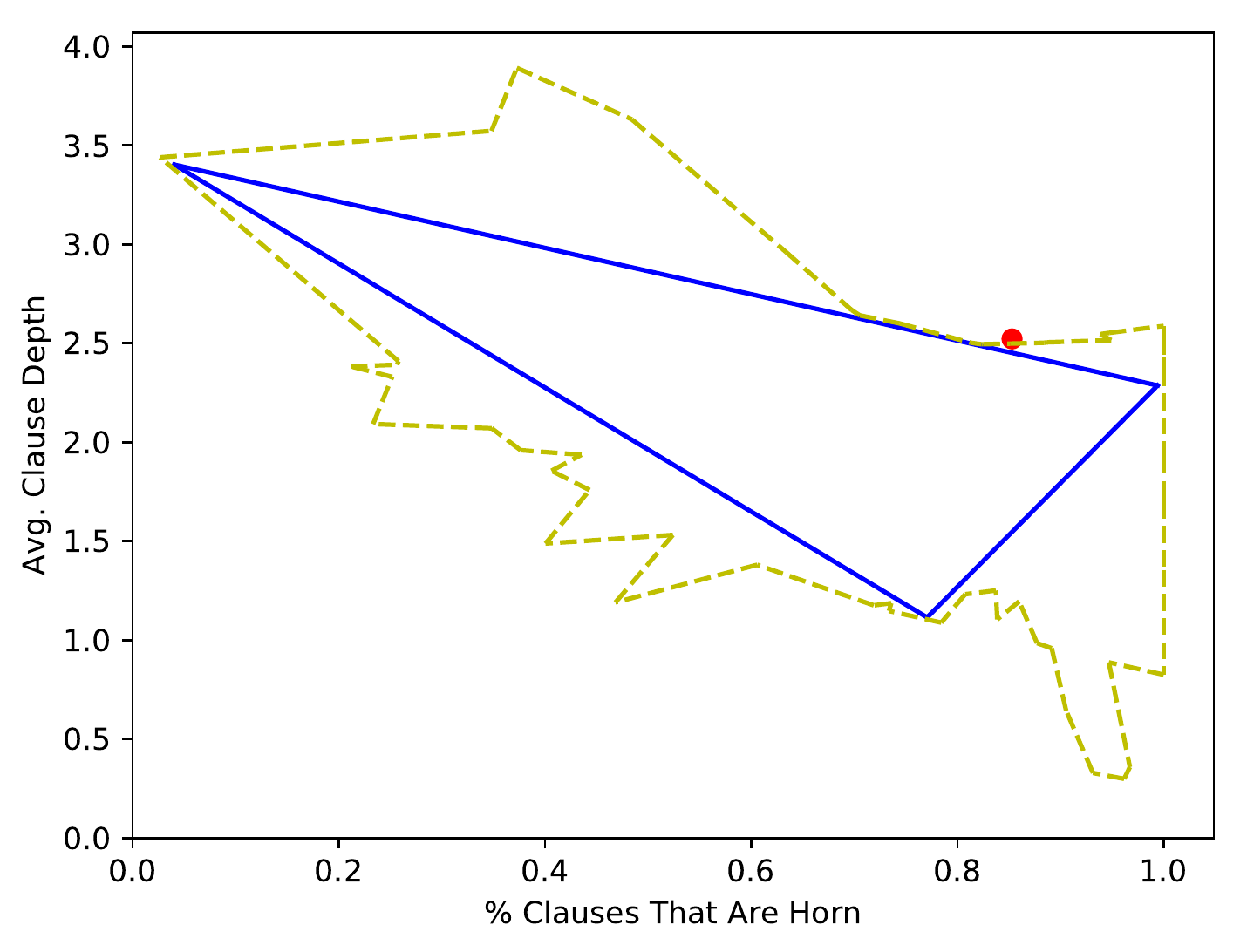} \\
	\includegraphics[width=0.5\textwidth]{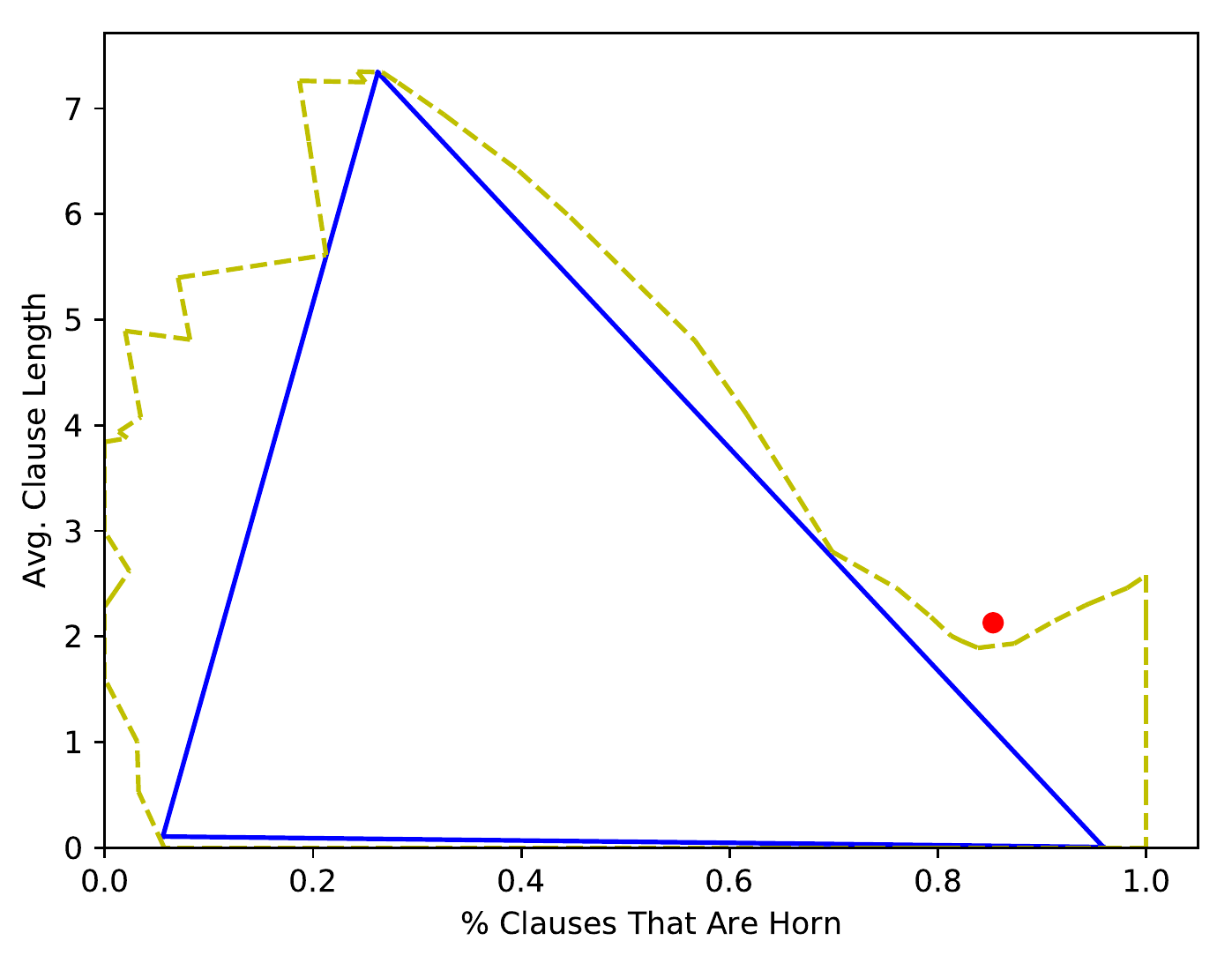}
	\includegraphics[width=0.5\textwidth]{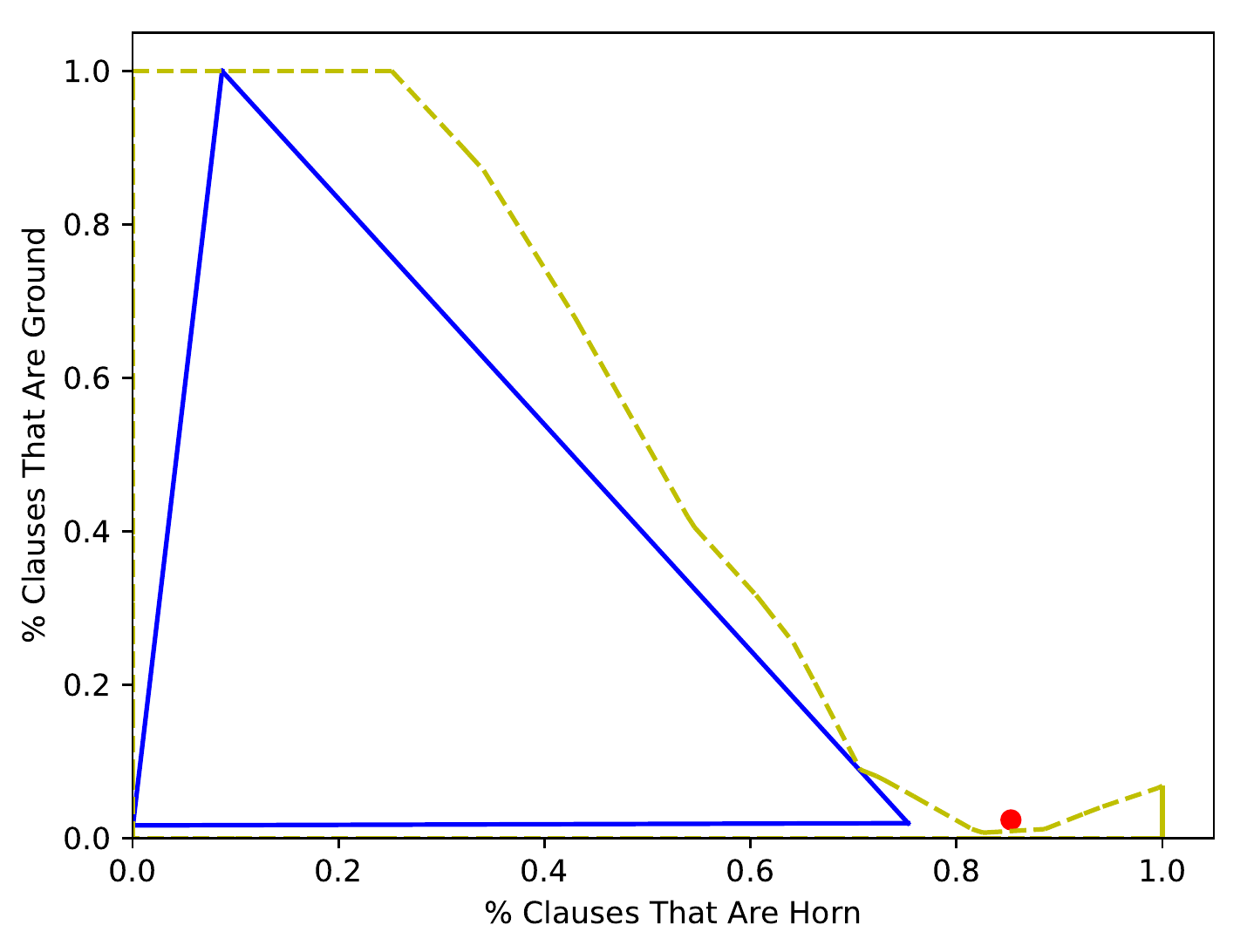} 
\end{figure}

\begin{figure}[t]
	\includegraphics[width=0.5\textwidth]{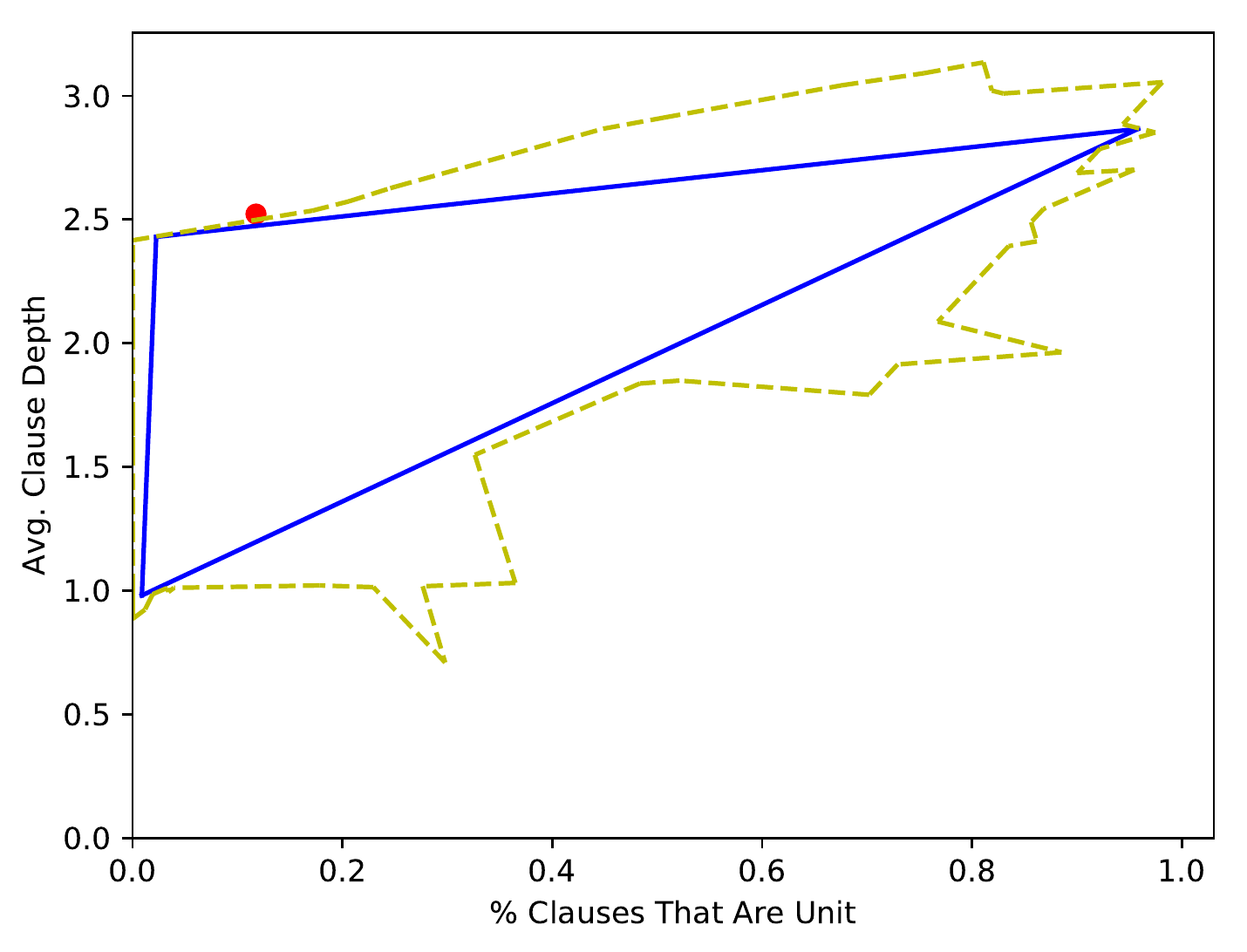}
	\includegraphics[width=0.5\textwidth]{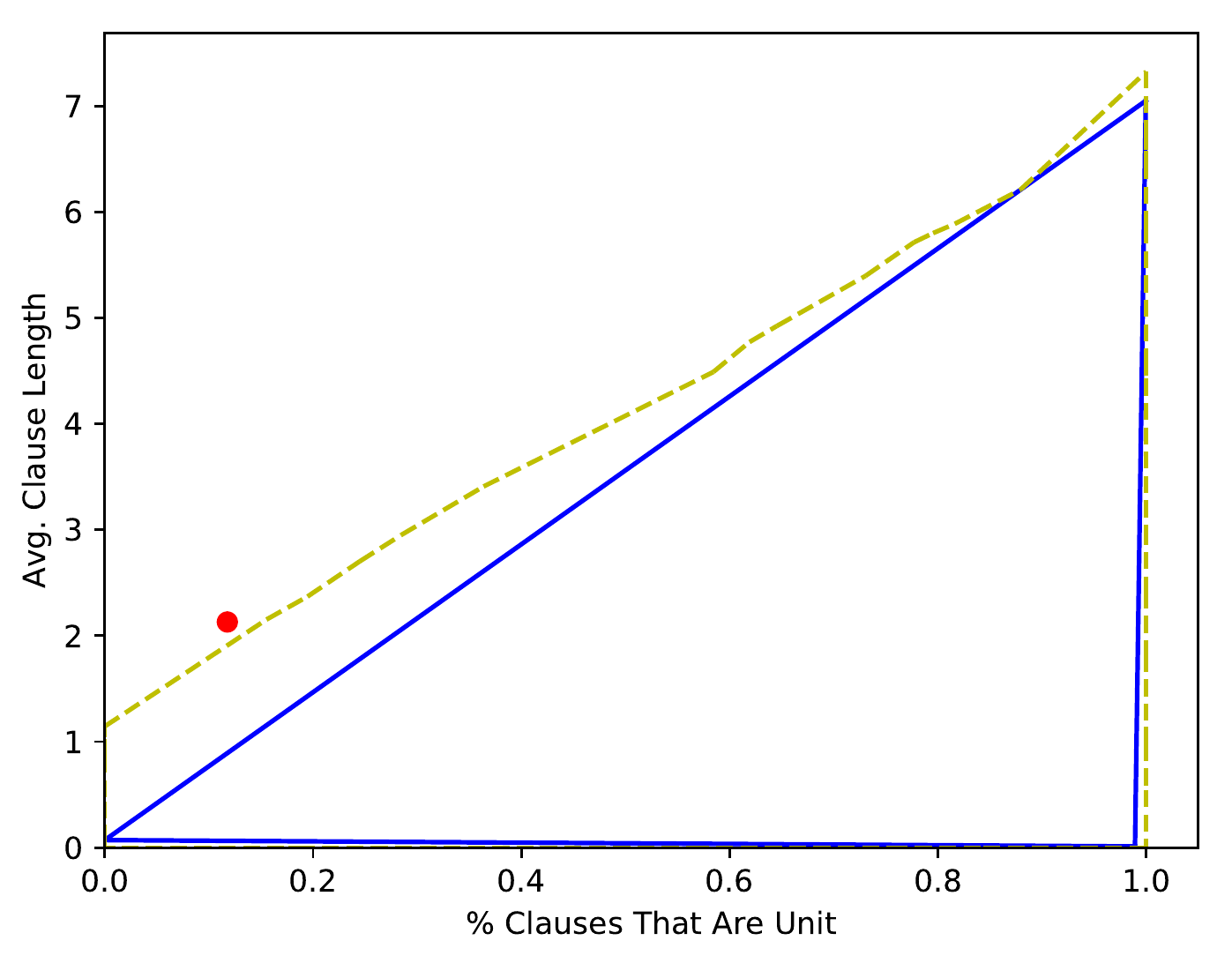} \\
	\includegraphics[width=0.5\textwidth]{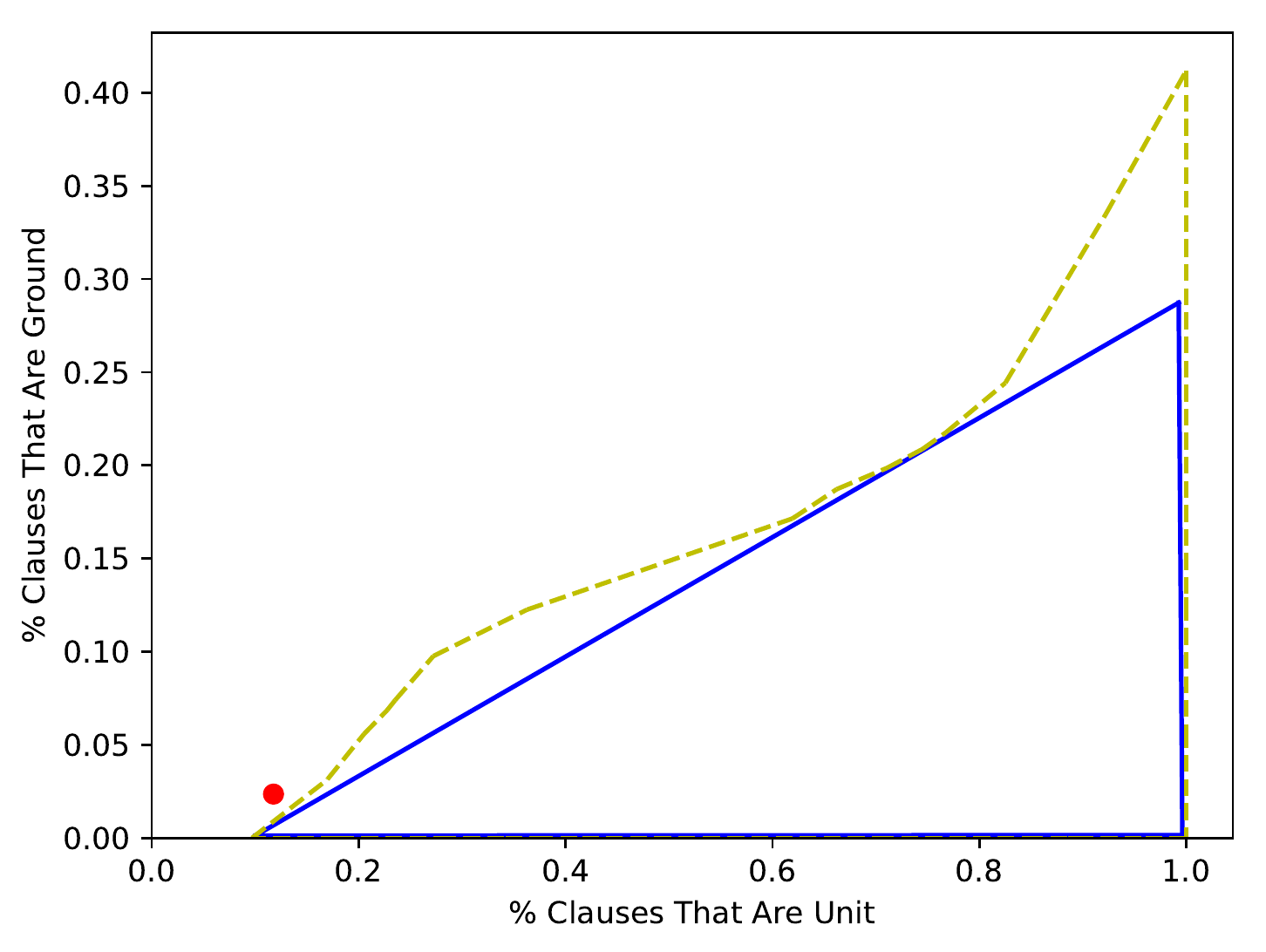}
	\includegraphics[width=0.5\textwidth]{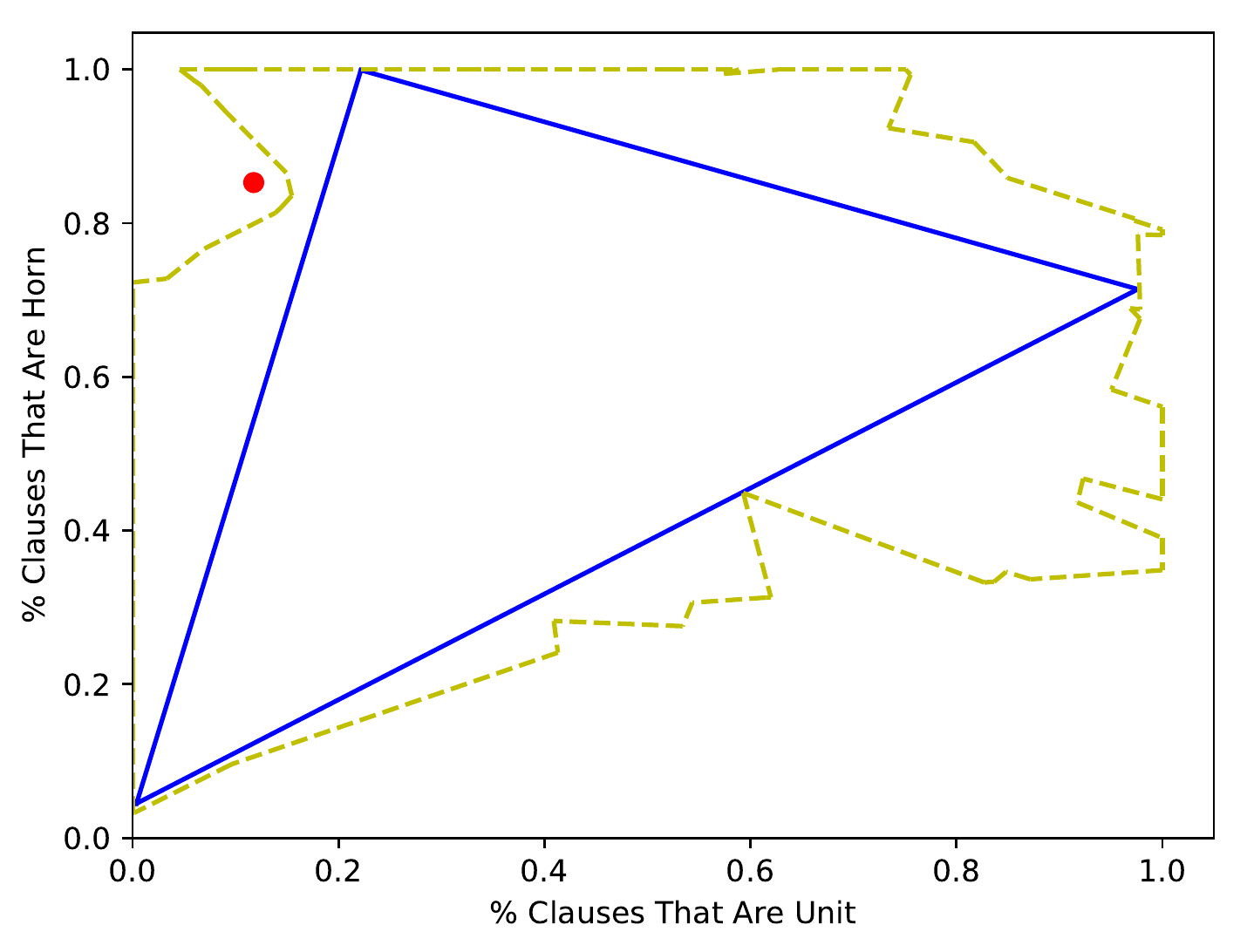} \\
	\includegraphics[width=0.5\textwidth]{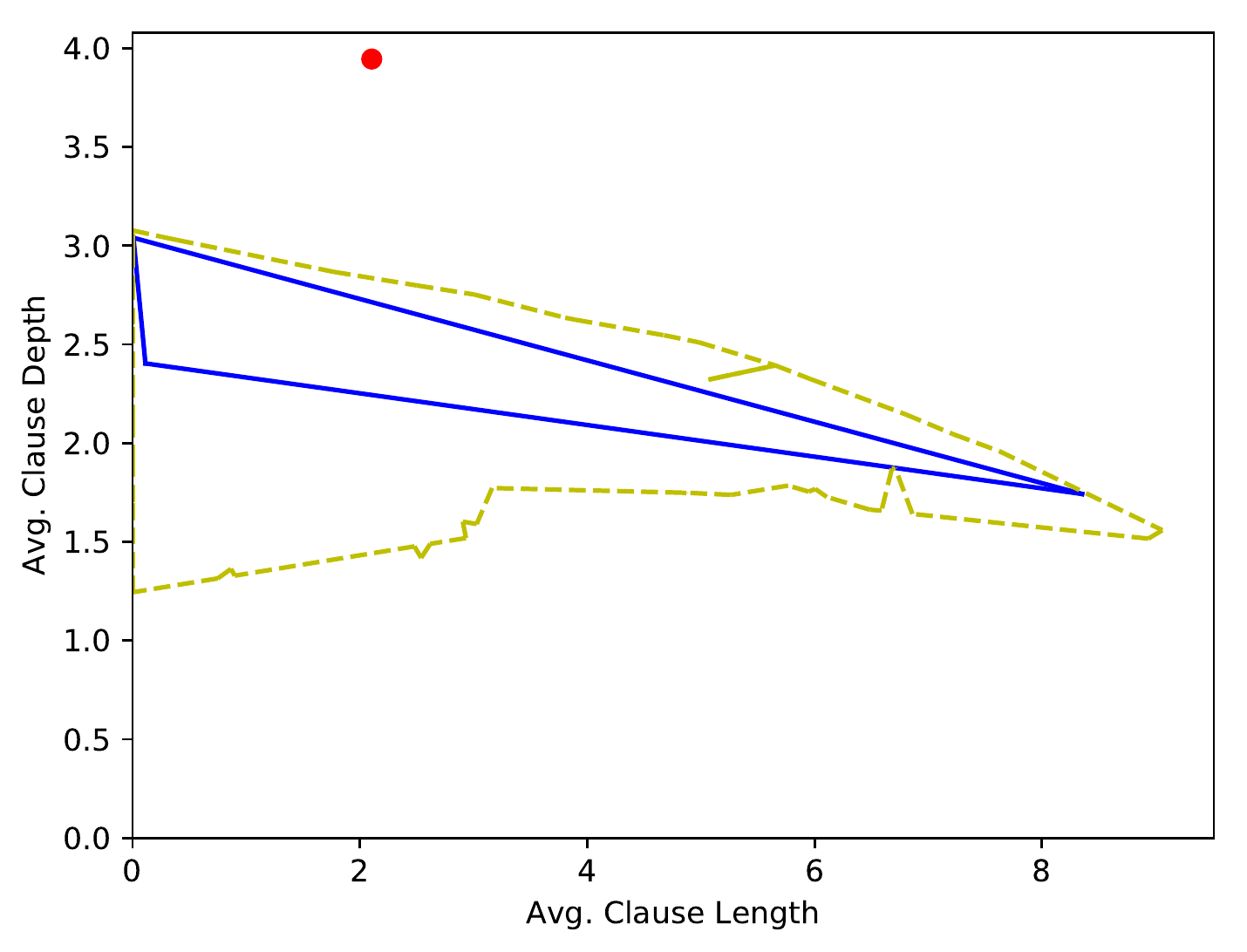}
	\includegraphics[width=0.5\textwidth]{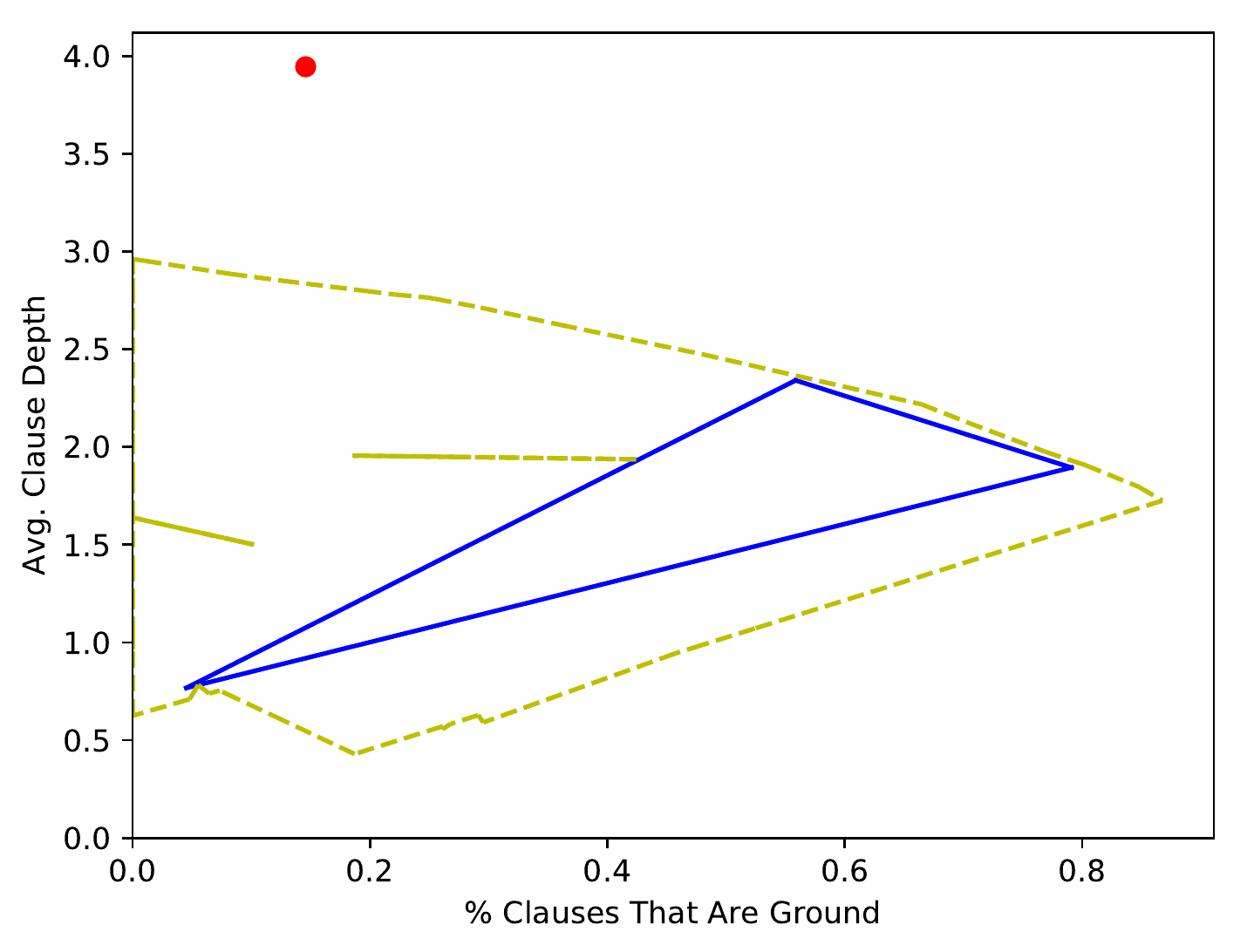} 
\end{figure}

\begin{figure}[h]
	\includegraphics[width=0.5\textwidth]{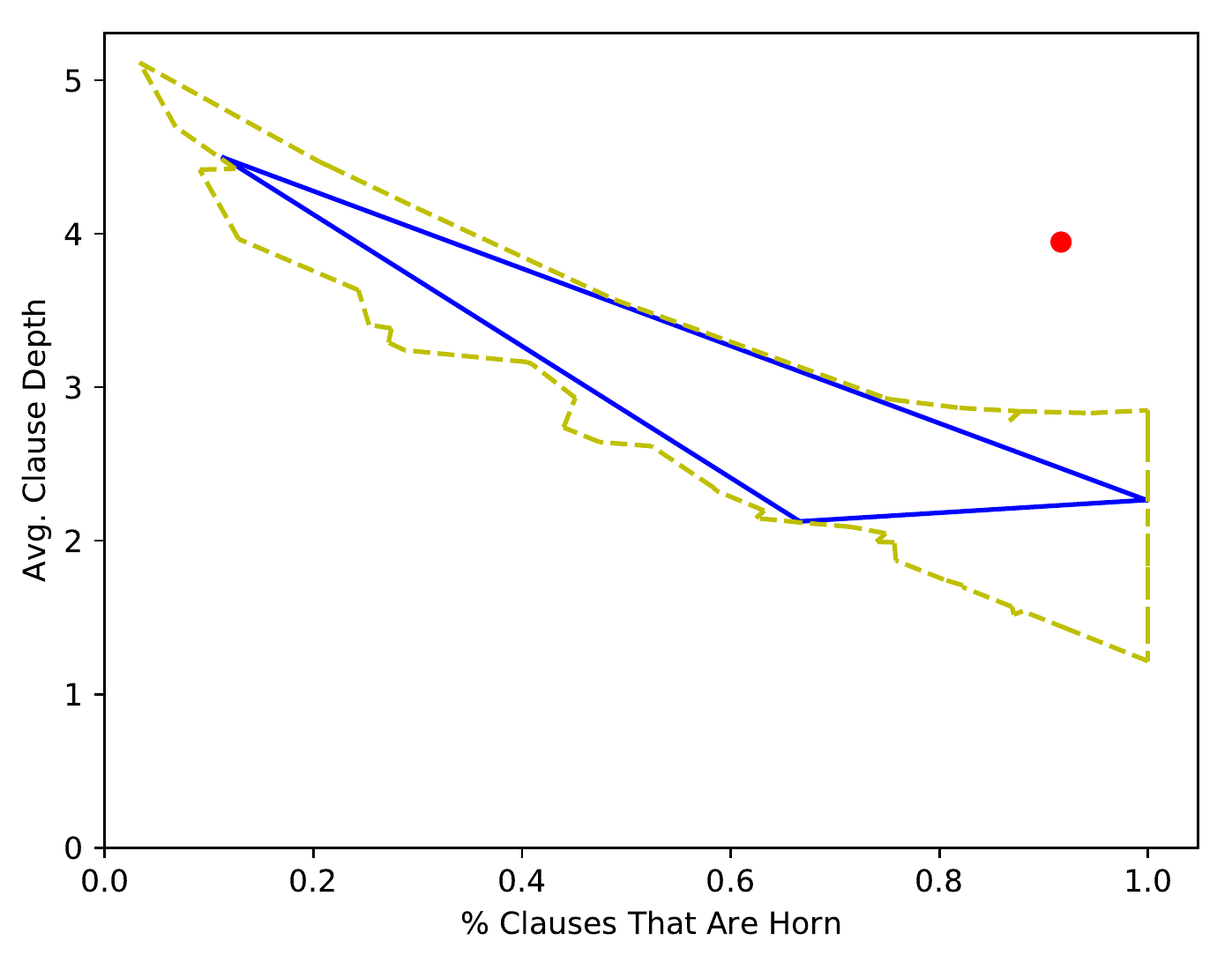}
	\includegraphics[width=0.5\textwidth]{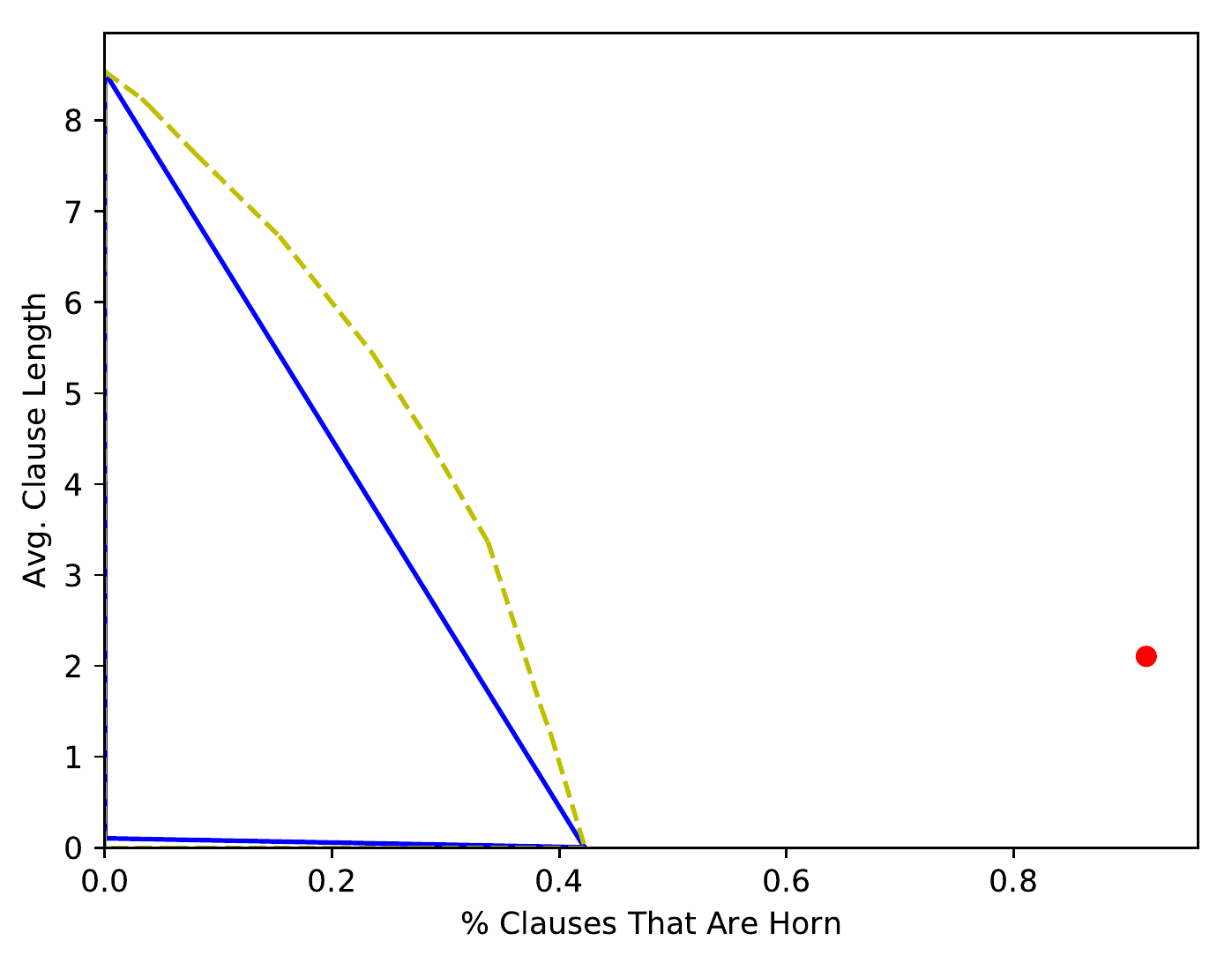} \\
	\includegraphics[width=0.5\textwidth]{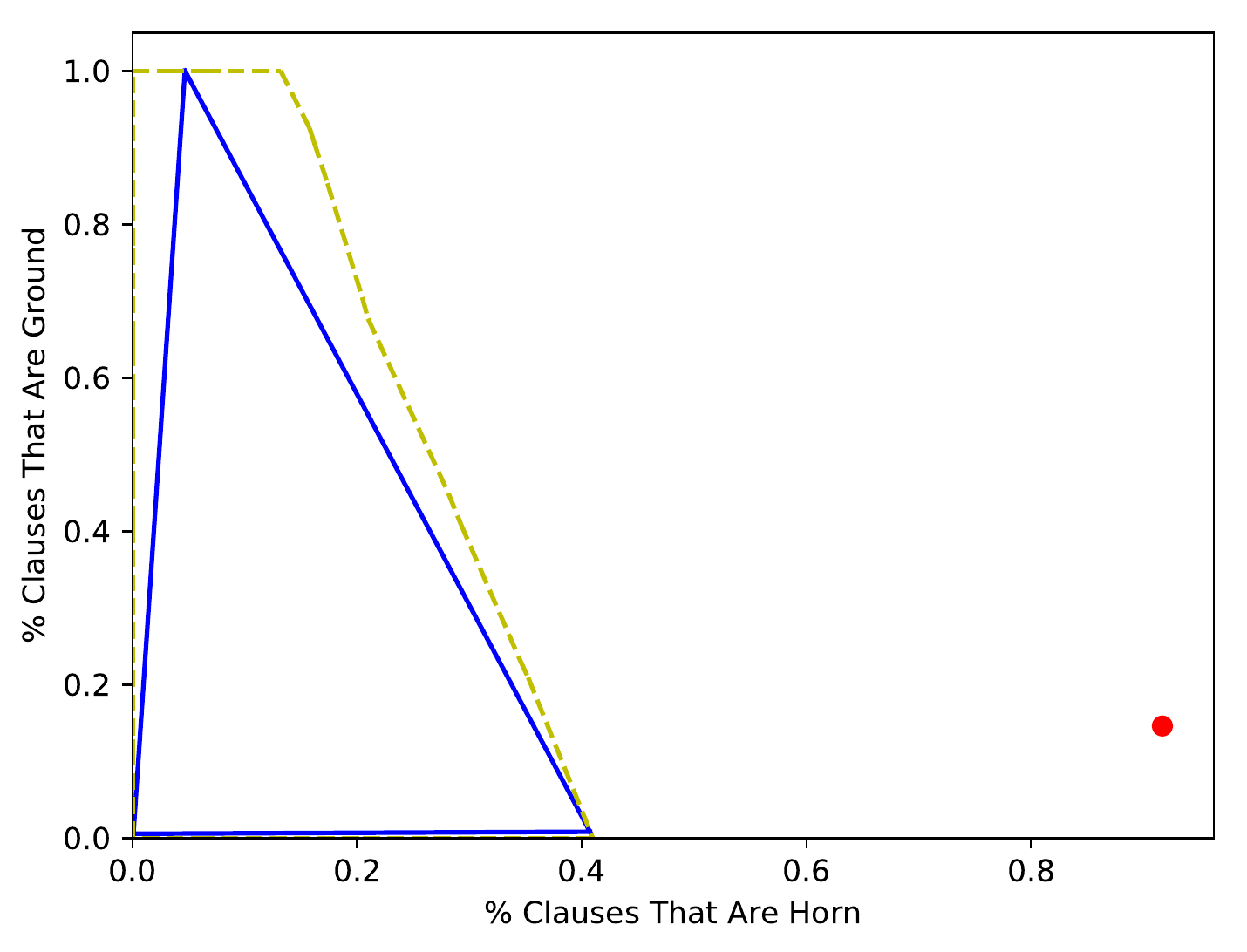}
	\includegraphics[width=0.5\textwidth]{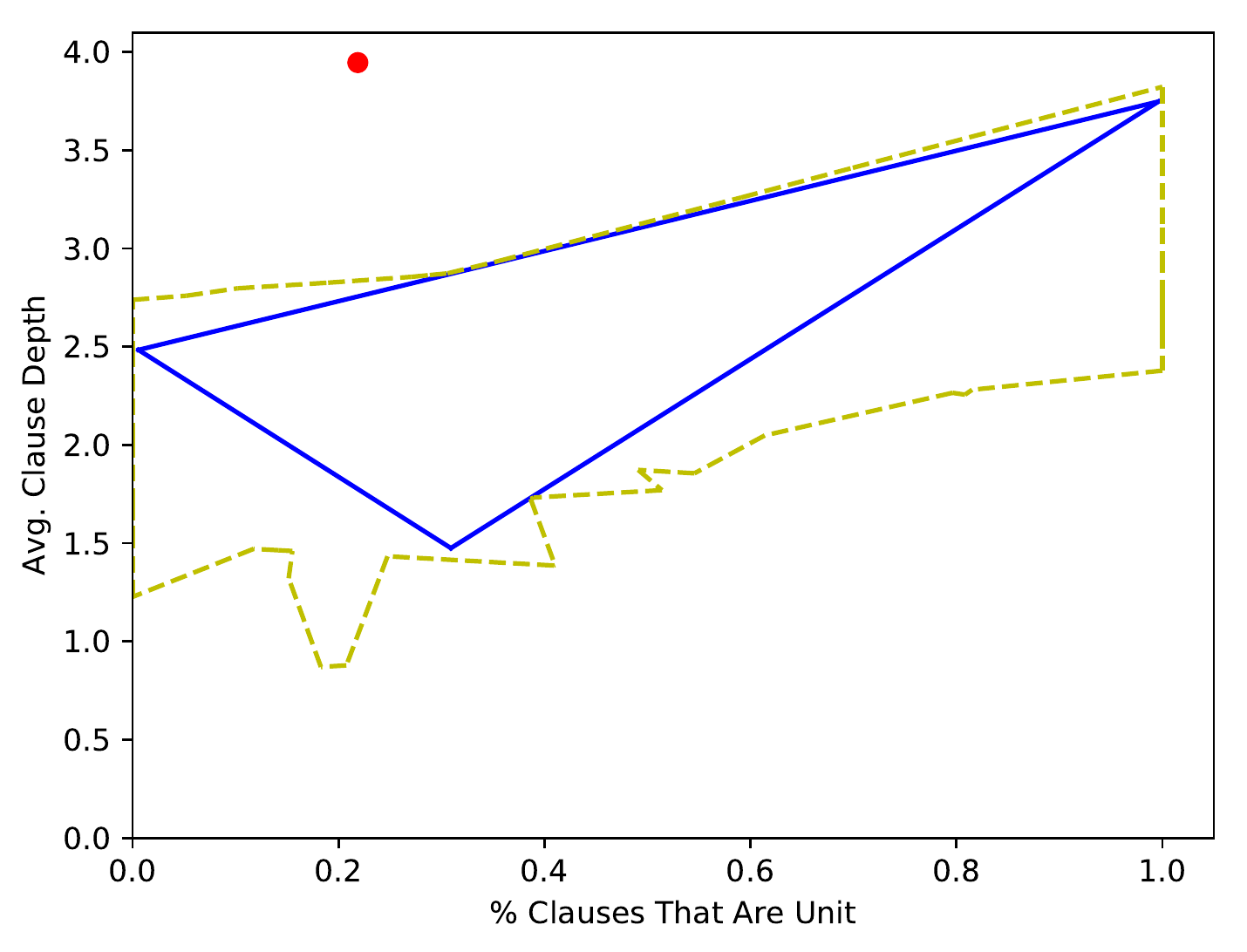} \\
	\includegraphics[width=0.5\textwidth]{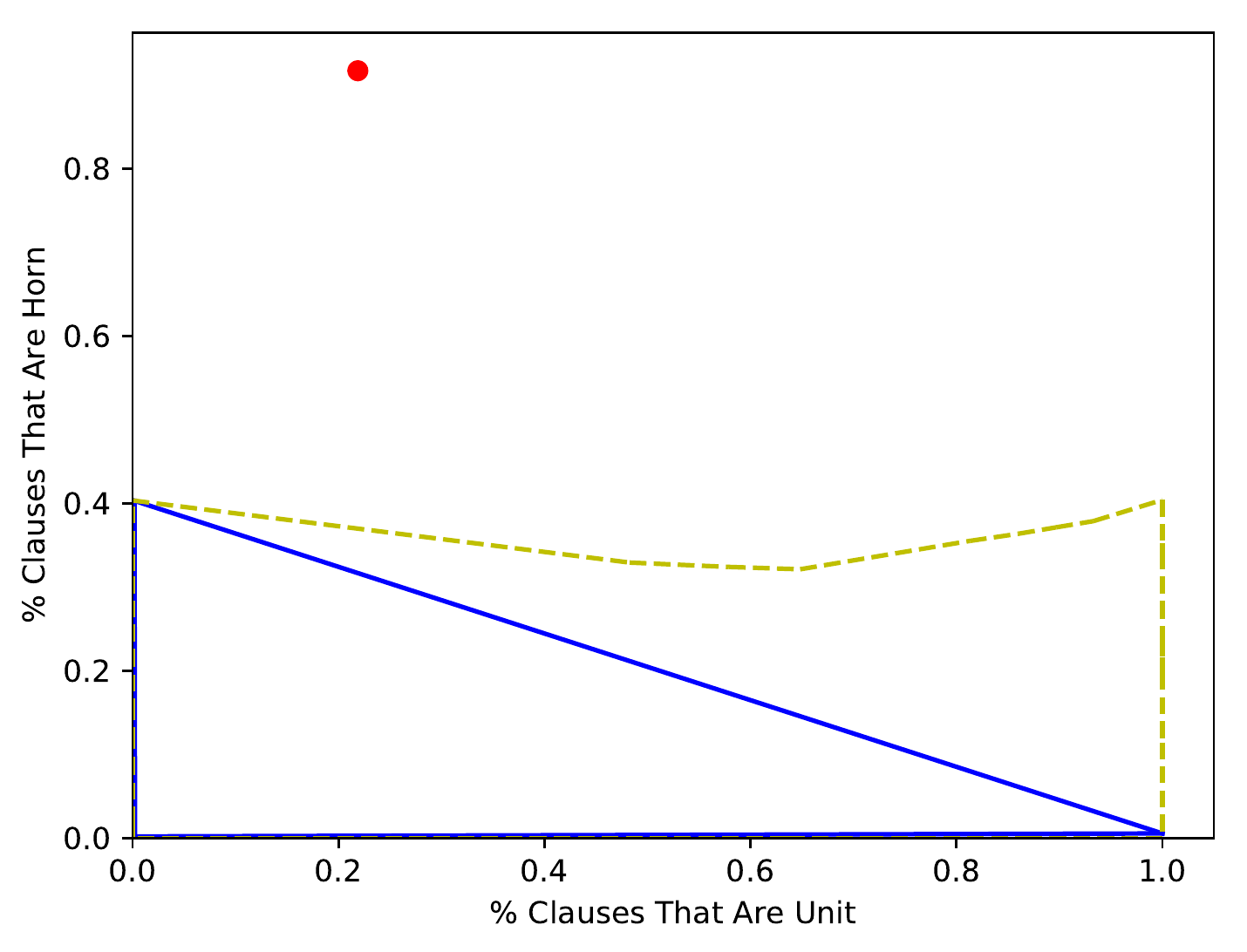}
	\includegraphics[width=0.5\textwidth]{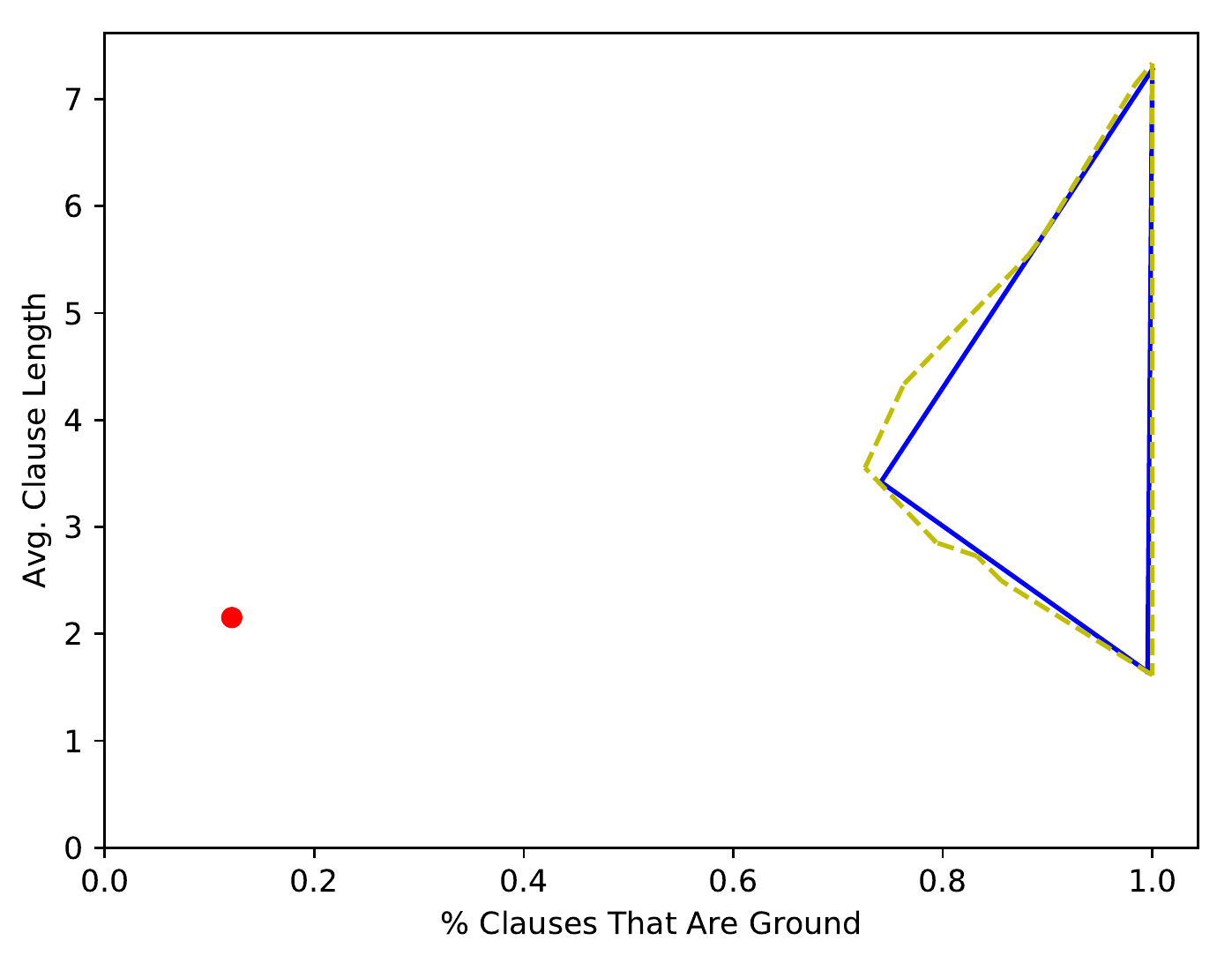} 
\end{figure}

\begin{figure}[t]
	\includegraphics[width=0.5\textwidth]{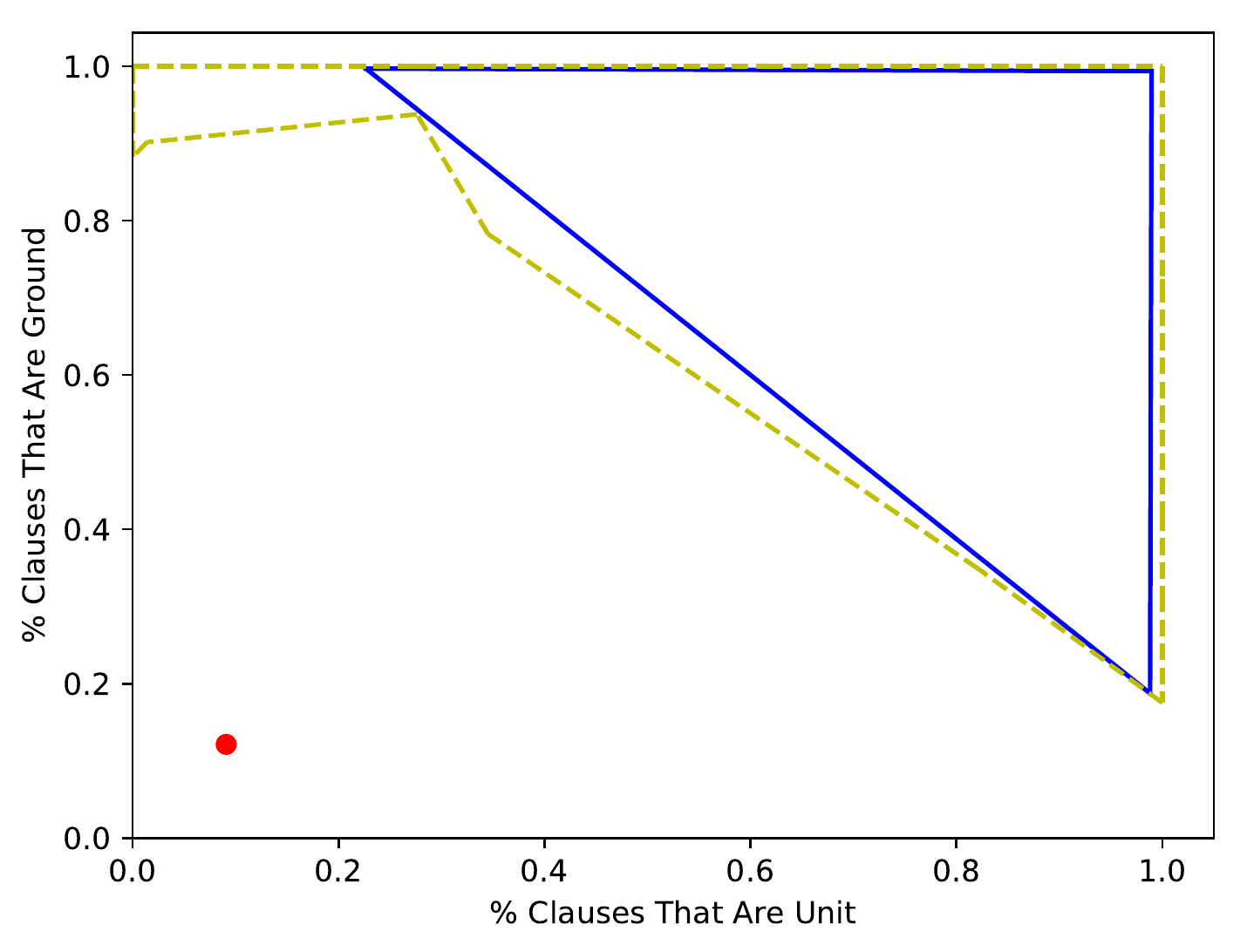}
	\includegraphics[width=0.5\textwidth]{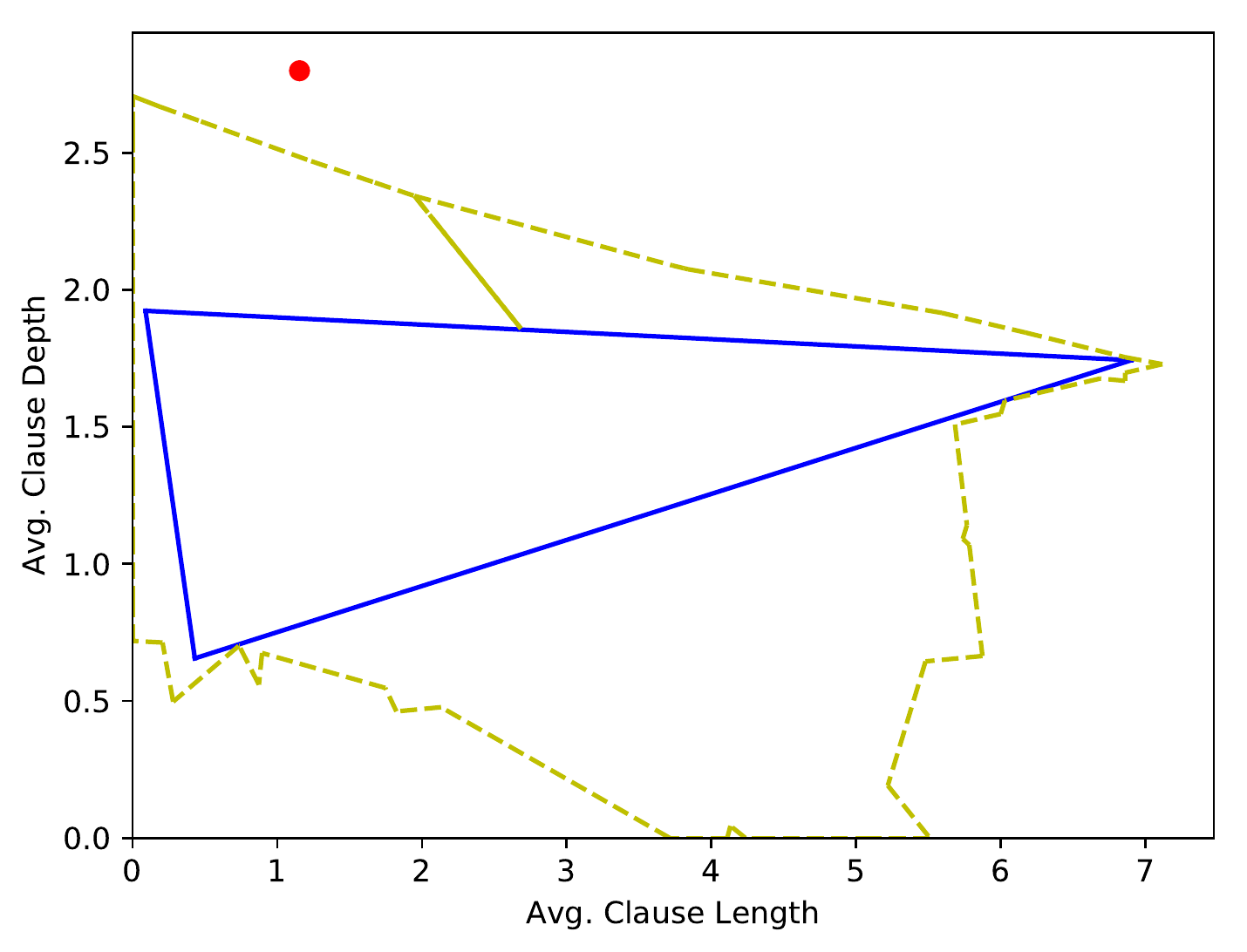} \\
	\includegraphics[width=0.5\textwidth]{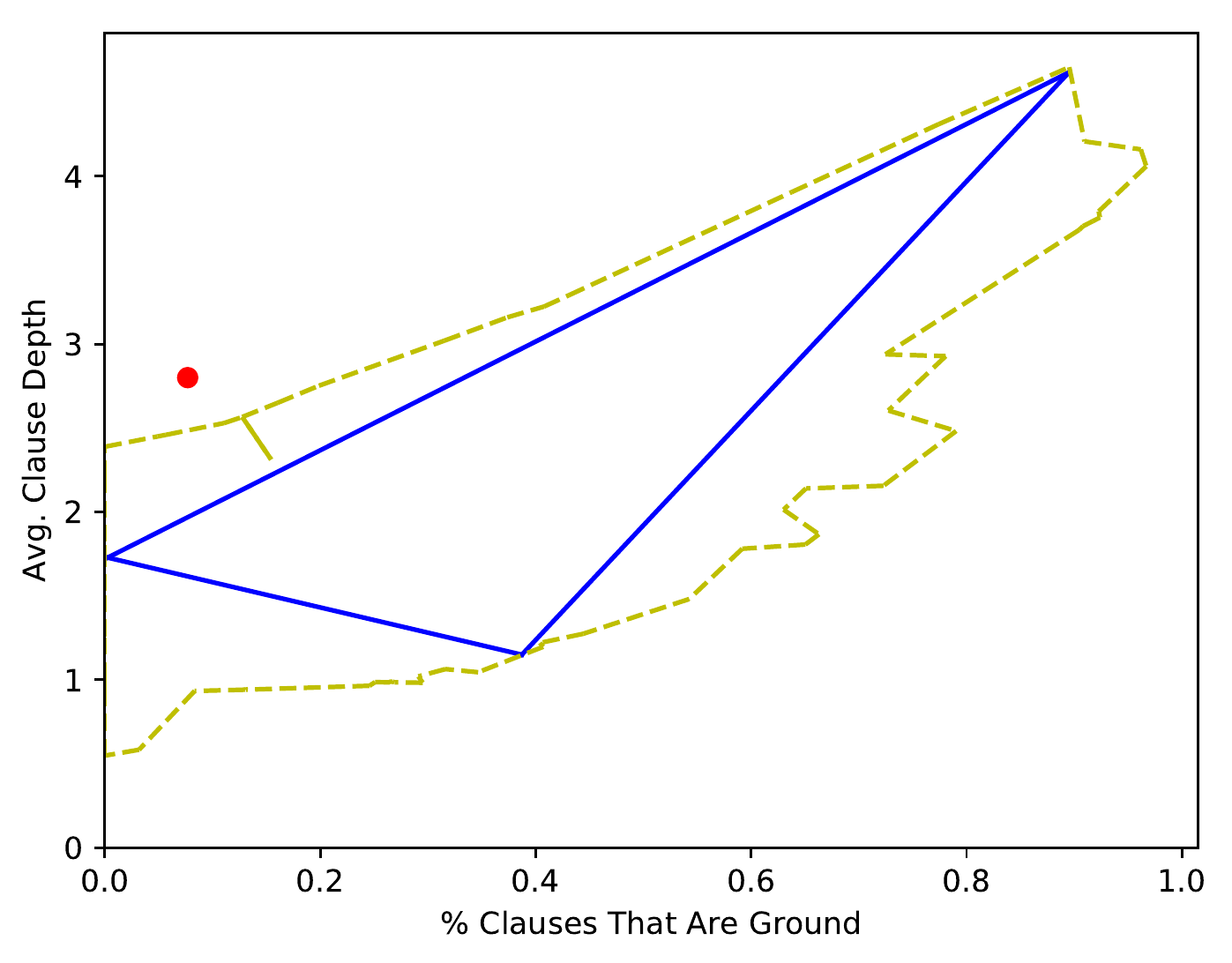}
	\includegraphics[width=0.5\textwidth]{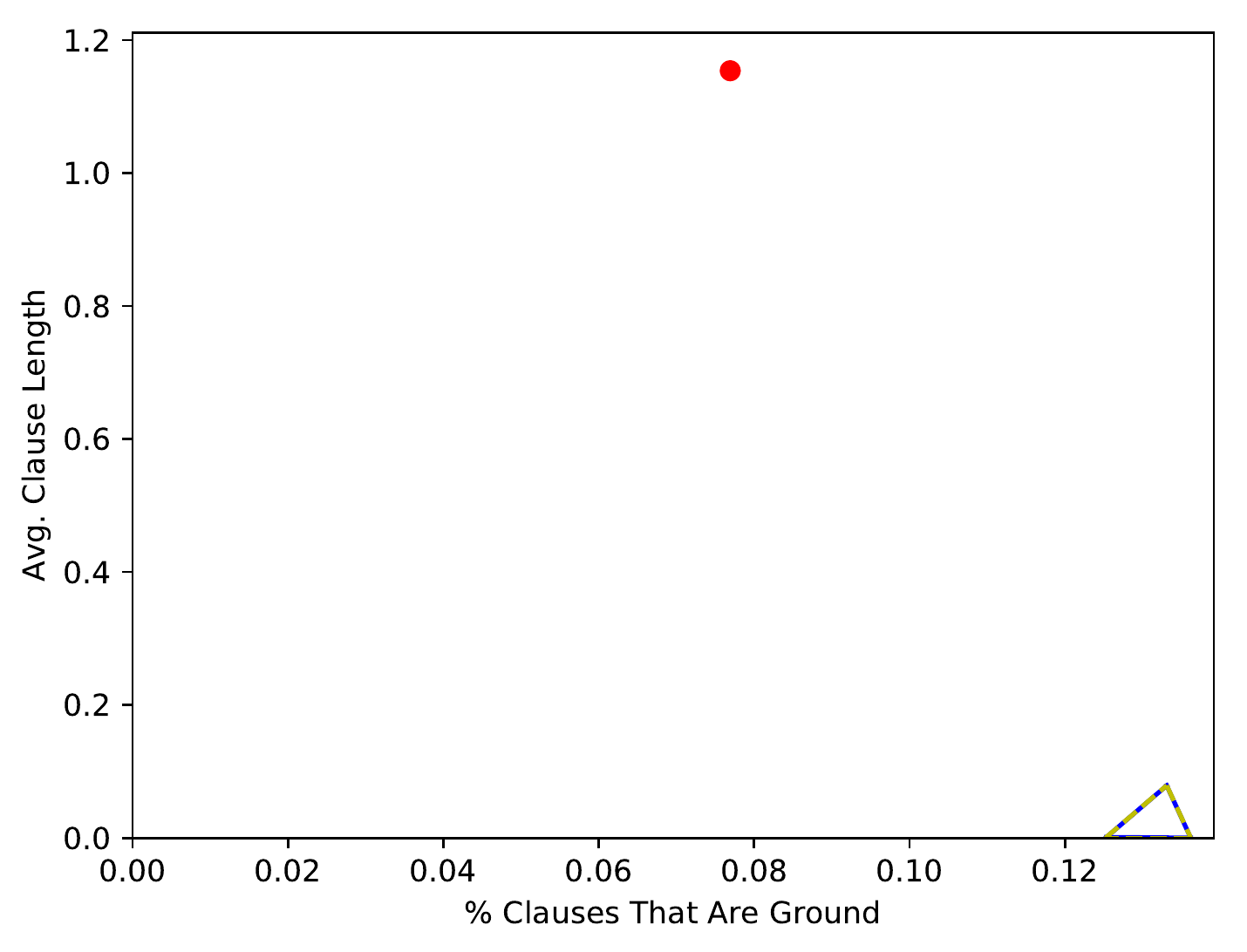} \\
	\includegraphics[width=0.5\textwidth]{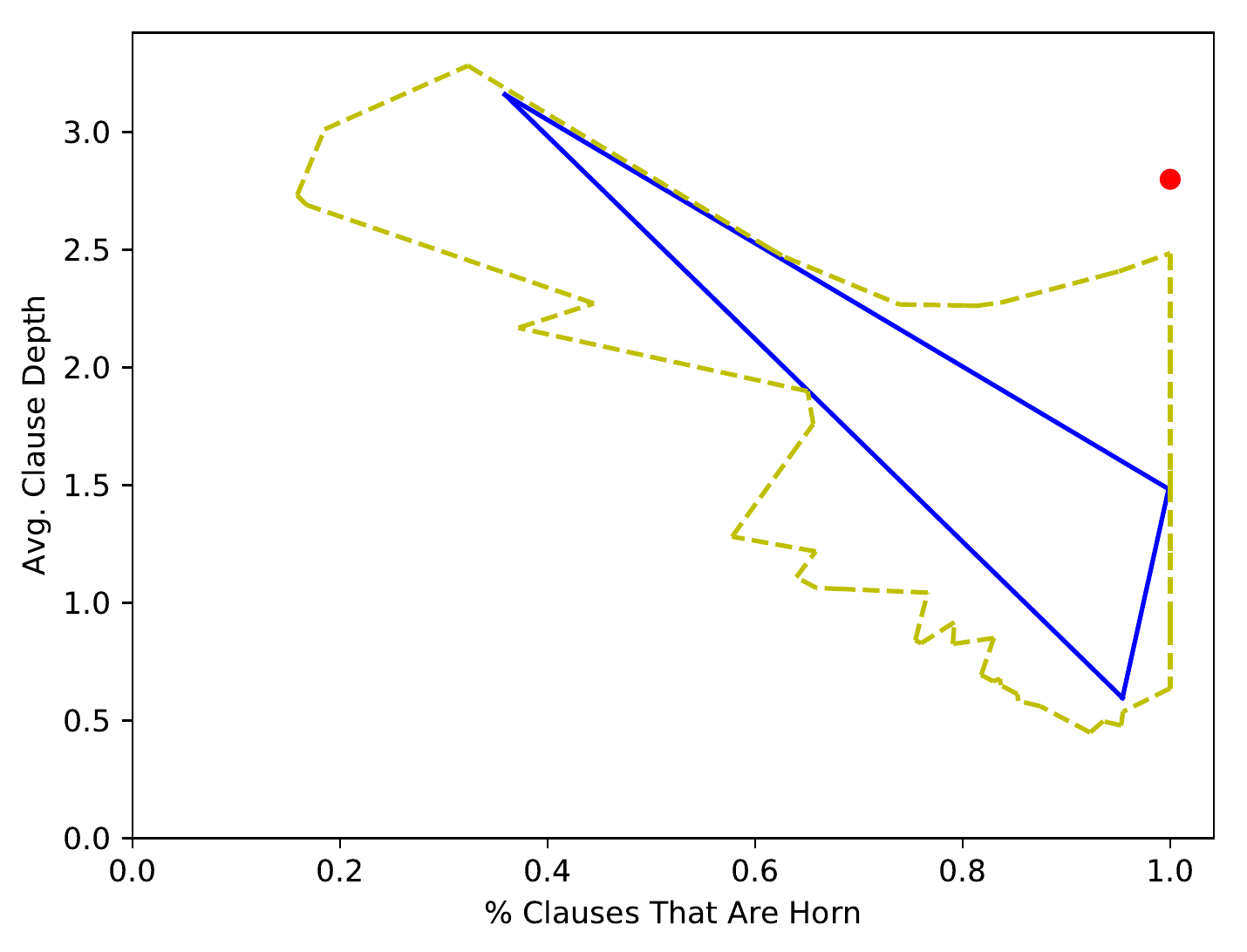}
	\includegraphics[width=0.5\textwidth]{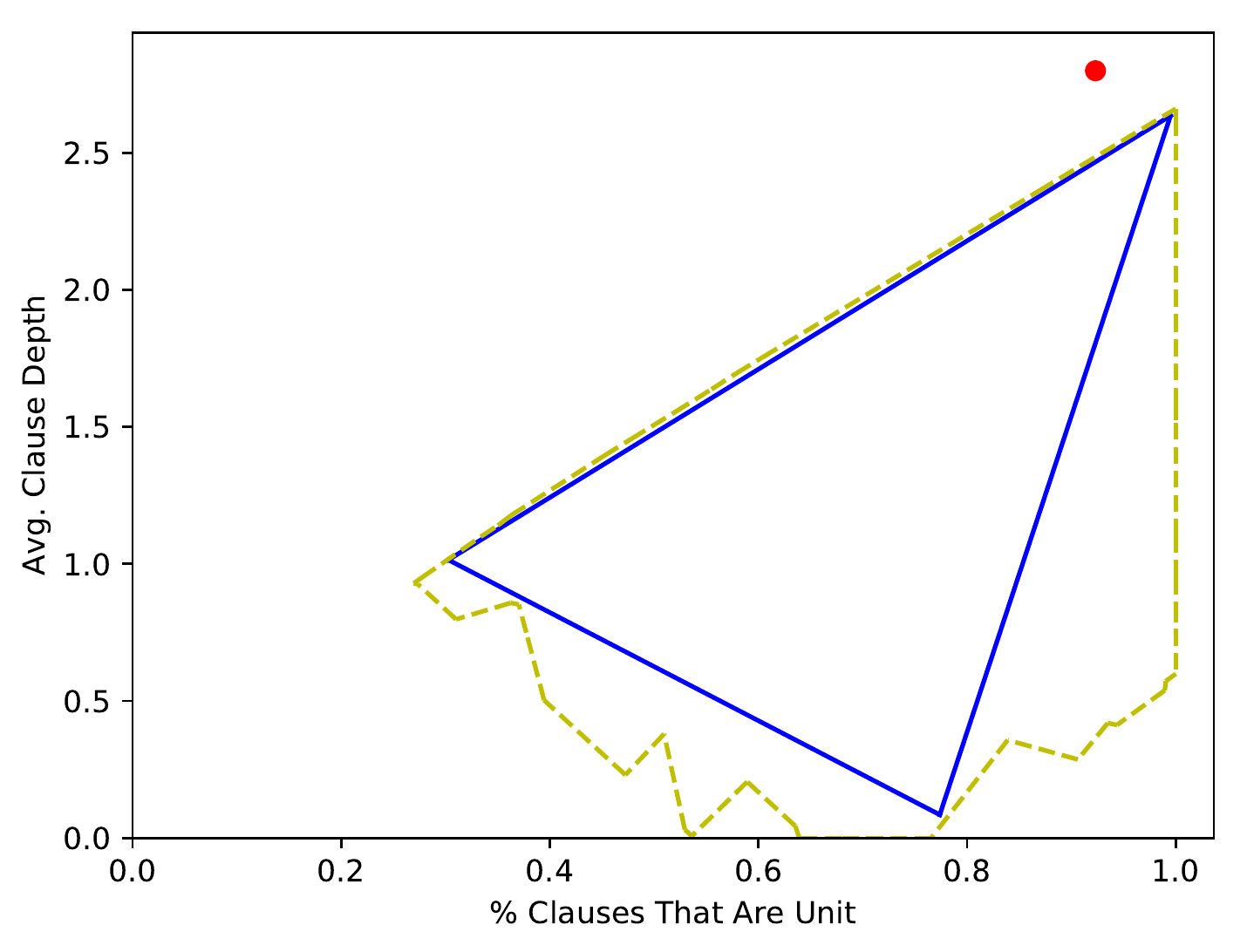} 
\end{figure}

\clearpage
\subsubsection{Drawing Recognition}
The blue lines except for the ones whose vertices fall into the red boxes are inputs.
The red boxes are the symbolic corrections.
Briefly, any sets of lines whose vertices all into them would make the drawing accepted by the neural network.

\begin{figure}[h]
	\includegraphics[width=0.5\textwidth]{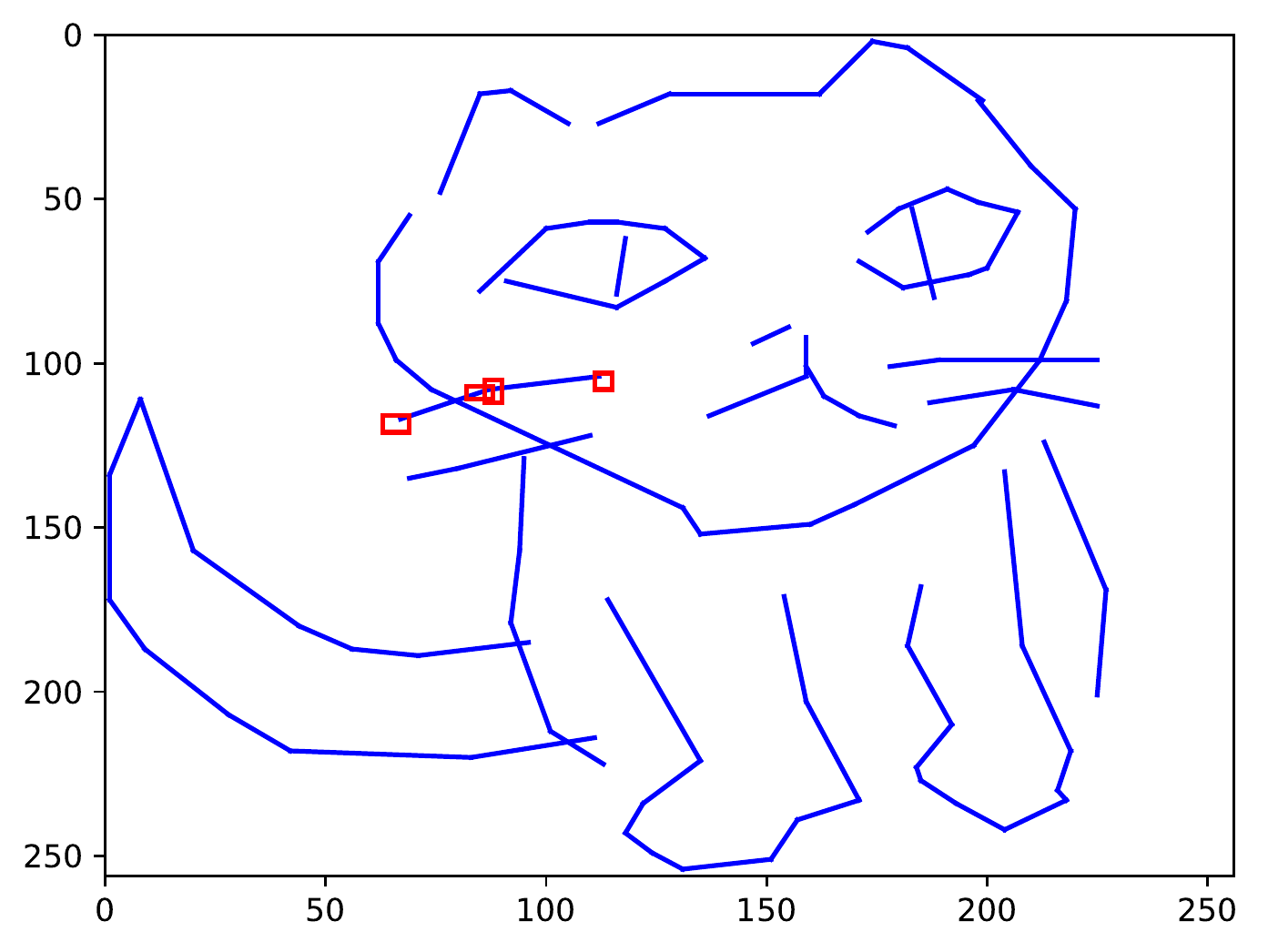}
	\includegraphics[width=0.5\textwidth]{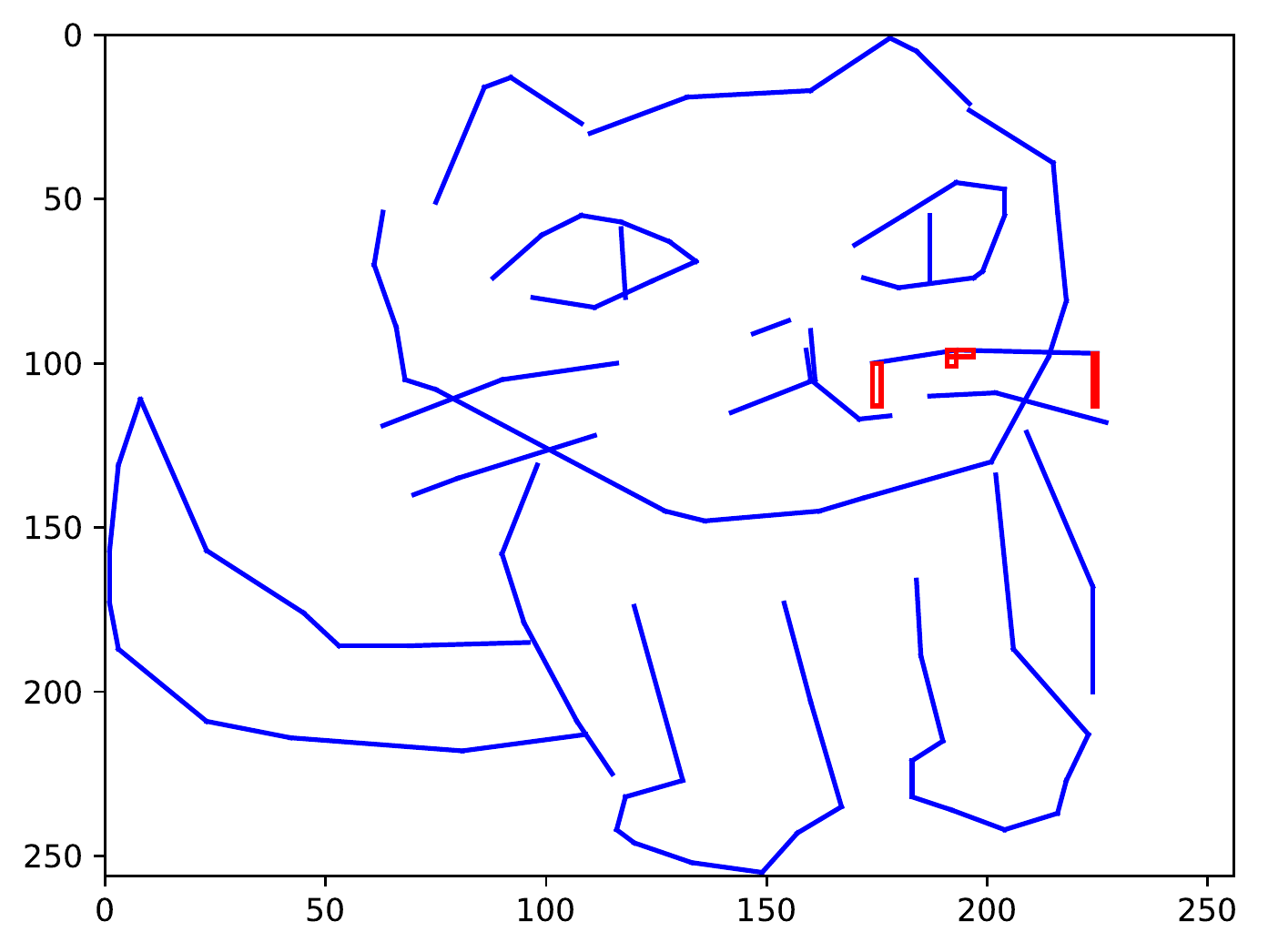} \\
	\includegraphics[width=0.5\textwidth]{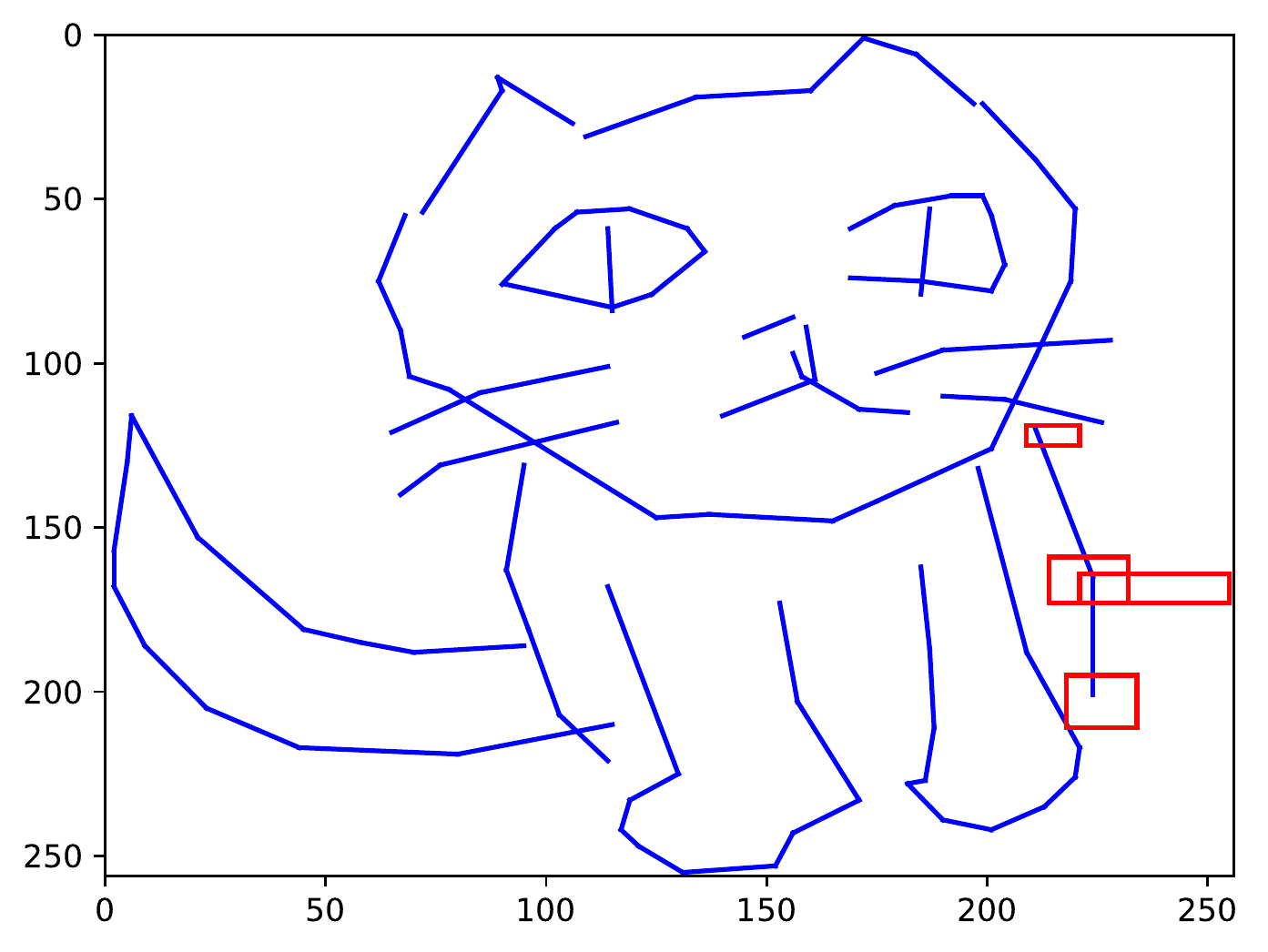}
	\includegraphics[width=0.5\textwidth]{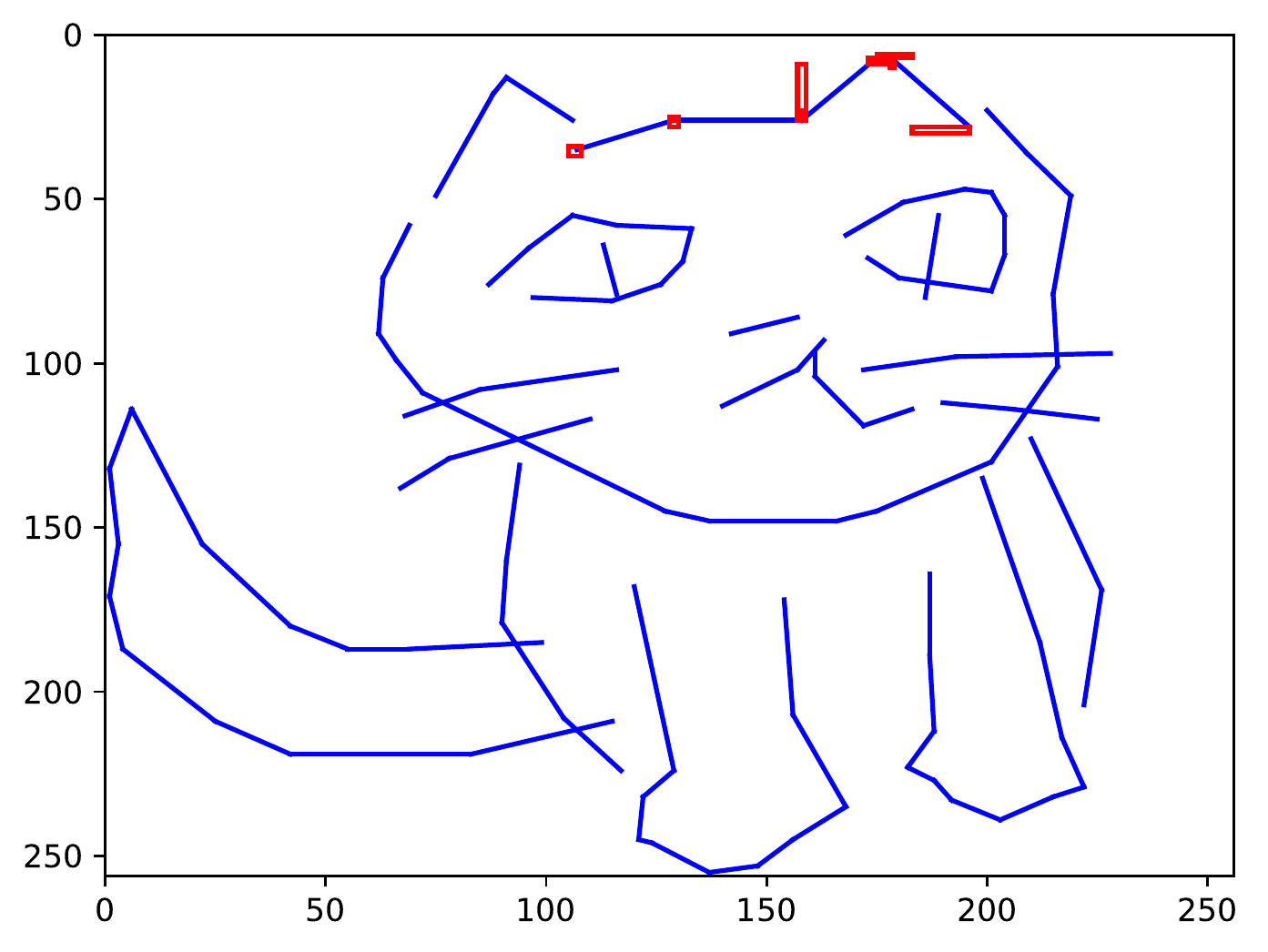} \\
	\includegraphics[width=0.5\textwidth]{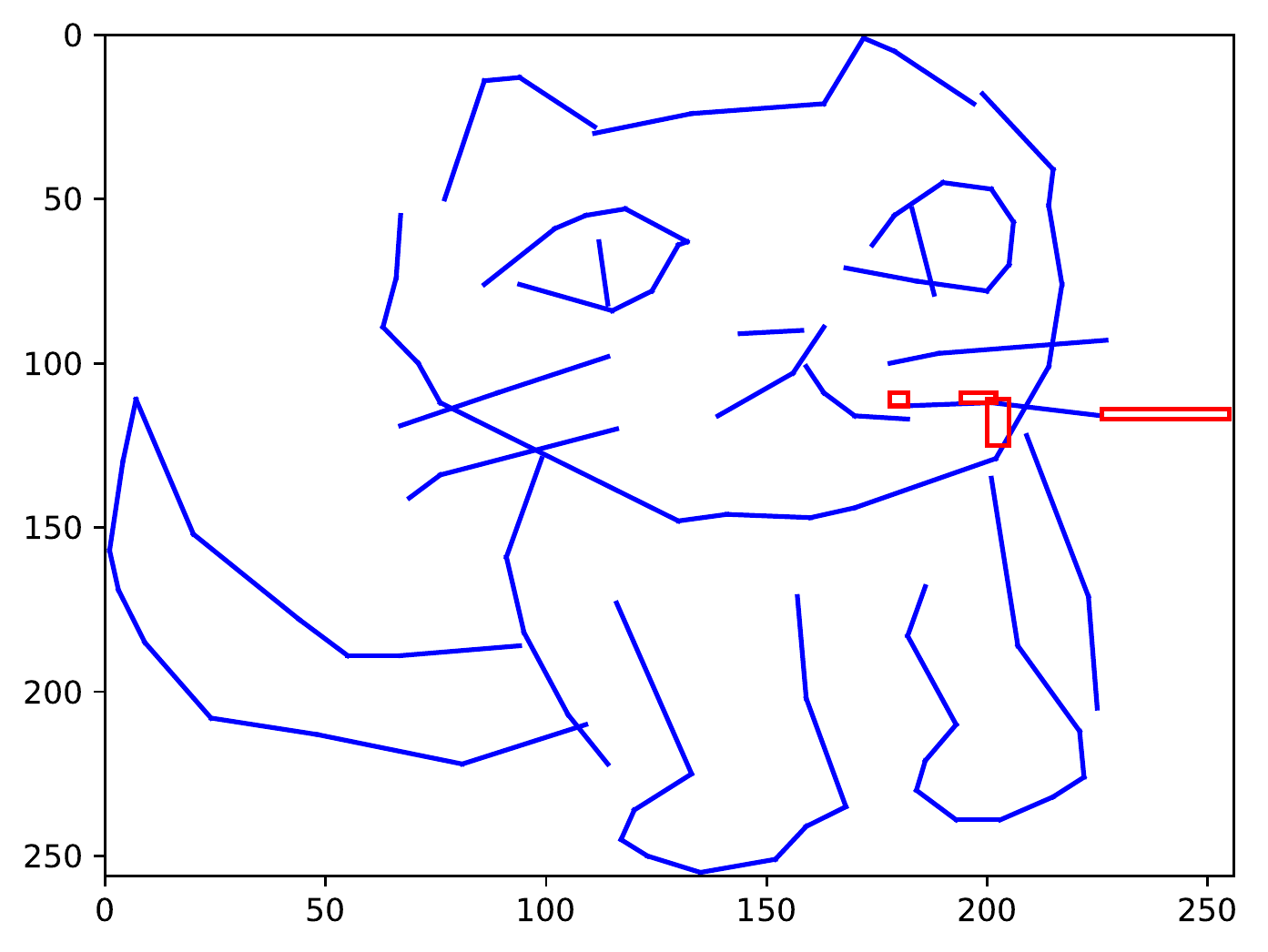}
	\includegraphics[width=0.5\textwidth]{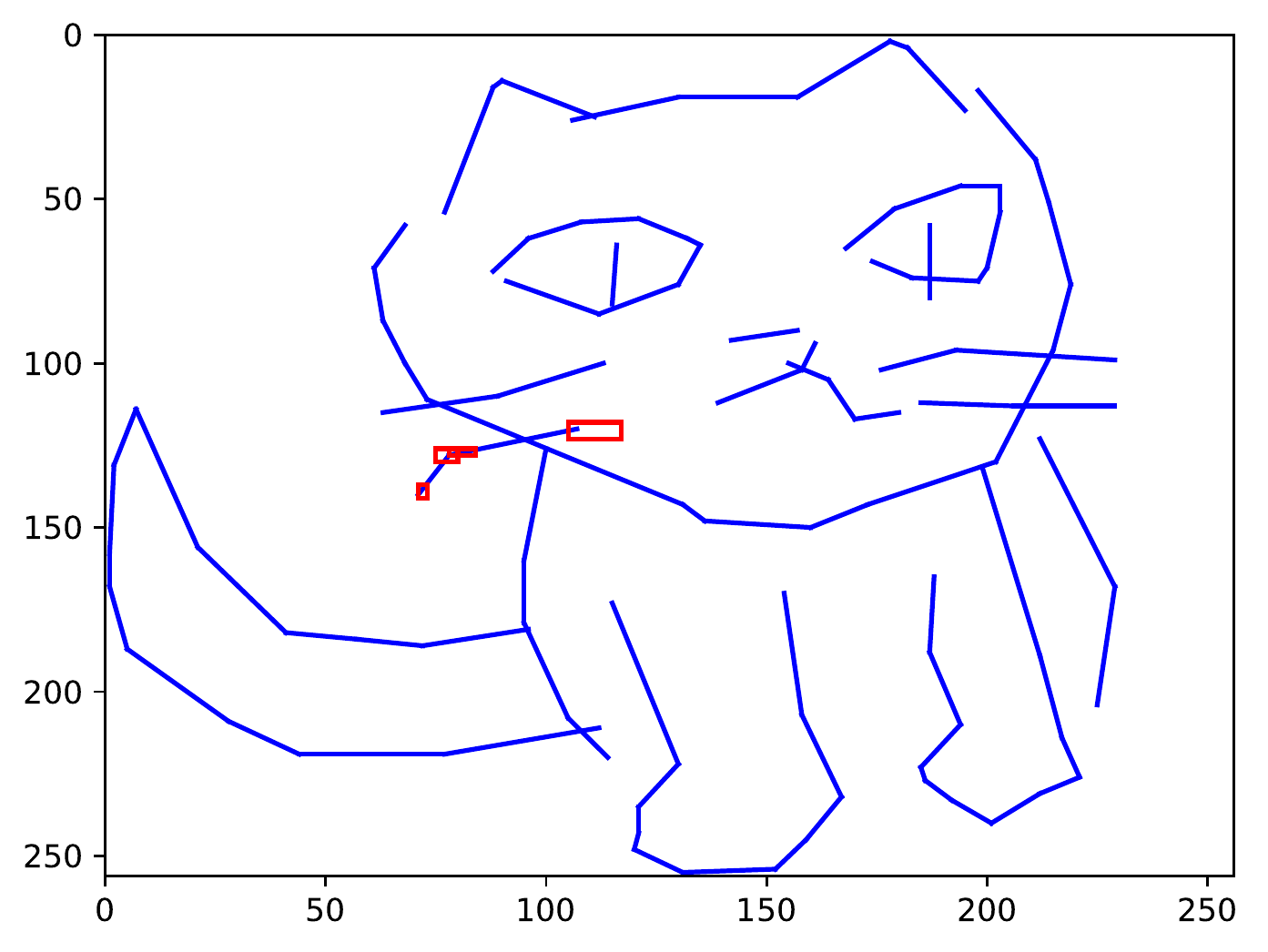} 
\end{figure}

\begin{figure}[t]
	\includegraphics[width=0.5\textwidth]{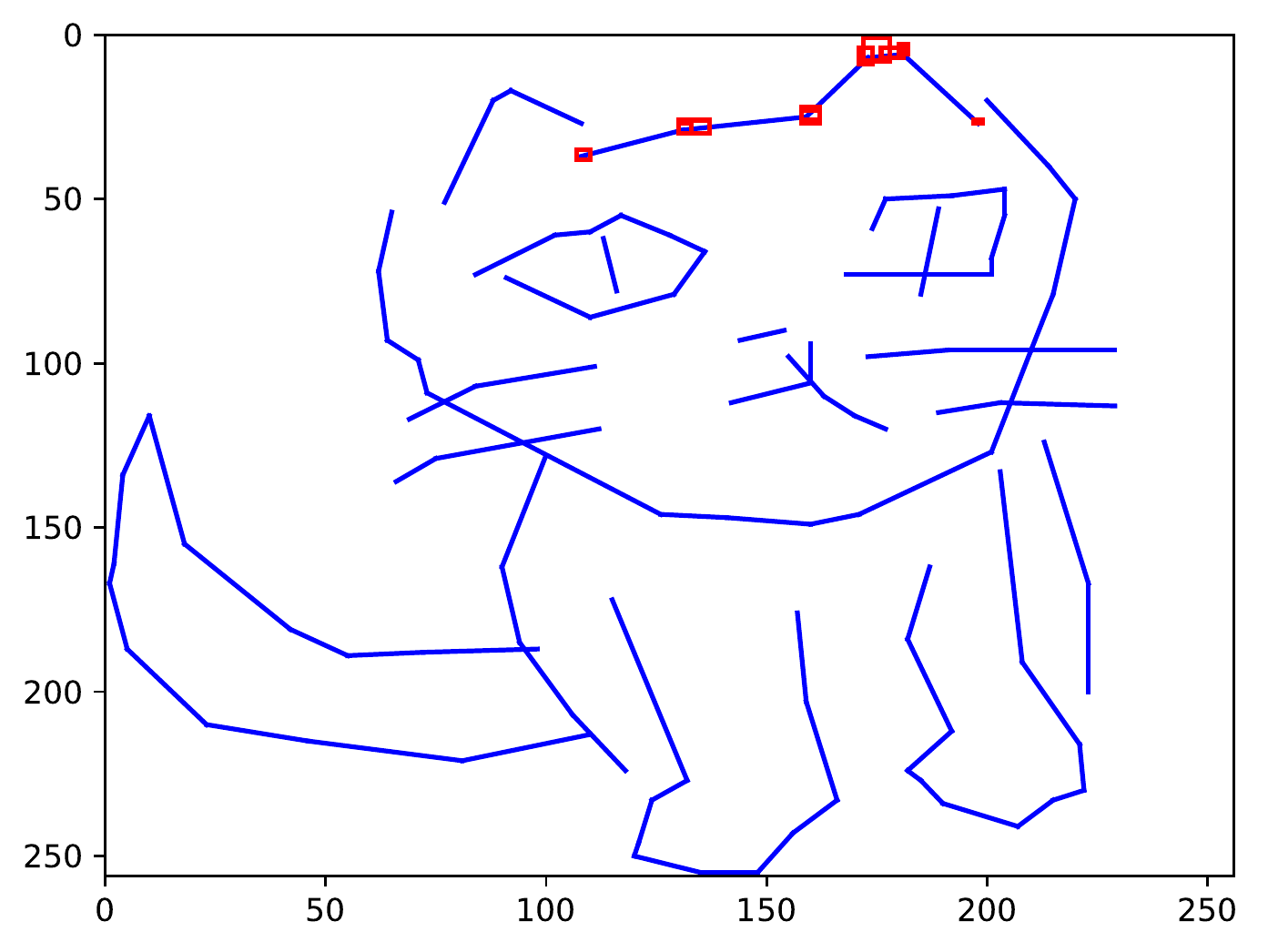}
	\includegraphics[width=0.5\textwidth]{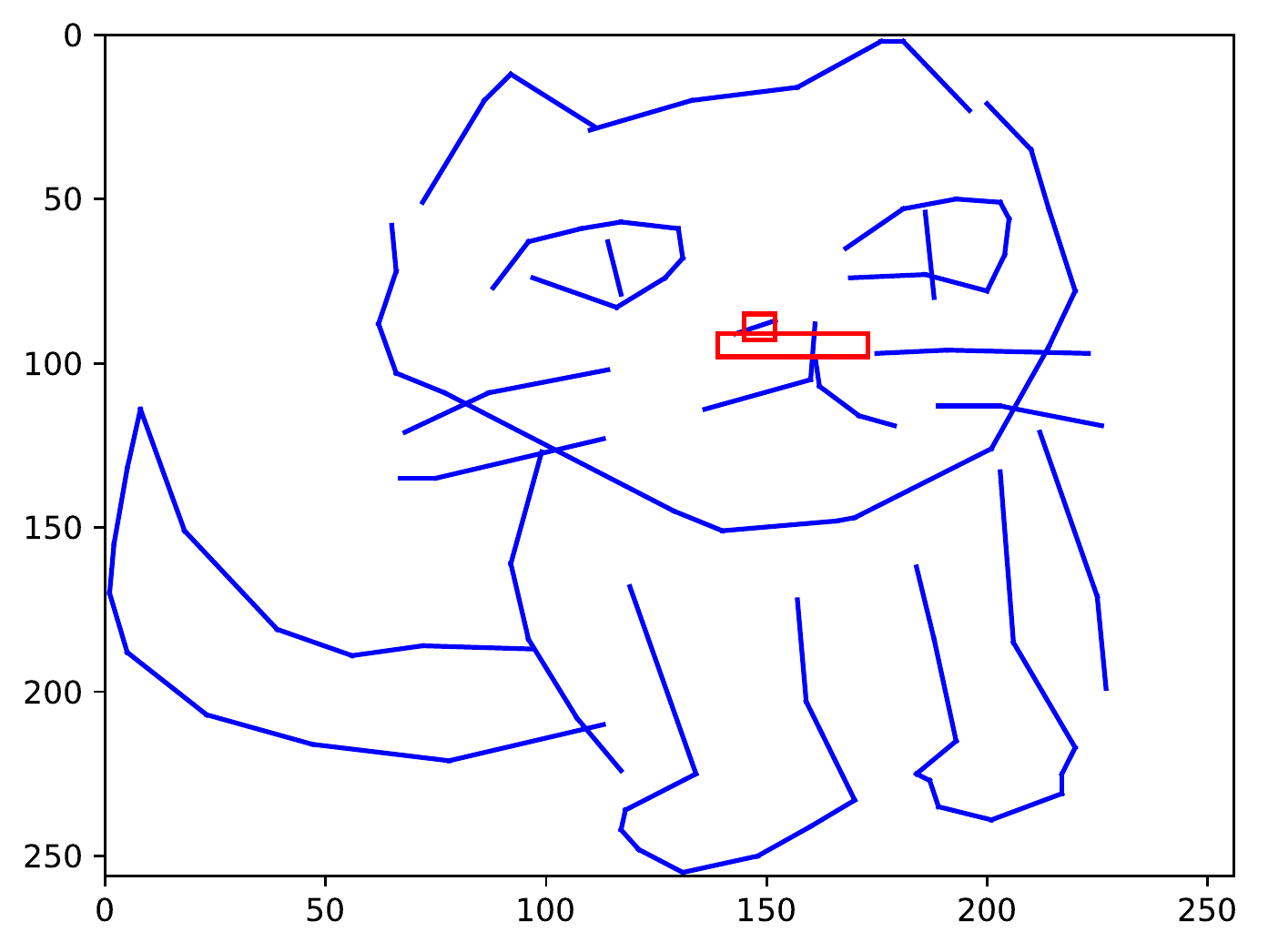} \\
	\includegraphics[width=0.5\textwidth]{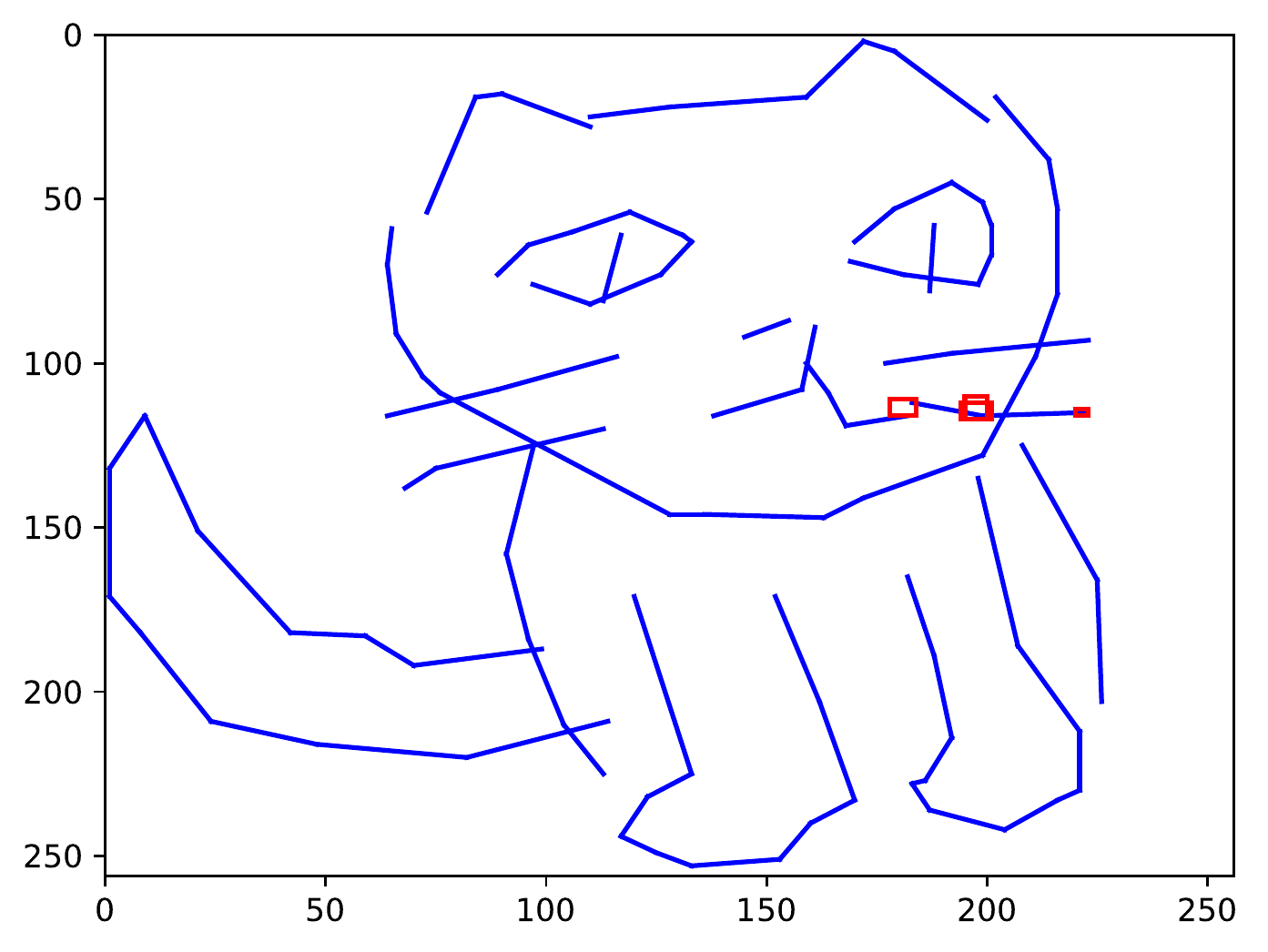}
	\includegraphics[width=0.5\textwidth]{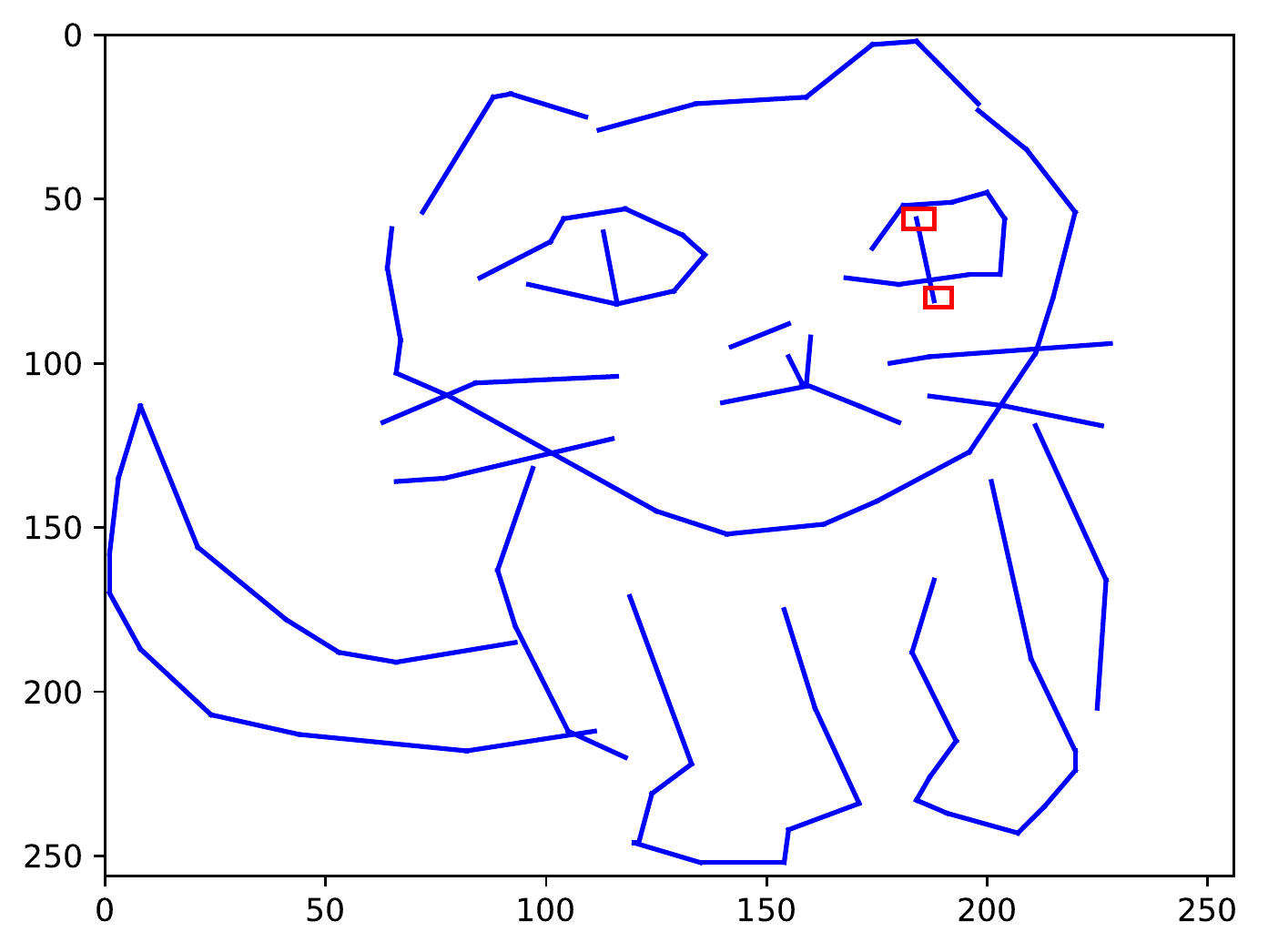} \\
	\includegraphics[width=0.5\textwidth]{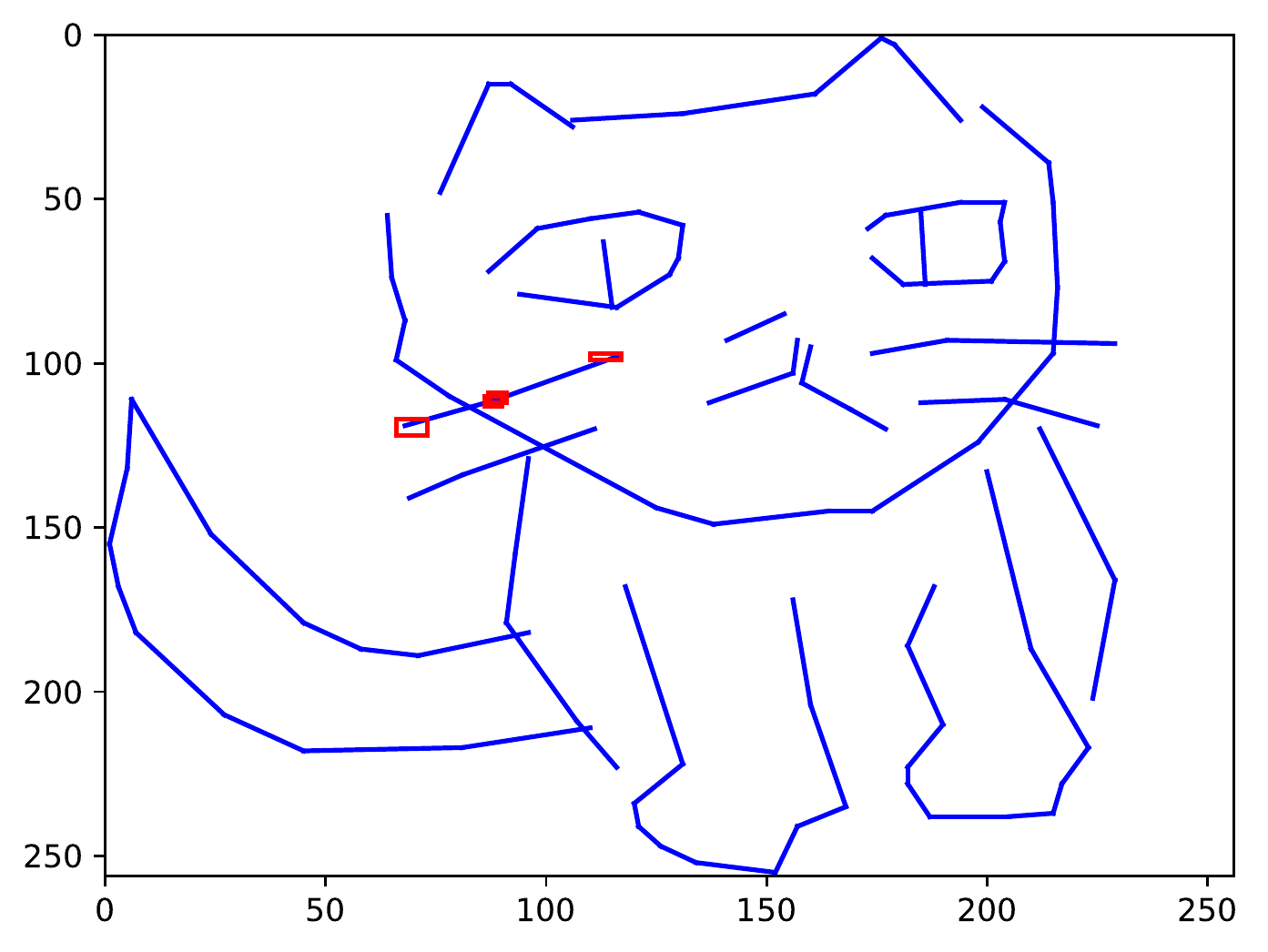}
	\includegraphics[width=0.5\textwidth]{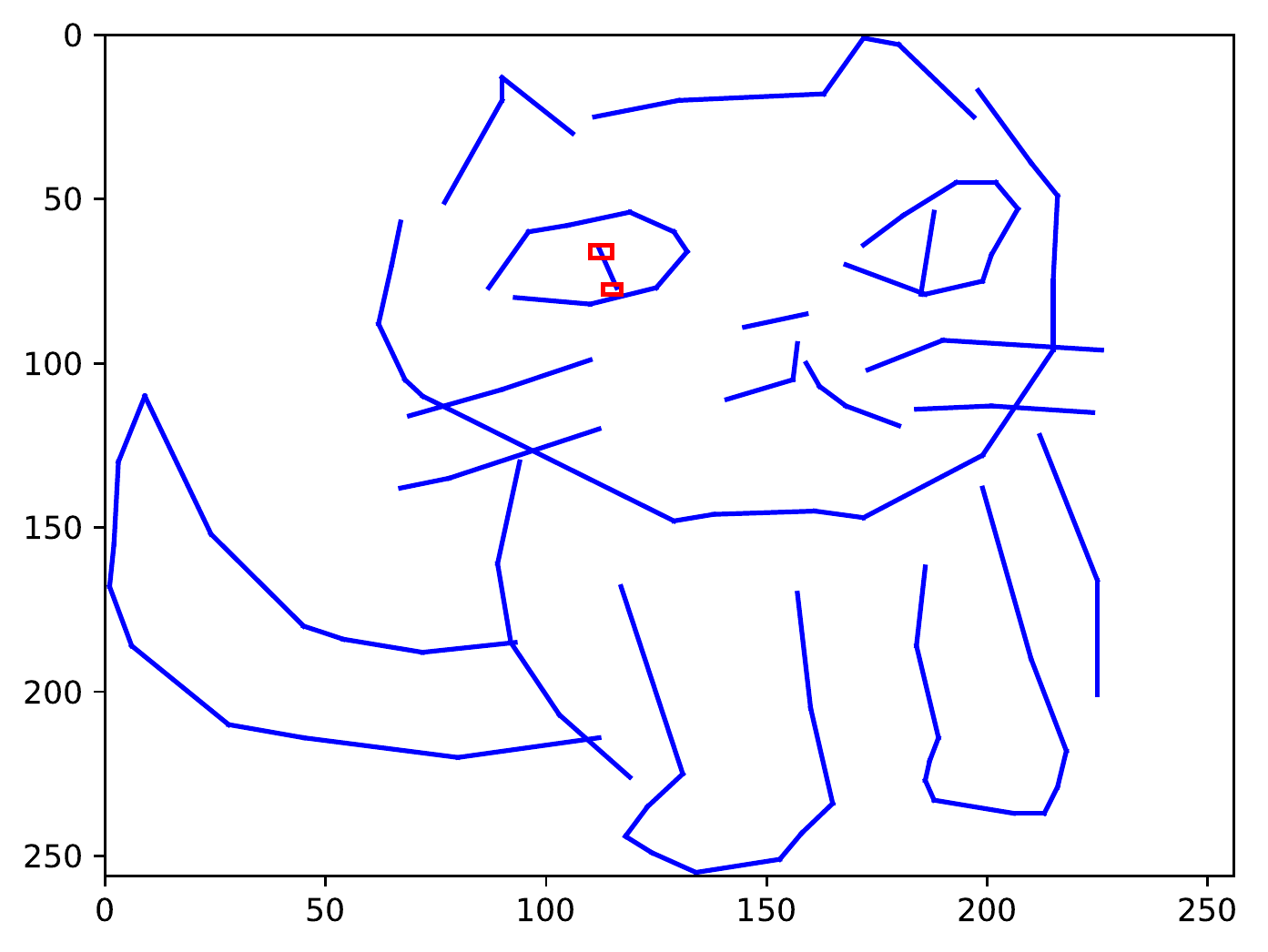} 
\end{figure}

\begin{figure}[t]
	\includegraphics[width=0.5\textwidth]{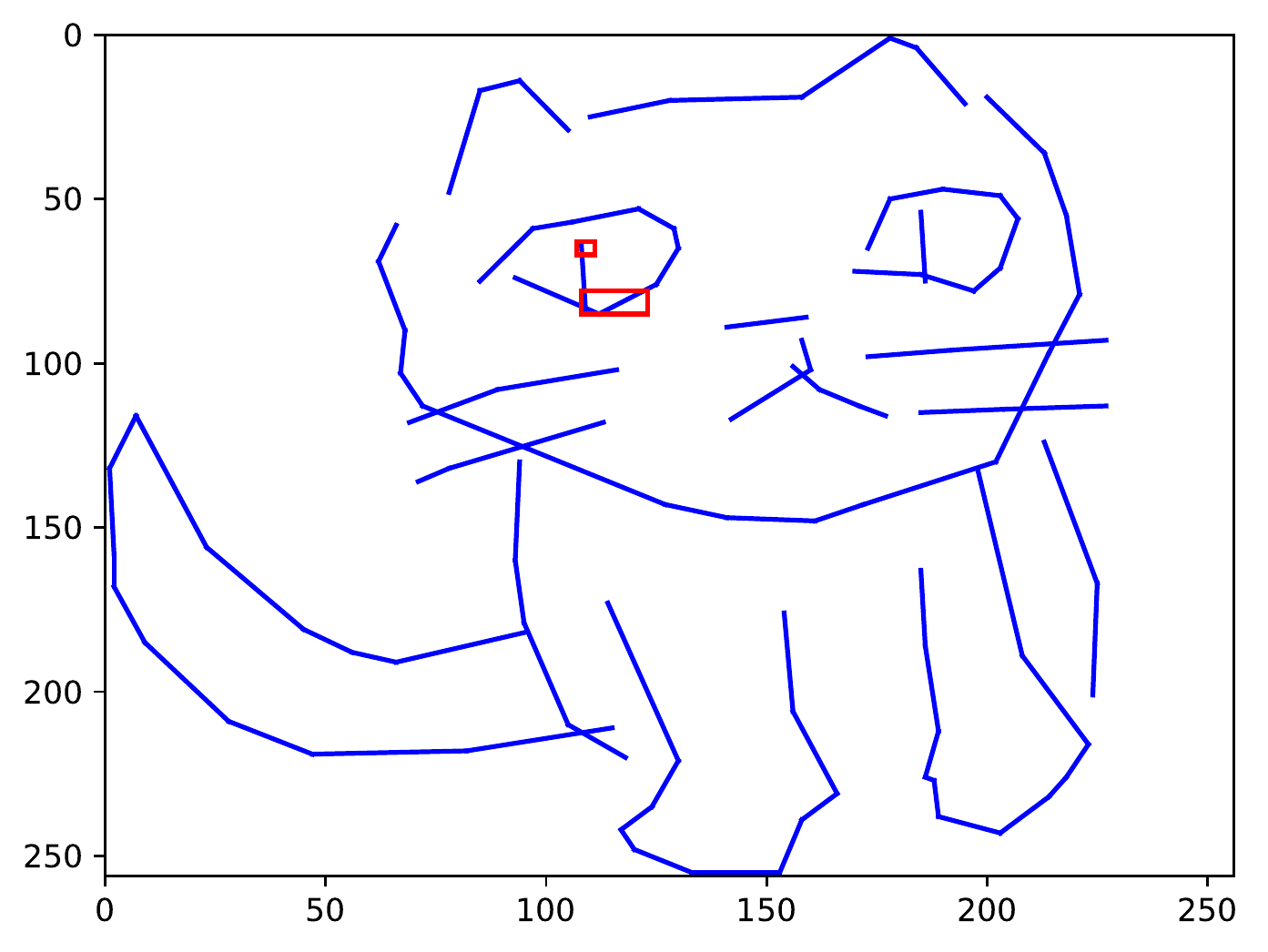}
	\includegraphics[width=0.5\textwidth]{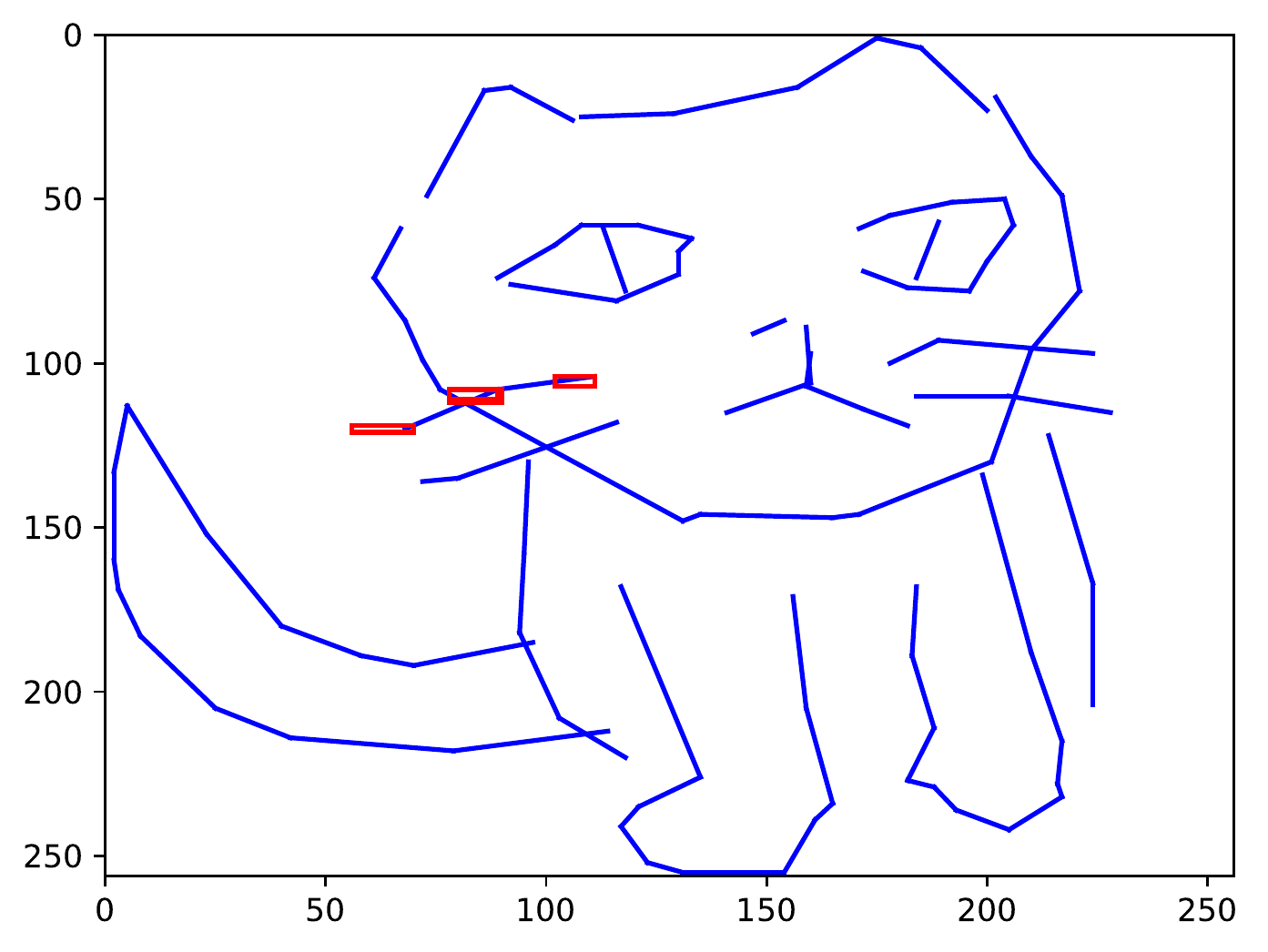} \\
	\includegraphics[width=0.5\textwidth]{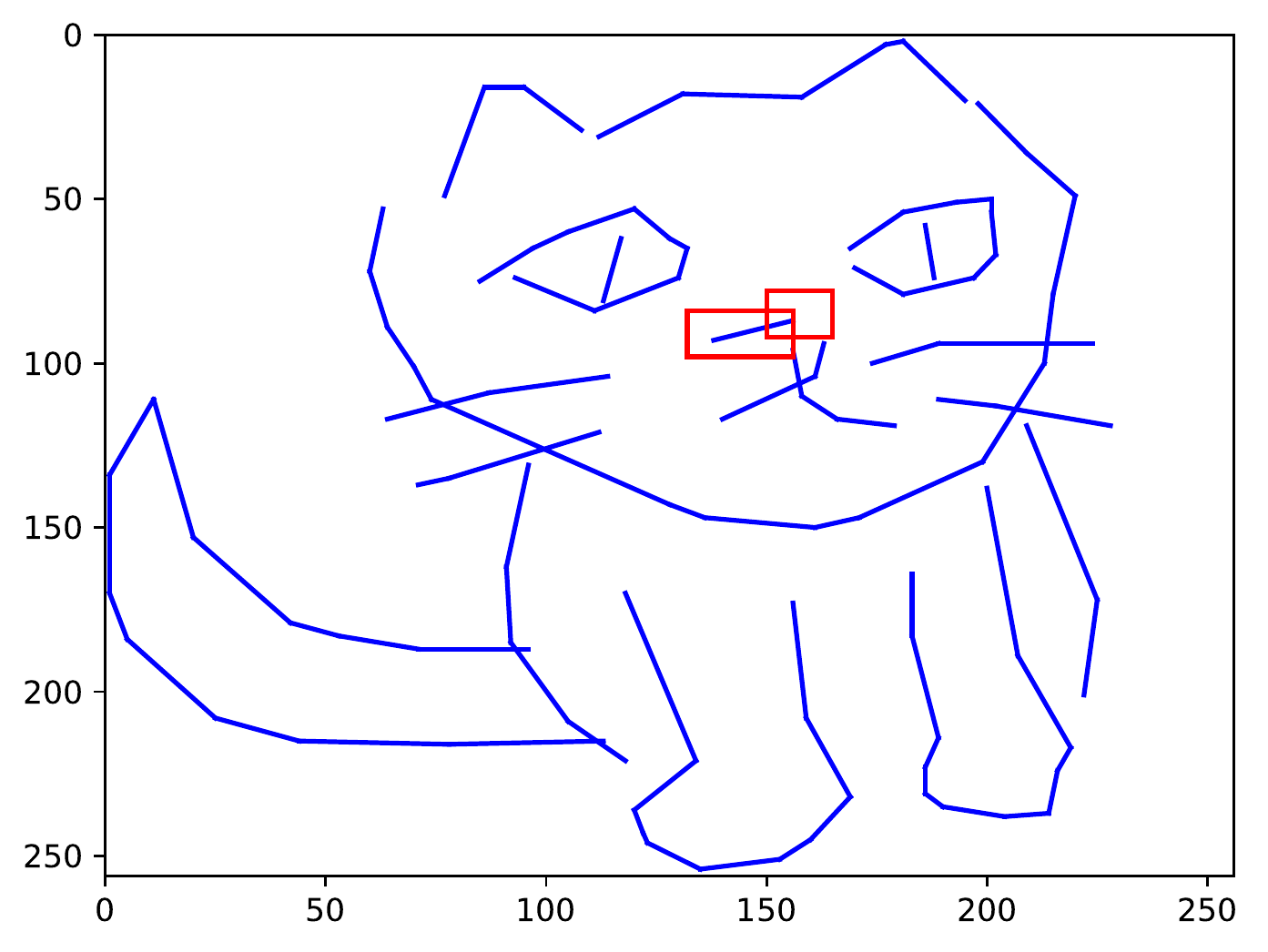}
	\includegraphics[width=0.5\textwidth]{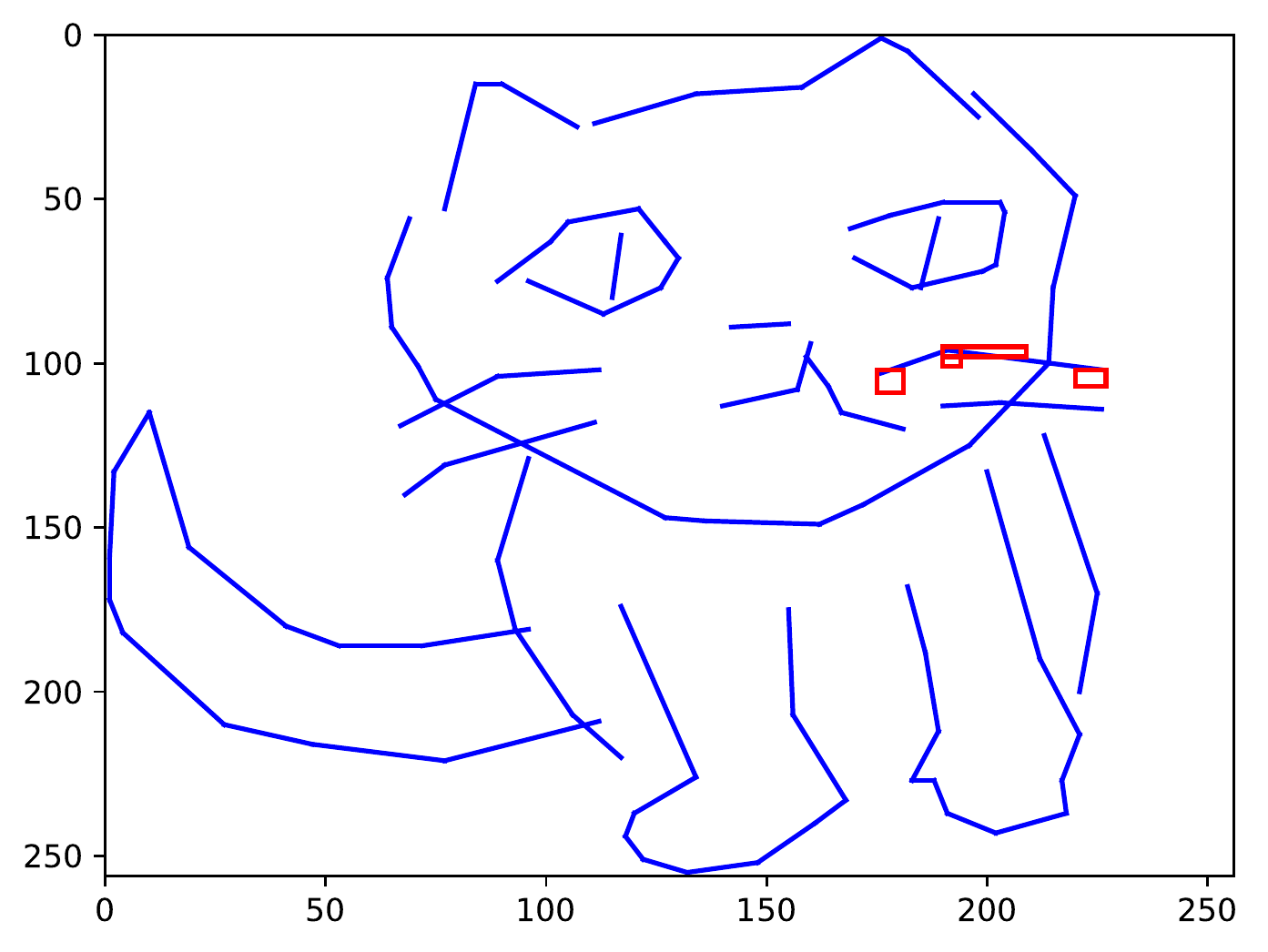} \\
	\includegraphics[width=0.5\textwidth]{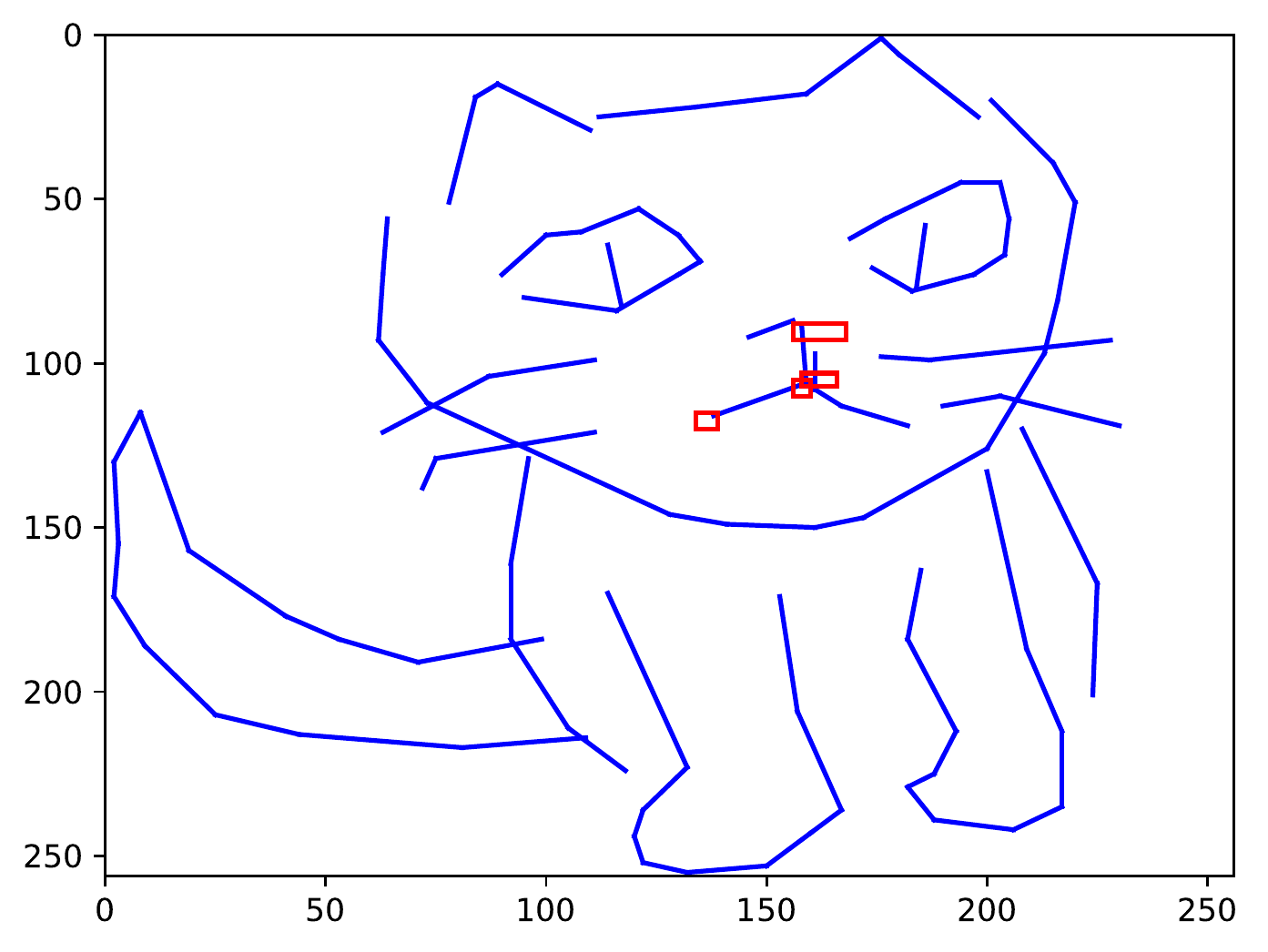}
	\includegraphics[width=0.5\textwidth]{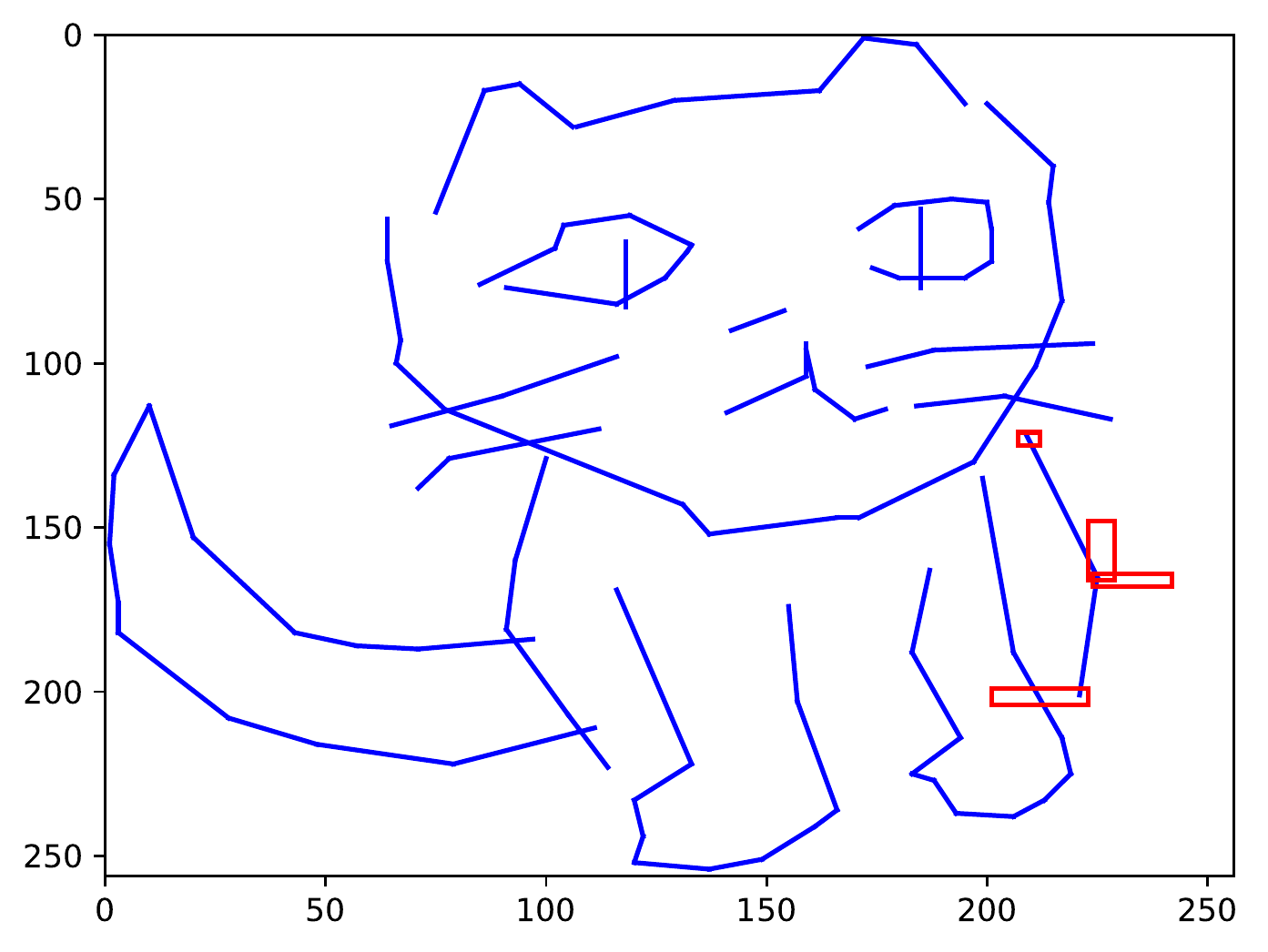} \\
		\includegraphics[width=0.5\textwidth]{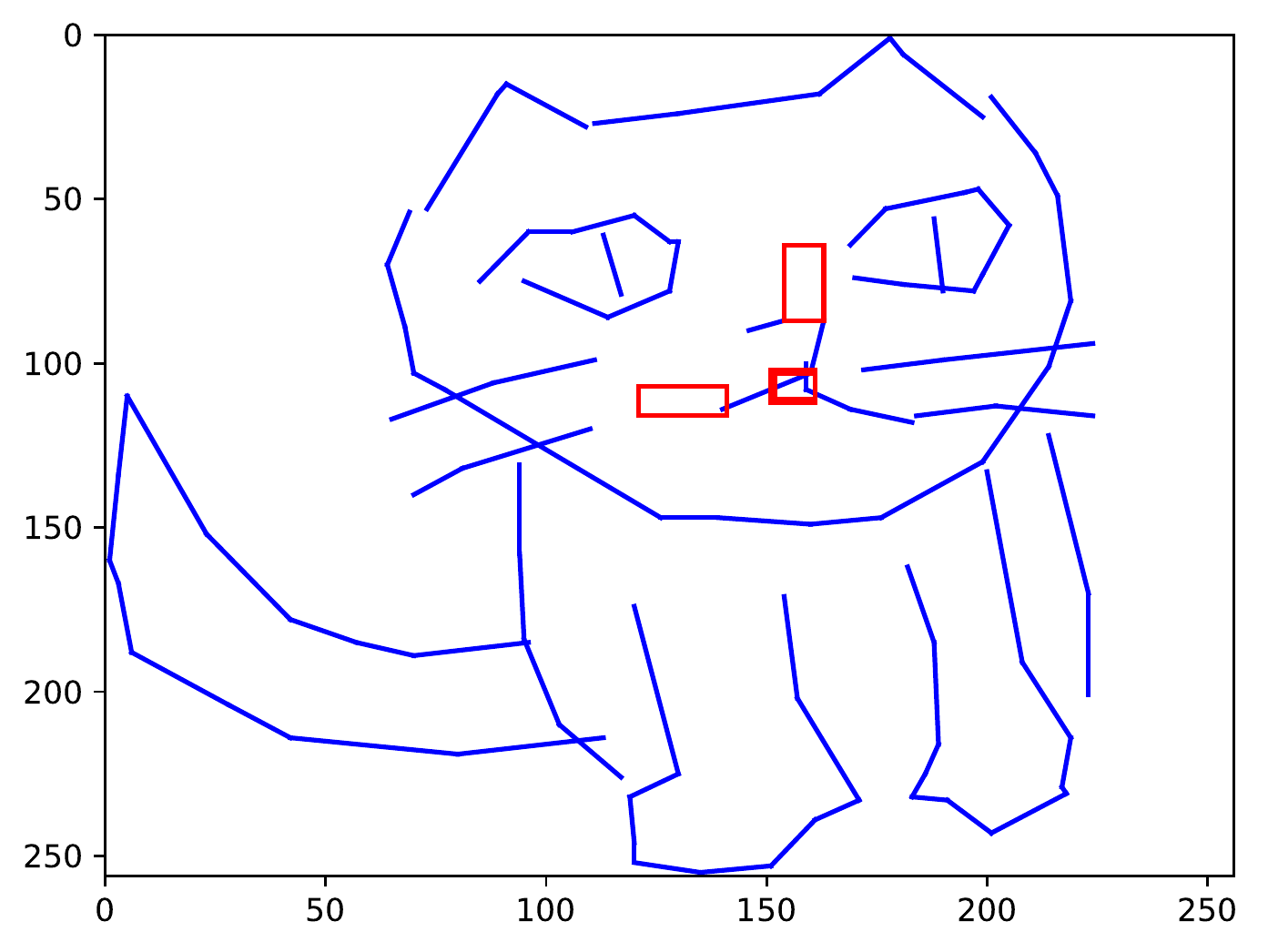}
		\includegraphics[width=0.5\textwidth]{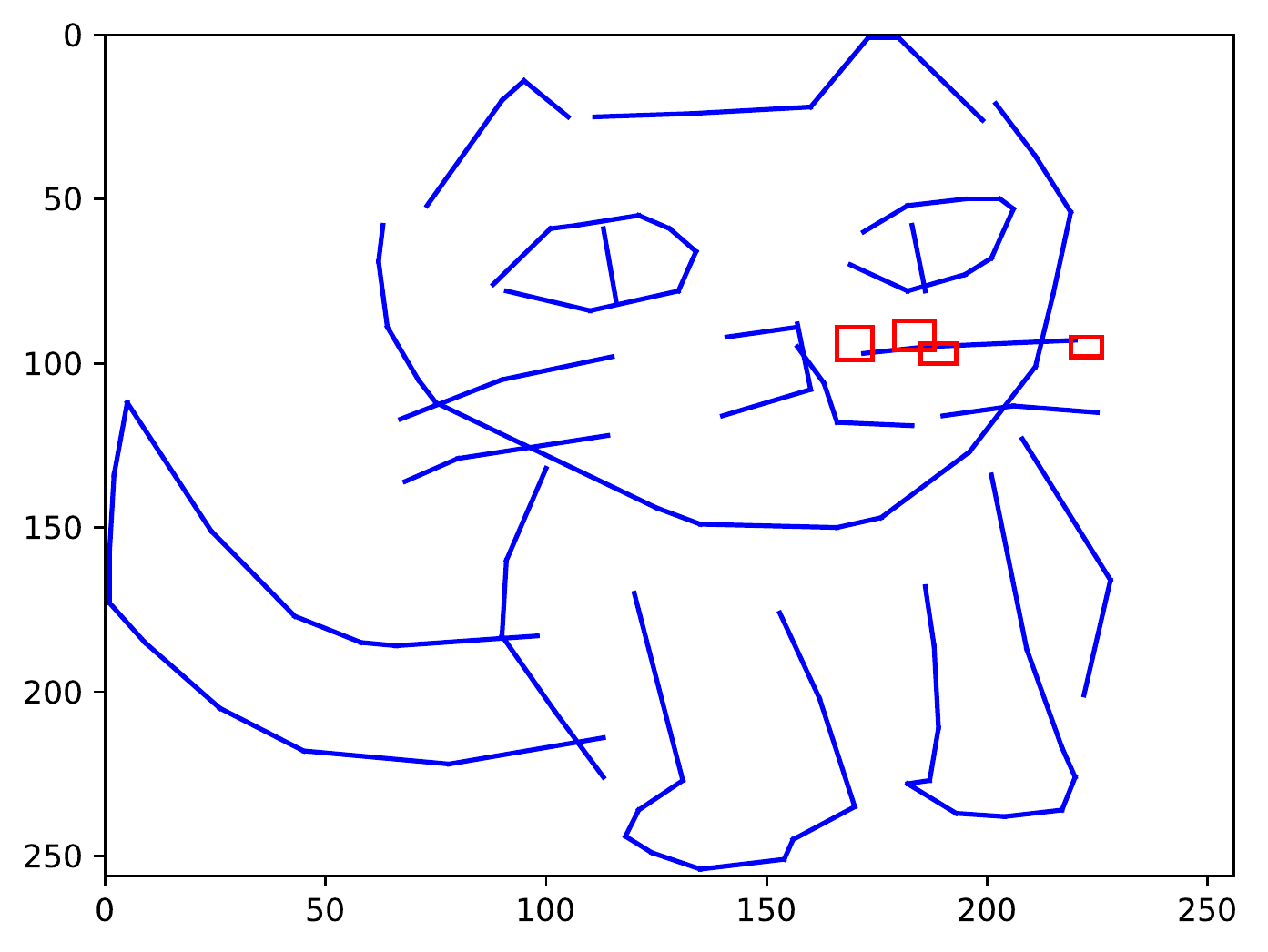} 
\end{figure}